\documentclass[sigconf, language=french,
language=german, language=spanish, language=english]{acmart}
\AtBeginDocument{%
  }

\usepackage{booktabs}  
\usepackage{array}     
\usepackage{multirow}  
\usepackage{subfig}

\acmConference[]{}{}

\renewcommand\footnotetextcopyrightpermission[1]{}
\acmSubmissionID{5258}

\begin{document}

\title{Flow Intelligence: Robust Feature Matching via Temporal Signature Correlation}

\author{Jie Wang}
\affiliation{
    \institution{SIGS, Tsinghua University}
    \city{Shenzhen}
    \country{China}
}
\email{jie-wang24@mails.tsinghua.edu.cn}

\author{CHEN YE GAN}
\affiliation{
    \institution{SIGS, Tsinghua University}
    \city{Shenzhen}
    \country{China}
}
\email{reachfelixgan@gmail.com}

\author{Caoqi Wei}
\affiliation{
    \institution{University of Electronic Science and Technology of China}
    \city{Chengdu}
    \country{China}
}
\email{2022090909004@std.uestc.edu.cn}

\author{Jiangtao Wen}
\affiliation{
    \institution{New York University}
    \city{Shanghai}
    \country{China}
}
\email{jw9263@nyu.edu}

\author{Yuxing Han}
\affiliation{
    \institution{SIGS, Tsinghua University}
    \city{Shenzhen}
    \country{China}
}
\email{yuxinghan@sz.tsinghua.edu.cn}


\begin{abstract}
Feature matching across video streams remains a cornerstone challenge in computer vision. Increasingly, robust multimodal matching has garnered interest in robotics, surveillance, remote sensing, and medical imaging. While traditional rely on detecting and matching spatial features, they break down when faced with noisy, misaligned, or cross-modal data. Recent deep learning methods have improved robustness through learned representations, but remain constrained by their dependence on extensive training data and computational demands. We present Flow Intelligence, a paradigm-shifting approach that moves beyond spatial features by focusing on temporal motion patterns exclusively. Instead of detecting traditional keypoints, our method extracts motion signatures from pixel blocks across consecutive frames and extract temporal motion signatures between videos. These motion-based descriptors achieve natural invariance to translation, rotation, and scale variations while remaining robust across different imaging modalities. This novel approach also requires no pretraining data, eliminates the need for spatial feature detection, enables cross-modal matching using only temporal motion, and it outperforms existing methods in challenging scenarios where traditional approaches fail. By leveraging motion rather than appearance, Flow Intelligence enables robust, real-time video feature matching in diverse environments.
\end{abstract}





\maketitle

\section{Introduction}

\begin{figure*}[t]
    \centering
    \begin{tabular}{c@{\hspace{1mm}}c@{\hspace{1mm}}c@{\hspace{1mm}}c}
        \raisebox{8pt}{\rotatebox{90}{}} &
        \subfloat[MINIMA\_LoFTR on RGB images under LVC]{\includegraphics[width=0.48\textwidth]{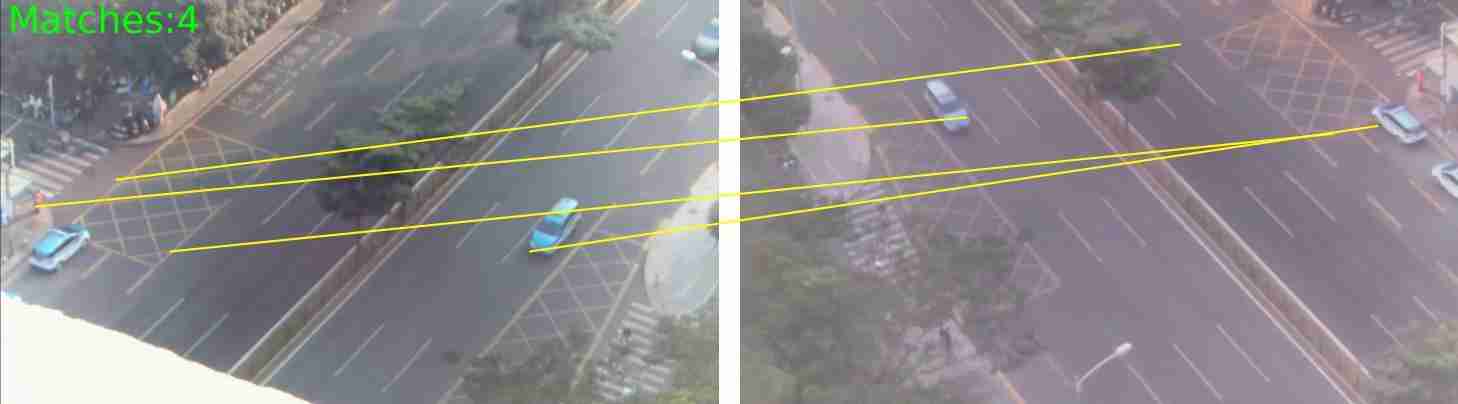}} &
        \subfloat[Flow Intelligence (Ours) on RGB images under LVC]{\includegraphics[width=0.48\textwidth]{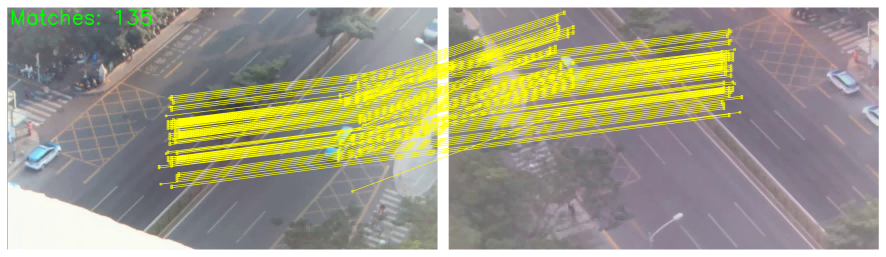}} \\[-2mm]
        
        \raisebox{8pt}{\rotatebox{90}{}} &
        \subfloat[MINIMA\_LoFTR on cross-modal (RGB to Infrared) images]{\includegraphics[width=0.48\textwidth]{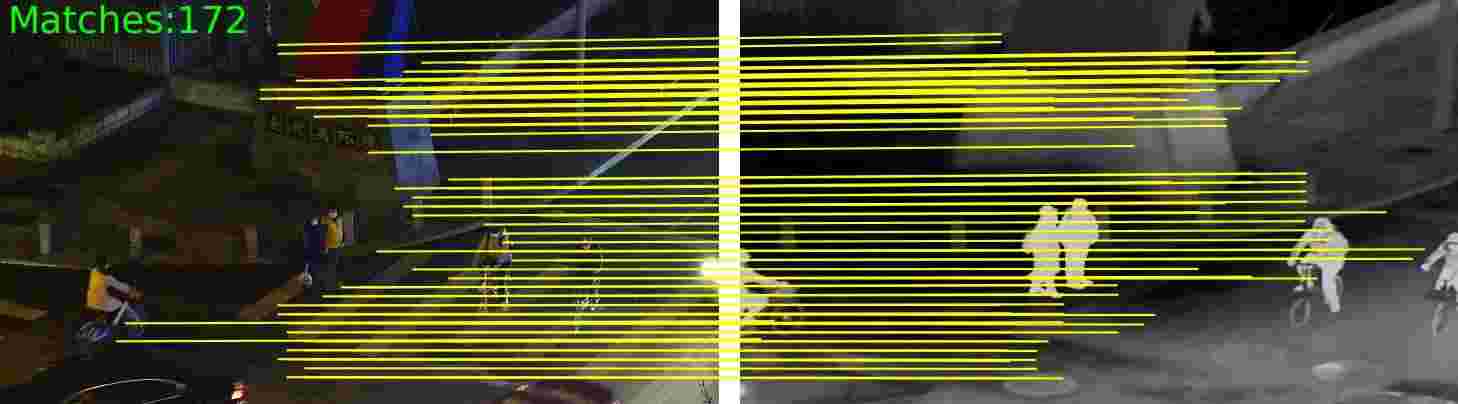}} &
        \subfloat[Flow Intelligence (Ours) on cross-modal (RGB to Infrared) images]{\includegraphics[width=0.48\textwidth]{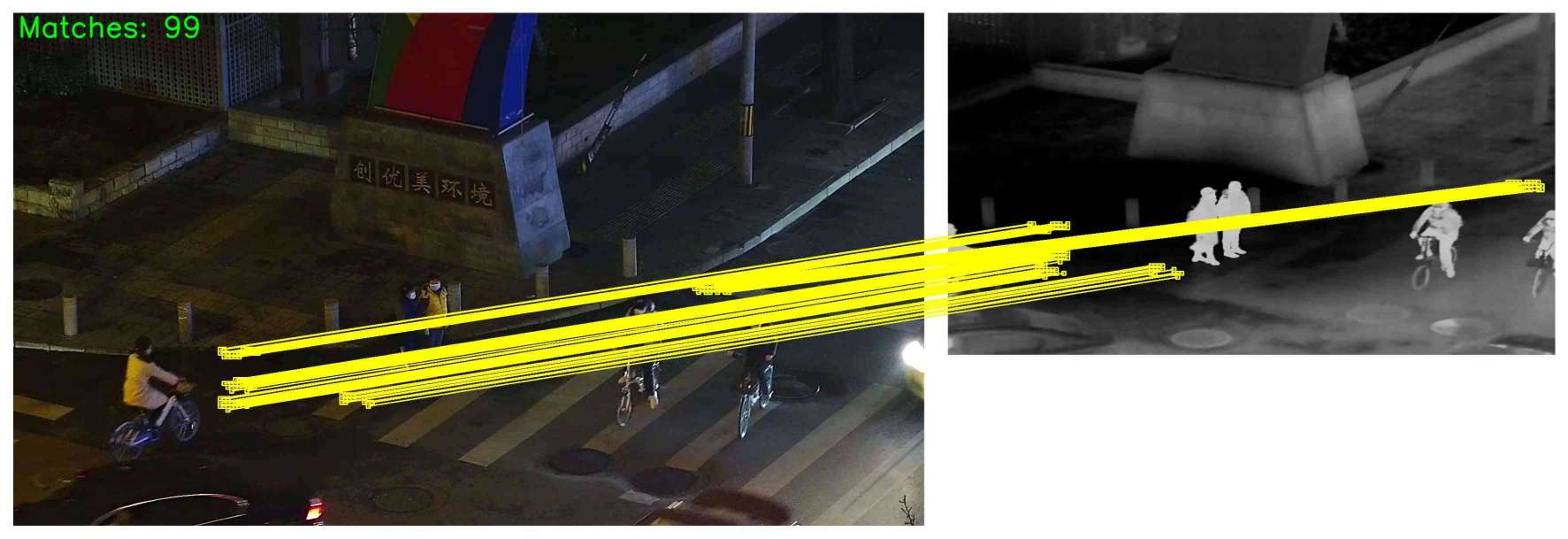}} \\[-3mm]
    \end{tabular}
    
    \caption1{Comparison between MINIMA\_LoFTR (deep learning SOTA) and our Flow Intelligence method. Under large viewpoint changes, MINIMA\_LoFTR fails to establish correct matches (a), while our method performs robustly (b). In RGB-IR matching (c vs d), our method achieves comparable results despite the lower resolution of IR images. MINIMA\_LoFTR resizes images to the same resolution. See Section~\ref{sec:Qualitative Matching Results} and the Appendix for details.}
    \label{fig:first_fig}
    \vspace{-4mm}
\end{figure*}

The fundamental challenge of establishing reliable point correspondences across video streams remains a cornerstone problem in computer vision, underpinning applications from 3D reconstruction to multi-view tracking. Specifically, cross-modal matching becomes more important in today's world, driven by applications such as combining RGB with thermal infrared imagery for night-time autonomous vehicle navigation, visible-light with depth sensors for robotics in visually challenging conditions, and multimodal medical imaging (MRI, CT, ultrasound) to improve diagnostic accuracy. In these systems, robust multimodal matching improve situational awareness, accurate object tracking, and enhanced 3D reconstruction. However, modality-specific visual differences pose unique conditions that challenge traditional appearance-based methods. 

Classical solutions like SIFT \cite{lowe_distinctive_2004} and SURF\cite{bay_speeded-up_2008}, has focused on detecting and matching distinctive spatial features across individual frames. BRIEF\cite{hutchison_brief_2010}, BRISK \cite{leutenegger_brisk_2011}, and ORB\cite{noauthor_orb_nodate} improves upon feature descriptions for efficiency and scale-sensitiveness. While these approaches revolutionized the field by introducing scale and rotation invariant descriptors, they fundamentally rely on spatial features. 

Recent advances in deep learning have pushed the capabilities of feature matching. Notable works include D2-Net\cite{dusmanu_d2-net_2019} which employs join detector-descriptor convolutional neural nets to identify robust keypoints, SuperGlue\cite{sarlin_superglue_2020} which uses graph neural nets to contextually match features, and UDIS++\cite{nie_parallax-tolerant_2023} which leverages a Resnet50 backbone for semantic feature extraction and CNN-based Contextual Correlation \cite{nie_depth-aware_2022} for feature correlation. Transformer-based approaches like LoFTR\cite{sun2021loftr} and ASpanFormer\cite{chen_aspanformer_2022} further advanced the field by eliminating explicit feature detection entirely. Despite their impressive performance, whether based on handcrafted features or learned representations, these methods inherently reply on spatial patterns within individual frames. Consequently, they require either careful engineering of invariant features or large-scale training datasets to achieve reasonable robustness.

Further, most existing solutions are tailored to spatial matching in RGB images, which assumes visual appearance consistency across videos—a premise that breaks down in multimodal scenarios. \cite{aytar2017cross} finds that spatial matching methods tend to extract features that generalize poorly across different sensing modalities, rendering cross-modal matching unreliable or even infeasible. 

In this paper, we tackle two critical challenges: achieving reliable matching independent of spatial appearance difference and developing a general purpose multi-modal matching framework. We present \textit{Flow Intelligence}, a paradigm-shifting approach that reconceptualizes feature matching in videos by focusing on temporal motion patterns rather than spatial appearances. Our findings suggest that corresponding points across different views or modalities should exhibit similar temporal motion patterns, even when their spatial appearances differ dramatically. Instead of following a classical detect and match spatial features framework, we take inspiration from video coding and adopt a motion-from-pixel block approach. Specifically, we partition each video frame into progressive block sizes (4x4, 8x8, etc.) and extract temporal motion signatures for each block across multiple frames so that blocks across videos can be matched by comparing their temporal correlation patterns.

This approach offers several compelling advantages. Unlike deep learning, it requires no training data or learned representations. It is not only robust, scaling and rotational invariant (like SIFT and SURF), but also achieves natural appearance invariance, enabling cross-modal matching. Finally, it remains computationally efficient by avoiding the overhead of feature learning and dense descriptor matching.

We recognize that videos captures from multiple viewpoints or across modalities may not be synchronized with each other. Our method is robust to moderate temporal misalignments, where even when videos are not perfectly synchronized, as long as capture devices record at a uniform rate, the method can continue to reliably matches temporal patterns(In our preliminary experiments, we tested scenarios with offsets of up to a dozen frames and observed that the matching performance remained stable under such slight desynchronization.). 


In addition, a novel temporal correlation-based framework is proposed that eliminates the need for spatial feature detection while providing competitive or superior performance compared to state-of-the-art methods, particularly in challenging scenarios where appearance-based methods typically fail. A recent survey\cite{xu_local_2024} highlighted ongoing challenges in cross-modal feature matching, a domain where traditional appearance-based methods struggle. By showing that temporal dynamics alone can enable robust feature matching, we open new possibilities for video understanding across different sensing modalities and extreme viewpoint changes. The newly collected multi-view dataset consisting of challenging angles between views used in the experiments reported in this paper can be used for training and benchmarking in research in this area.

\begin{figure*}[t]
    \centering
    \includegraphics[width=0.95\textwidth]{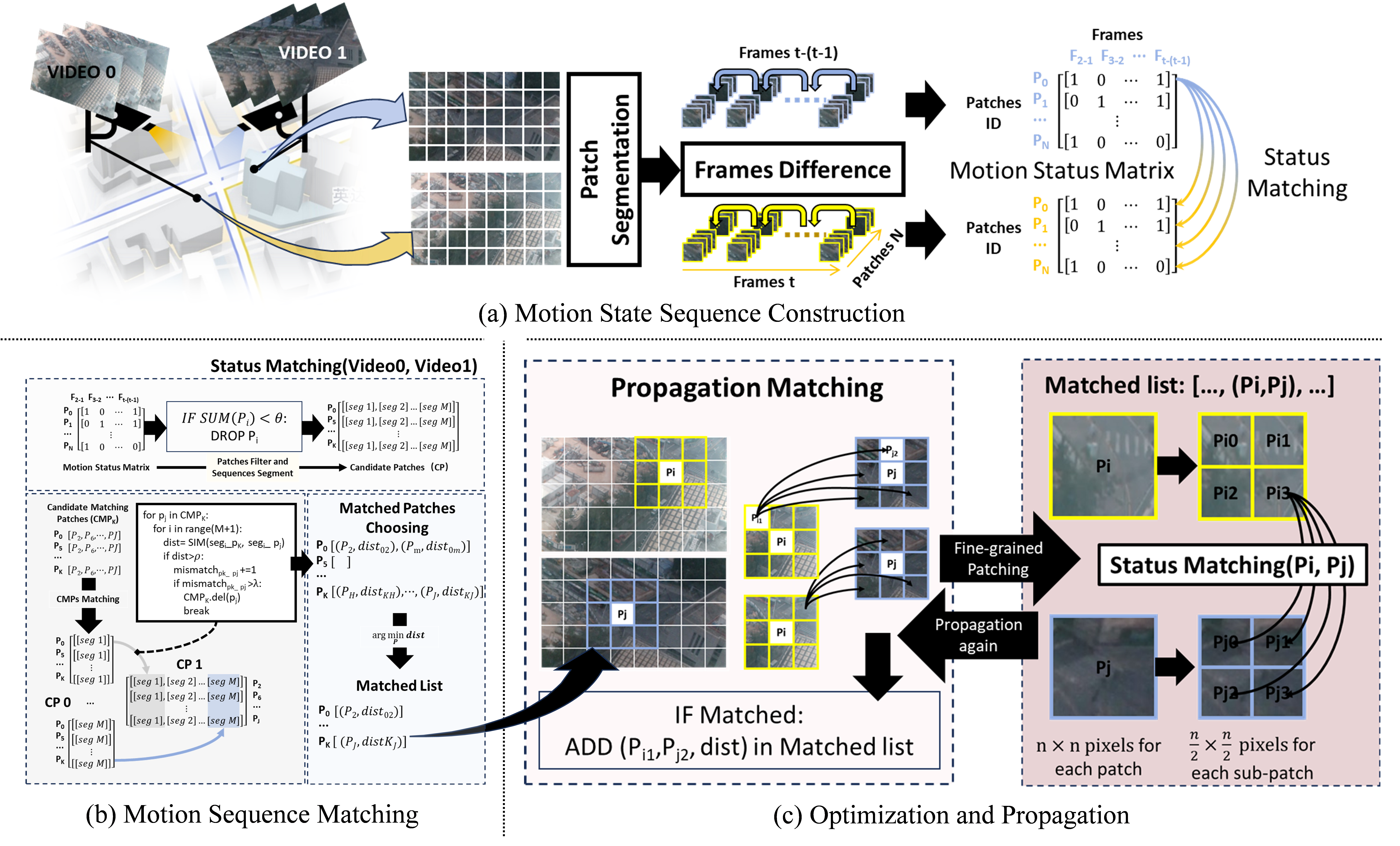} 
    \caption{Overview of the Flow Intelligence architecture. Matching is performed from coarse to fine scales by iteratively comparing temporal motion signatures to refine correspondences. \textbf{State Sequence Construction Module} extracts temporal motion patterns from each pixel block.  \textbf{Correlation Computation Module} matches blocks by evaluating correlations between their state sequences. \textbf{Optimization and Propagation Module} refines and expands these initial matches.
    }
    \Description{Detailed}
    \label{fig:fig1}
    \vspace{-3mm}
\end{figure*}

\section{Related Work}

\subsection{RGB Image Matching}

The predominant solutions focus on detecting and matching distinct spatial features across individual frames. Traditional approaches \cite{lowe_distinctive_2004, noauthor_orb_nodate, bewley_simple_2016, yang2025multimodal} and state-of-the-art deep models \cite{sun2021loftr, chen_aspanformer_2022, tuzcuouglu2024xoftr, ViNet-Li-2024, ren2025minima} alike still only deal with image features, and tend to fail when confronted with dramatic geometric transformations, image quality deteriorates, or modality shifts (e.g., RGB to infrared), leading to brittleness when scenes are noisy, cameras are misaligned, or imaging conditions vary. They fail to exploit rich temporal signals present in videos, which can be a crucial and effective constraint when appearances differ significantly. Some works have incorporated temporal patterns, (SORT\cite{bewley_simple_2016}, DeepSORT\cite{wojke_simple_2017}), which combines deep learning-based appearance features with Kalman filtering for robust object tracking, or LAMV \cite{baraldi_lamv_2018} which matches videos using kernelized temporal layers, the complexity of these models significantly increase computational demands. 

\subsection{Cross-Modality Matching}

Solutions tailored to multi-modal matching have emerged, focusing on finding modality-invariant features descriptors for matching across different sensor data. \cite{CHEN2019-visible-infrared} match edges that maintain invariant orientation and scale across visible and infrared images using Log-Gabor filters. FILER\cite{yang2025multimodal} explores frequency-domain information that remains invariant across modalities. RIFT \cite{LI-RIFT-2020} uses phrase congruence and gradient information for cross-modal matching.
Deep learning approaches mirror the core idea behind handcrafted descriptors by learning modality-invariant representations. For instance, XoFTR\cite{tuzcuouglu2024xoftr} improved upon LoFTR \cite{sun2021loftr} by training on pseudo-thermal augmented images from visible light images. VINet\cite{ViNet-Li-2024} uses an Inception-V3 backbone with attention to capture modality invariant features. These attempts show that learning how features transform from one modality to another can be effective for matching. MINIMA \cite{ren2025minima} furthers this idea by tackling the data-scarcity problem for cross-modal images generating original RGB images pairs into multiple others modalities (infrared, depth, sketch, etc), building a synthetic dataset of multi-modal image pairs. However, existing detectors, both handcrafted and learned, predominantly rely on spatial appearance, rendering them vulnerable to noise introduced by geometric transformations, camera artifacts, and varying viewpoints. We propose using temporal sequences that remain completely invariant to such limitations. Although \cite{ren2025minima} demonstrated that deep learning can achieve multi-modal generalization with sufficient training data, the substantial computational costs required for training remain a significant drawback. In contrast, Flow Intelligence requires no training, eliminating all training-related computational overhead, while maintaining performance regardless of dataset availability.

\section{METHODOLOGY}


As shown in Fig.~\ref{fig:fig1}, Flow Intelligence consists of three main parts: 1) \textbf{Motion State Sequence Construction Module}; 2) \textbf{Motion Sequence Matching}; 3) \textbf{Optimization and Propagation Module}. 

\subsection{Motion State Sequence Construction}

In contrast to previous methods in image/video matching, where visual information like pixel RGB values may fail under certain conditions (e.g., low light or large viewing angles), we propose an innovative motion-signature matching approach. Common motion detection algorithms, such as optical flow and frame difference methods, calculates precise pixel correspondence but remains highly computationally extensive. Flow Intelligence identifies motion as moving/not-moving signatures of blocks over time, at considerably lower complexity that can be performed at the “edge”, e.g. on cameras. 

We compare two or more consecutive frames to identify motion regions. For two frames, \( I_t(x, y) \) and \( I_{t-1}(x, y) \), motion detection result \( M_t(x, y) \)is computed as follows:
\begin{equation}
M_t(x, y) = 
\begin{cases}
1, & \text{if } |I_t(x, y) - I_{t-1}(x, y)| > T_1 \\
0, & \text{otherwise}
\end{cases}
\end{equation}
Here, \( T_1 \) represents the threshold. The motion count within a patch is given by $\Phi(P_{i,t}) = \sum_{(x,y) \in P_i} M_t(x, y)$, 
the motion state of Patch \( i \) at frame \( t \) $s_{P_{i,t}}$ set to 1 if $\Phi(P_{i,t}) > T_2$ or 0 otherwise. 
The motion state sequence of Patch \( i \) is $S_{P_{i}} = [s_{P_{i,1}}, s_{P_{i,2}}, \cdots, s_{P_{i,n}}]$.

The size of the patches impacts both matching precision and computational cost. Larger patches yield coarser matches with lower computational cost, and vice versa. To optimize the trade-off between computational cost and accuracy, we propose an iterative matching scheme that starts with large patches and progressively divides them into smaller patches. In each matching round, a large patch will be recursively subdivided into \(n_s^2\) smaller patches, where $n_s \ge 2$. Under a limited partitioning iterations, the origin image will be divided into \(N \times N\) small patches.

Let \(x_t\) represent the number of matches required in the \(t\)-th round, and \(S_t\) denote the accumulated number of matches up to the \(t\)-th round, resulting in $S_t = \sum_{\tau=1}^{t} x_{\tau}$. The image begins as a \(1 \times 1\) patch. In the following rounds, each patch from the previous iteration partitions into \(n_s \times n_s\) smaller subpatches. During the \(t\)-th round, the image consists of \(n_s^{t-1} \times n_s^{t-1}\) patches and only patches with \(n_s^4\) matches continue to be undergo partitioning. The number of matches required in the \(t\)-th round is calculated as $x_t = n_s^4 \cdot n_s^{2t-2} = n_s^{2t+2}$, and the total number of matches after \(t\) rounds is $S_t = \sum_{\tau=1}^t x_{\tau} = \frac{n_s^4}{n_s^2 - 1} \cdot (n_s^{2t} - 1)$.
Suppose the original image divides into \(N \times N\) small patches after \(T\) rounds, which can be expressed as \( n_s^T = N \). Replacing \(t\) with N by \( T = \log_{n_s} N \) yields the total number of matches required:
\begin{equation}
\label{eq:optimal-quadtree}
S_T = \frac{n_s^4}{n_s^2-1} \cdot (N^2 - 1).
\end{equation}

For various values of \(n_s\), the image is all to be partitioned into \(N\times N\) subpatches. So \(N\) is a constant and \(S_T\) is a function of \(n_s\). Since the derivative of \(S_T\) is increases for any \(n_s \ge 2\), \(S_T\) strictly increases with \(n_s\). Therefore, the optimal value of \(n_s\) is \(n_s = 2\), which minimizes the number of required matches. Instead of separately calculating the state sequence for each patch size using frame difference, we derive the state of a large patch as the sum of the states of its small patches:
\begin{equation}
s_{P_{i,t}} = 
\begin{cases}
1, & \text{if } \sum_{p_i \in P} \Phi(p_{i,t}) > T_3 \\
0, & \text{otherwise}
\end{cases}
\end{equation}
This approach ensures that when the number of motion pixels in all small patches of a large patch exceeds a threshold \(T_3\), the large patch is considered to have undergone "sufficient" motion.

\subsection{Motion Sequence Matching}

Based on the spatiotemporal consistency of motion in overlapping regions, we complete region correspondence by measuring sequence similarity. For the constructed binary motion state sequences (\(0/1\)), we design a logical matching algorithm to efficiently compute this similarity. Let \( \text{seq}_1 \) and \( \text{seq}_2 \) denote the two sequences to be matched:
\begin{equation}
\text{distance}(\text{seq}_1, \text{seq}_2) = 1 - \frac{2 \sum (\text{seq}_1 \& \text{seq}_2)}{ \sum \text{seq}_1 + \sum \text{seq}_2}
\end{equation}
Here, \( \& \) denotes the bitwise AND operation, and summations count the number of 1s in each sequence. This distance metrics measures the number of frames in which the two patches exhibit synchronized motion.

We note that in scenes with insufficient motion (e.g., a street at night) or overly complex content, visual noise may affect short-term matching performance, but the accuracy improves as sequence length increases.

Although the state consistency between two corresponding patches includes both simultaneous motion and simultaneous rest, the focus here is on determining correspondence based on motion. As there are always a large number of stationary regions (state sequences full of zeros), if we were to consider both motion and stationary states, errors in the constructed state sequences (e.g., some motion frames incorrectly marked as 0 or some 0 frames incorrectly marked as 1) could cause sequences with little motion to mistakenly match with the all-0 sequence, especially if the sequence length is long. The denominator is the sum of the number of 1s in both sequences, ensuring that sequences with less motion do not match with sequences that have frequent motion.

To improve calculation efficiency when matching two sequences that are clearly not corresponding, only a small segment of the sequence may be needed to determine that they do not match. To address this, we divide the sequence into multiple segments and calculate the similarity for each segment. During the matching process, we record the number of times the distance exceeds a threshold. When the threshold is exceeded \(n\) times (in the experiment, \(n = 1\)), we discard the entire sequence, thus avoiding similarity calculations for the entire long sequence. 

Furthermore, to reduce unnecessary computation, we exclude low-motion patches before matching. The remaining patches form a candidate set for each reference patch, restricting comparisons to meaningful regions and reduces computational load. 

Given the viewpoint discrepancy between cameras, not all motion regions have overlapping counterparts, and the correspondence is not strictly one-to-one. As illustrated in Fig.~\ref{fig:fig1}, we avoid naive nearest-patch assignments. Instead, each match is represented as a triplet \( (P_{1,i}, P_{2,j}, \text{Dist}) \), where \( \text{Dist} \) is the computed similarity. Final matches are selected by sorting the triplets and retaining those with distances below a threshold, allowing for one-to-many mappings while effectively filtering incorrect matches.

\subsection{Optimization and Propagation}


To address the one-to-many correspondence between image patches arising from viewpoint changes, we introduce a stride parameter during image partitioning. When the stride equals the block size, adjacent blocks are non-overlapping. By adjusting the stride, we effectively control the spatial granularity of the search space, enabling more precise matching and facilitating the identification of accurate correspondences.


To reduce the complexity of querying similar blocks in a large space, we recursively build a quadtree image structure for the video frames, replacing the global query on the entire image with local searches within the same level and branch of the image tree that proceeds as follows. Suppose that after \(i\) iterations, a block pair \(p_A^{(i)} \in A\) and \(p_B^{(i)} \in B\) have been successfully matched. Then, in the \((i+1)\)-th iteration, we first divide the block \(p_A^{(i)}\) and \(p_B^{(i)}\) into smaller blocks \(p_A^{(i+1)} \in p_A^{(i)}\) and \(p_B^{(i+1)} \in p_B^{(i)}\), and then, for all \(p_A^{(i+1)}\), we search for the most similar block in \(p_B^{(i)}\) and match them, instead of searching globally in \(p_B\).

Hierarchical optimization of the image according to a quadtree grants the patch matching a multi-scale search capability. At the top of the image tree, the large-scale patch search is a spatially localized fuzzy match, which provides a rough direction for the lower-level search, significantly reducing the spatial redundancy in matching. At the bottom of the tree, small-scale patches perform a detailed local search, which improves the algorithm's ability to handle image details and enhances the accuracy of patch matching. This multi-scale search approach reduces the spatiotemporal complexity while minimizing the smallest processing unit that can be handled, balancing both matching speed and accuracy. Eq.\ref{eq:optimal-quadtree} proves that quadtree structure is optimal.


Due to strict hierarchical matching and similarity filtering, the number of blocks that can be successfully matched is not large. Furthermore, due to the locality of tree-based search, we cannot guarantee that the successfully matched blocks are globally optimal. To ensure the accuracy of the matching results, we introduce propagation optimization, which extends the algorithm's global search capability and robustness based on existing matches. Unlike \cite{Barnes2009PatchMatchAR}, we first use random search to improve the algorithm's ability to escape from local optima and find the global optimum. Random search is an iterative process. For a given well-matched patch pair $(P_1, P_2)$, we introduce random offsets around $P_2$ to explore and optimize the best match for $P_1$. The random offsets are generated according to an exponentially decaying distance relative to the current good match.

After the random search, the propagation process utilizes the natural structure of the image and the known best matches to rapidly propagate these matches to surrounding regions. The spatial continuity and consistency of image content means that when we have already found a best-matched patch in one image and the surrounding region of this patch in the other image has corresponding overlapping parts, the matching region of these adjacent blocks in the other image will likely be located near the initially found best-matched patch. Thus, the propagation algorithm can expand the matching results to the neighboring regions to find more matches.

\section{Experiment}

\subsection{Experiment Preparation}


The proposed algorithm is designed for processing multi-view synchronized videos and is specifically optimized for challenging conditions, including low-light environments, high noise levels, and multispectral scenarios. To evaluate its performance in cross-modal settings, we conducted experiments on two datasets: OTCBVS\cite{DAVIS2007162}, which captures a busy road intersection on the Ohio State University campus, and LLVIP\cite{jia2021llvip}, a high-quality visible-infrared paired dataset tailored for low-light vision tasks. From LLVIP, we specifically utilized the video data.

The OTCBVS dataset, however, has limited content, and its visible and infrared frames exhibit minimal viewpoint variation due to near-perfect alignment. To introduce diversity, we manually applied multiple perspective transformations to each video pair to generate an augmented dataset, OTCBVS-Aug. Additionally, due to a lack of synchronized multi-view video datasets under low-light and high-noise conditions, we collected 15 pairs of urban street videos featuring significant viewpoint variations and low-light conditions. This newly collected dataset is referred to as CityData-MV (MultiView).

Given the extreme difficulty of manually annotating overlapping matching regions in our captured data, we developed an alternative approach for quantitative evaluation. For each video in CityData-MV, we apply two distinct perspective transformations to create a corresponding video pair. Since these transformation relationships were manually defined, we could accurately compute patch correspondences within each pair. This augmented dataset is denoted as CityData-Aug. A summary of these datasets is provided in the table \ref{tab:datasets}, with additional samples available in the Appendix B.

\begin{table}[htbp]
    \vspace{-3mm}
    \centering
    \small  
    \setlength{\tabcolsep}{2.5mm}  
    \caption{Dataset Specifications Overview (IR: Infrared)}
    \label{tab:datasets}
    \begin{tabular}{@{}l@{\hspace{2.5mm}}c@{\hspace{2.5mm}}c@{\hspace{2.5mm}}c@{\hspace{2.5mm}}c@{\hspace{2.5mm}}c@{}}
    \toprule
    \textbf{Dataset} & \textbf{FPS} & \textbf{Resolution} & \textbf{Duration} & \textbf{Modality} & \textbf{Pairs} \\
    \midrule
    CityData-Aug & 30 & 1920$\times$1080 & 2$\sim$3min & RGB-RGB & 32 \\
    \addlinespace[3pt]
    CityData-MV & 30 & 1920$\times$1080 & 2$\sim$3min & RGB-RGB & 28 \\
    \addlinespace[3pt]
    OTCBVS-Aug & 30 & 320$\times$240 & 30s & RGB-IR & 36 \\
    \addlinespace[3pt]
    LLVIP & 25 & \begin{tabular}[c]{@{}c@{}}1920$\times$1080(RGB)\\1280$\times$720(IR)\end{tabular} & 2$\sim$11min & RGB-IR & 15 \\
    \bottomrule
    \end{tabular}
    \vspace{-2mm}
\end{table}


We adopted a three-frame difference method, where the parameter \( T_1 \) is set to 4, and \( T_2 \) and \( T_3 \) are set to 1/6 of the patch pixel count, respectively. During the construction of the candidate set, patches with motion elements less than 1/30 of the total sequence length are excluded, while the length of a single segment (seg) in the complete sequence is set to 500. During the matching process, the distance threshold between two sequences is set to the first quartile (1/6 quantile, the distance factor $\lambda$ is configured as 6 in the implementation) of all distances during the matching of the first segment, and the maximum number of unmatched segments is limited to 1. For the videos in OTCBVS and LLVIP, we process all frames in the video for matching. In contrast, for the CityData videos, we limit the processing to the first 3000 frames for matching. In the following experiments, unless otherwise specified, we set the number of hierarchical matching iterations to 4, where the patch size is progressively reduced from $(64, 64)$ to $(8, 8)$. After each matching stage, three propagation operations are performed.

All experiments were performed on a high-performance workstation with a 13th Gen Intel\textsuperscript{\textregistered} Core\texttrademark{} i9-13900KF CPU and an NVIDIA GeForce RTX 4090 GPU.

\subsection{Baselines}

We conduct a comprehensive evaluation of our method against both traditional feature-based techniques (SIFT, ORB, AKAZE) and state-of-the-art deep learning approaches--\textbf{MINIMA}~\cite{ren2025minima}, which is pre-training on RGB data and fine-tuned on synthetic cross-modal data, achiving strong generalization in both in-domain and zero-shot settings, outperforming modality-specific methods. Our experiments adopt the official implementations of \textbf{MINIMA$_{LG}$}~\cite{ren2025minima, lindenberger2023lightglue} and \textbf{MINIMA$_{LoFTR}$}~\cite{ren2025minima, sun2021loftr}, which include RANSAC-based geometric verification as a post-processing step. As all baselines are designed for static images, we uniformly sample one frame every 300 frames from each video in our datasets for fair comparison.

\subsection{Results}
\subsubsection{Quantitative Matching Analysis}

\begin{table}[htbp]
    \vspace{-2mm}
    \centering
    \caption{Comparison of Feature Matching Methods on CityData-Aug}
    \label{tab:feature_comparison}
    \begin{tabular}{l@{\hspace{2.5mm}}c@{\hspace{2.5mm}}c@{\hspace{2.5mm}}c@{\hspace{2.5mm}}c@{\hspace{2.5mm}}c@{\hspace{2.5mm}}c}
    \toprule
    \textbf{Metric} & SIFT & AKAZE & ORB & M\_LG & M\_LoFTR & Ours \\
    \midrule
    \textbf{Matches} & 1232 & 2309 & 714 & 509 & 1477 & 2924 \\
    \textbf{Dist (px)} & 33.73 & 31.08 & 131.47 & 1.79 & 1.46 & 2.59 \\
    \bottomrule
    \end{tabular}

    \vspace{1mm}
    \raggedright
    \footnotesize
    \textbf{Note:} \textbf{Matches}: Average matches per frame (per scene for ours); \textbf{Dist (px)}: Average pixel distance.
    
    \vspace{-3mm}
\end{table}

\begin{figure}[htbp]
    \vspace{-4mm}
    \centering
    \subfloat[Accuracy comparison under different error thresholds on CityData-Aug]{\includegraphics[width=0.46\textwidth]{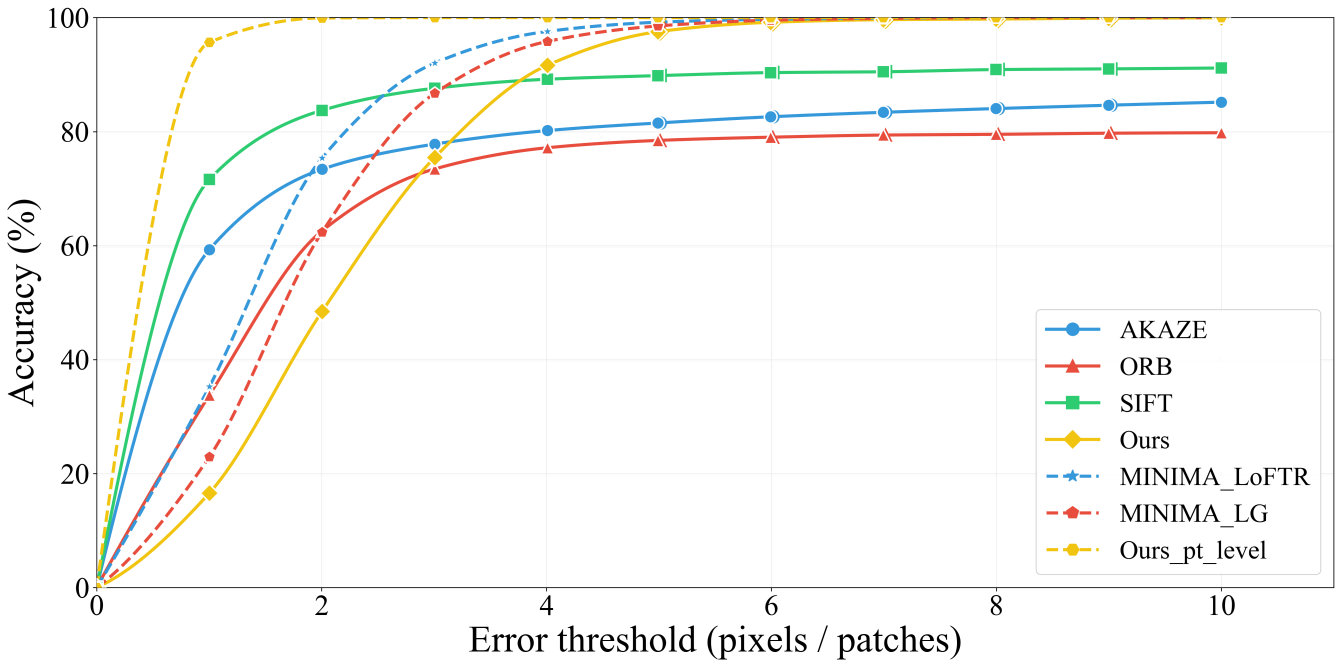}}
    \hfill
    \subfloat[Accuracy comparison under different error thresholds on OTCBVS-Aug]{\includegraphics[width=0.46\textwidth]{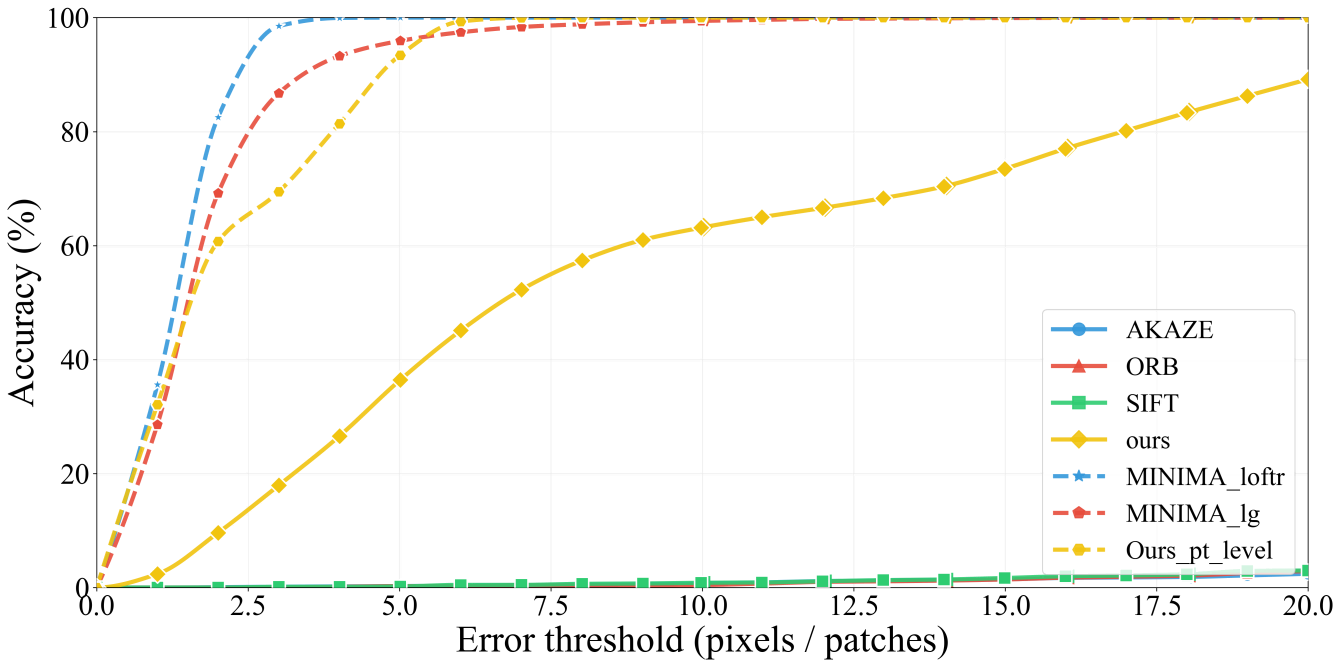}}
    \caption{Accuracy curves of different feature matching methods, showing performance variations with increasing error threshold.(a) Evaluated on CityData-Aug; (b) Evaluated on OTCBVS-Aug. Note that due to the lack of ground truth labels, the accuracy on OTCBVS-Aug dataset is computed using $MINIMA_{LoFTR}$ as pseudo labels.}
    \label{fig:performance_analysis}
    \vspace{-2mm}
\end{figure}

The quantitative results on CityData-Aug are shown in Table~\ref{tab:feature_comparison} and Fig.~\ref{fig:performance_analysis}(a). Our method achieves significantly higher matching quantities, averaging 2,924 matches per scene—outperforming traditional approaches such as AKAZE (2,309) and SIFT (1,232), and nearly doubling the semi-dense \texttt{MINIMA\textsubscript{LoFTR}} (1,477). Despite operating at the patch rather than pixel level, our approach achieves high matching density.

This superior performance can be attributed to two key factors. First, our patch-based matching strategy effectively captures local structural information and exhibits strong consistency under viewpoint transformations. Second—and more importantly—the number of correspondences produced by our method is primarily governed by the extent of motion, rather than the distribution of pixel-level features within the frames. This characteristic enables our approach to establish reliable correspondences even in regions with limited visual texture or repetitive patterns—conditions under which traditional feature-based methods typically deteriorate. As long as there is sufficient motion between frames, our method is capable of generating a dense set of patch correspondences, regardless of the visual distinctiveness of local features.

In terms of matching accuracy, our approach achieves an average distance error of 2.59 pixels, positioning between traditional methods (SIFT: 33.73, AKAZE: 31.08, ORB: 131.47 pixels) and learning-based methods (MINIMA\_LoFTR: 1.46, MINIMA\_LG: 1.79 pixels). This performance can be interpreted from two perspectives. First, operating on $8\times 8$ pixel patches rather than individual pixels inevitably introduces quantization errors when computing distances between patch centers. Second, the presence of perspective variations between frames hinders precise one-to-one correspondence between rectangular patch cells, as such non-affine transformations which prevents exact spatial alignment across viewpoints.

To better analyze the effectiveness of our patch-level matching strategy, we further introduce a patch-level error threshold, which quantifies the distance between predicted and ground-truth patches in units of patches rather than pixels. Specifically, a prediction is considered correct under a patch-level threshold $T$ if the center of the predicted patch lies within a square region of side length $T \cdot P$ centered at the ground-truth patch, where $P$ denotes the patch size. The patch-level distance is computed as $d_{\text{patch}} = \left\lceil \frac{2 \cdot \max(|x_1 - x_2|, |y_1 - y_2|)}{P} \right\rceil$ (see Appendix A for details).

As shown in Fig.~\ref{fig:performance_analysis}, under this patch-level metric (Ours\_pt\_level), our method achieves 95.6\% accuracy at a 1-patch threshold and reaches 100\% accuracy at a 2-patch threshold. These results indicate that, even when accounting for the inherent spatial granularity of patch-wise representations, our model consistently predicts correspondences within a minimal spatial deviation. The rapid accuracy saturation suggests strong local consistency and precise regional localization of our method. 

It should be noted that these results were obtained on datasets with controlled perspective transformations. The performance on more challenging CityData-MV dataset, which contains more complex real-world variations, are detailed in Section~\ref{sec:Qualitative Matching Results}.

For the OTCBVS-Aug dataset, where ground truth correspondences are unavailable, we observed that MINIMA methods yielded high matching quality in our experiments (detailed in Appendix C). Accordingly, we used the pixel correspondences from \texttt{MINIMA\textsubscript{LoFTR}} to estimate the perspective transformation matrix between image pairs. This matrix was then used to generate pseudo-labels by mapping corresponding pixels across viewpoints, enabling quantitative evaluation of our method and traditional feature-based approaches.Due to this label generation strategy, the MINIMA methods (LoFTR and LG variants) unsurprisingly achieve near-perfect accuracy—approaching 100\% within a 2.5-pixel error threshold. In contrast, traditional feature-based methods perform poorly on this RGB-IR dataset, with accuracy remaining below 5\% across all thresholds. Our method achieves moderate performance, with accuracy increasing progressively as the error tolerance grows. \textbf{However, it should be noted that the reported accuracy of our method might be conservative due to the inherent errors propagated from the pseudo-labels themselves.}

\begin{figure}[htbp]
    \vspace{-6mm}
    \centering
    \subfloat[On CityData-Aug]{\includegraphics[width=0.495\columnwidth]{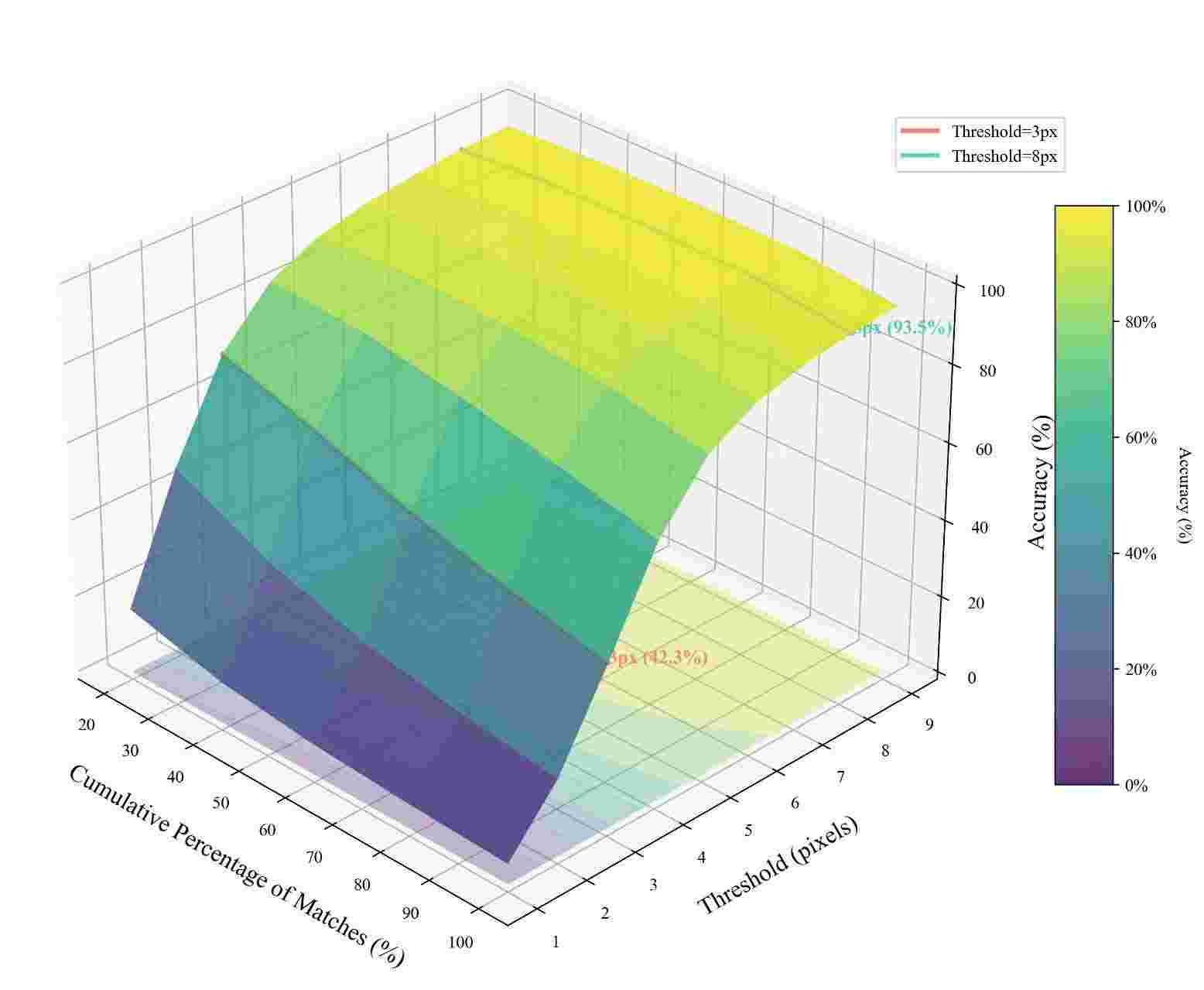}}
    \hfill
    \subfloat[On OTCBVS-Aug]{\includegraphics[width=0.495\columnwidth]{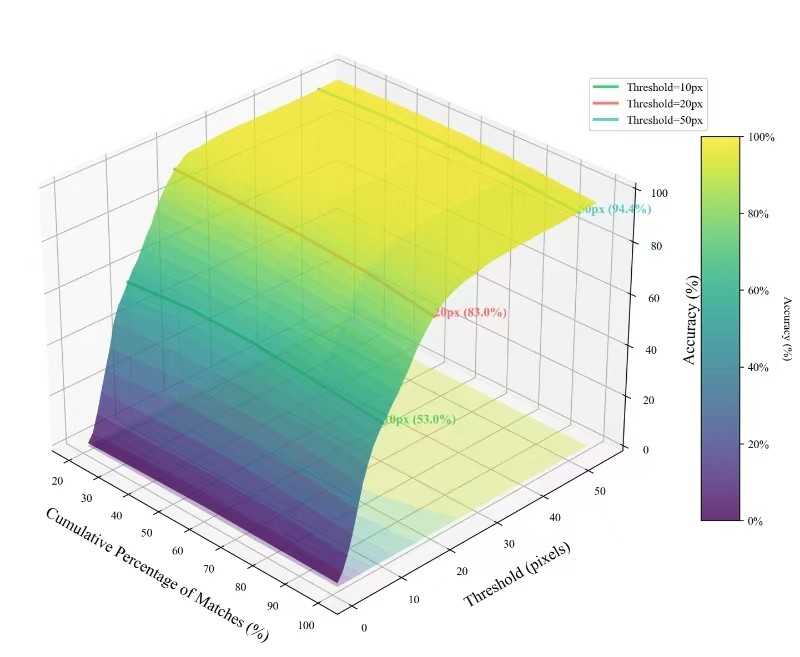}}
    \caption{Distribution of matching accuracy, error threshold, and cumulative percentage of confidence-ranked matches for Flow Intelligence.}
    \label{fig:3d_analysis}
    \vspace{-6mm}
\end{figure}

Fig.~\ref{fig:3d_analysis}  shows a 3D visualization of matching accuracy distribution on CityData-Aug and OTCBVS-Aug datasets. The surfaces represent the relationship between matching accuracy, error threshold, and cumulative percentage of confidence-ranked matches.

The smooth gradient of the surface indicates that our confidence ranking effectively correlates with matching accuracy. Specifically, when considering only the top-ranked matches (20-30\% cumulative percentage), the accuracy improves compared to using all matches, suggesting the reliability of our confidence estimation mechanism.

\subsubsection{Qualitative Matching Results}
\label{sec:Qualitative Matching Results}

\begin{figure*}[htbp]
    \centering
    \begin{tabular}{c@{\hspace{0.5mm}}c@{\hspace{0.5mm}}c@{\hspace{0.5mm}}c}
        & SIFT & MINIMA\_LoFTR & Ours \\
        \raisebox{0pt}{\rotatebox{90}{Scene 1}} &
        \includegraphics[width=0.33\textwidth]{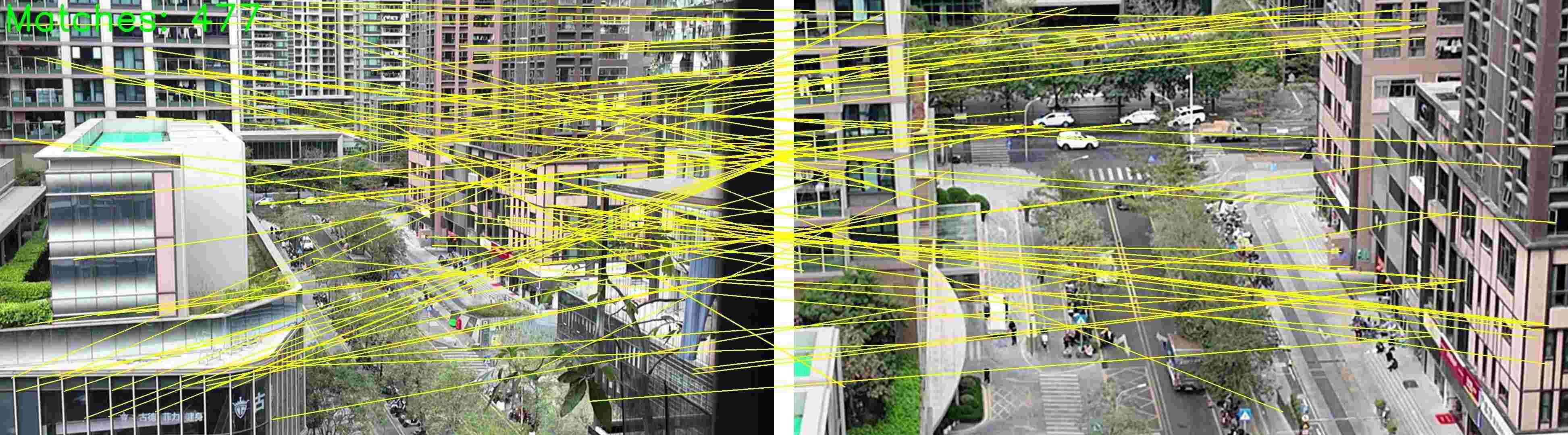} &
        \includegraphics[width=0.33\textwidth]{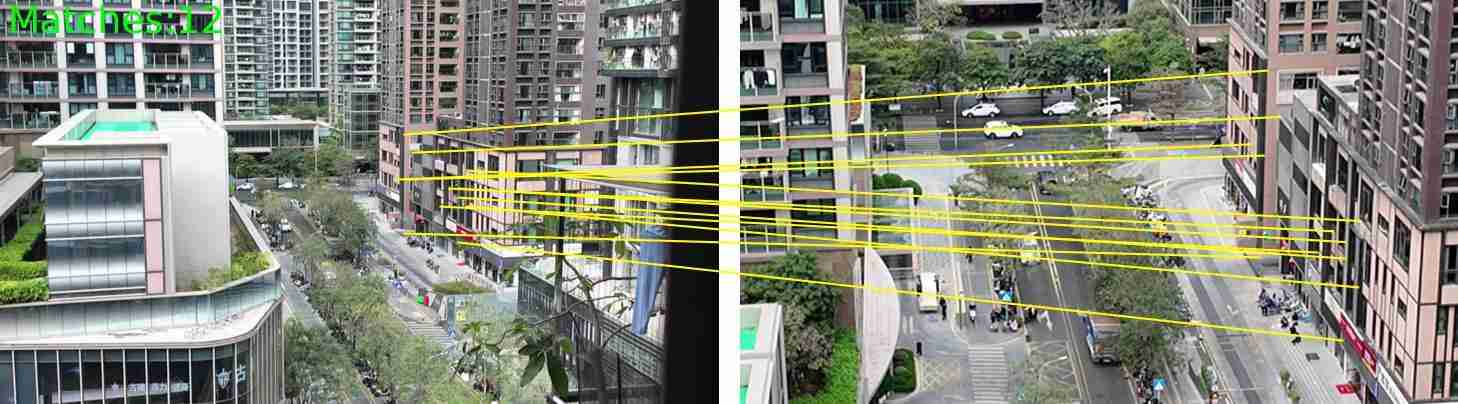} &
        \includegraphics[width=0.33\textwidth]{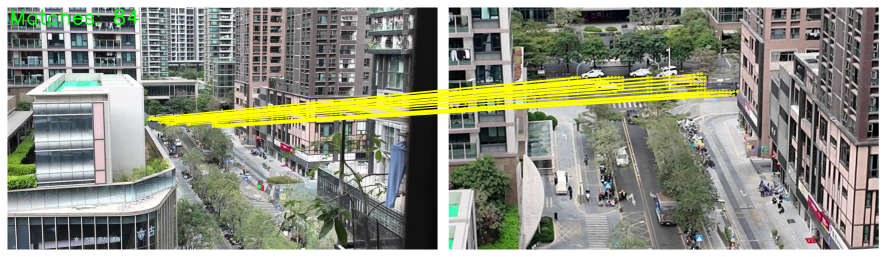} \\[1mm]

        \raisebox{0pt}{\rotatebox{90}{Scene 2}} &
        \includegraphics[width=0.33\textwidth]{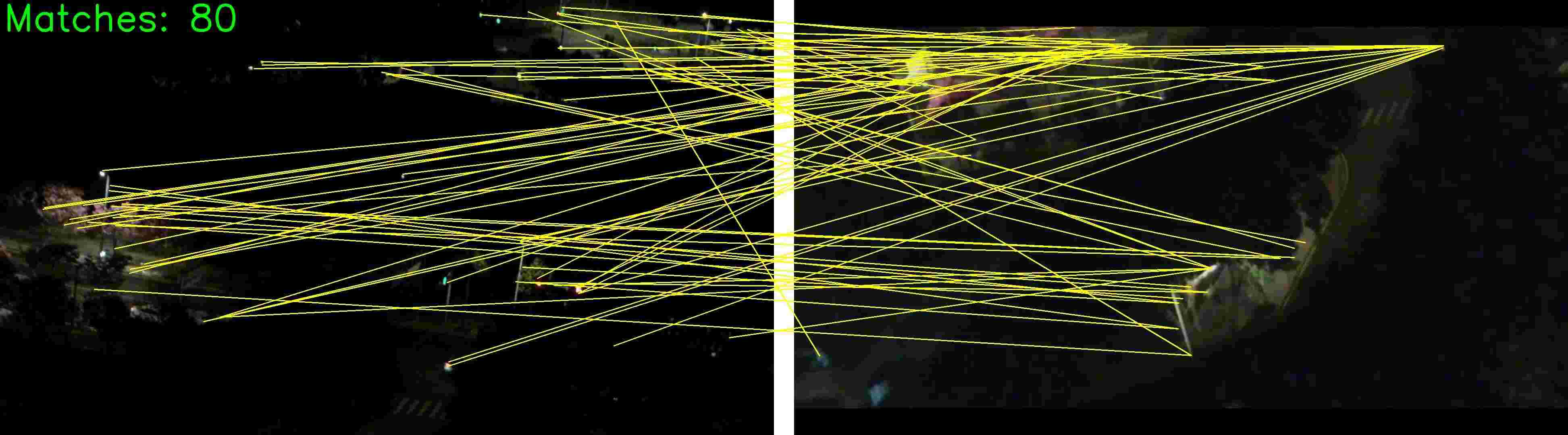} &
        \includegraphics[width=0.33\textwidth]{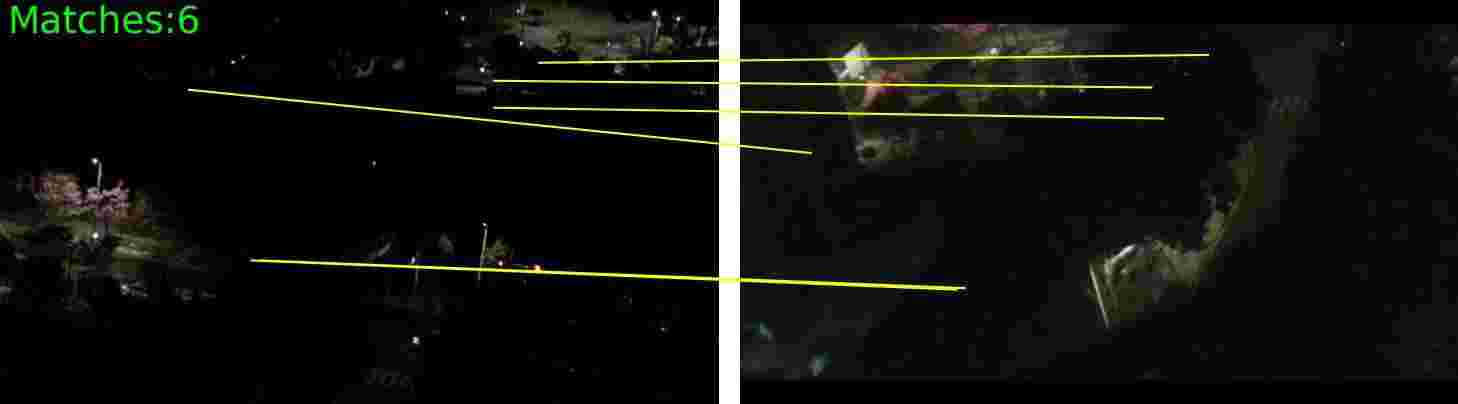} &
        \includegraphics[width=0.33\textwidth]{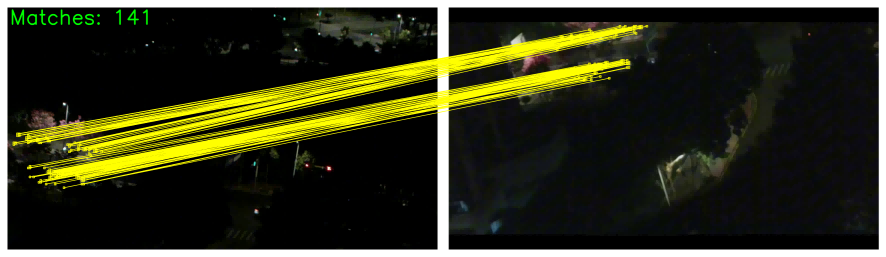} \\[1mm]
        
        \raisebox{0pt}{\rotatebox{90}{Scene 3}} &
        \includegraphics[width=0.33\textwidth]{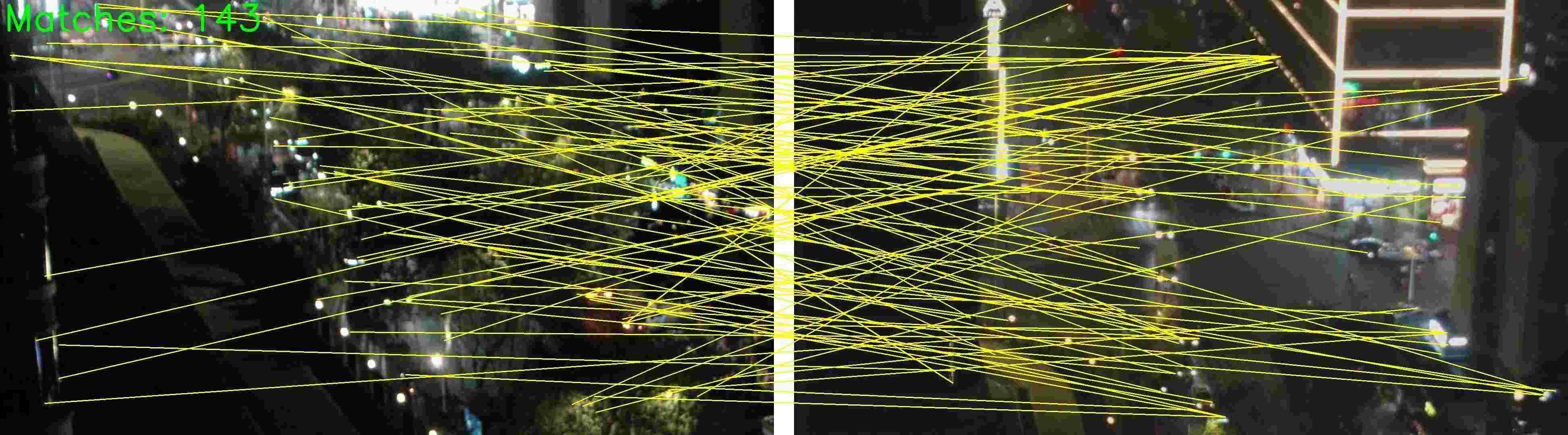} &
        \includegraphics[width=0.33\textwidth]{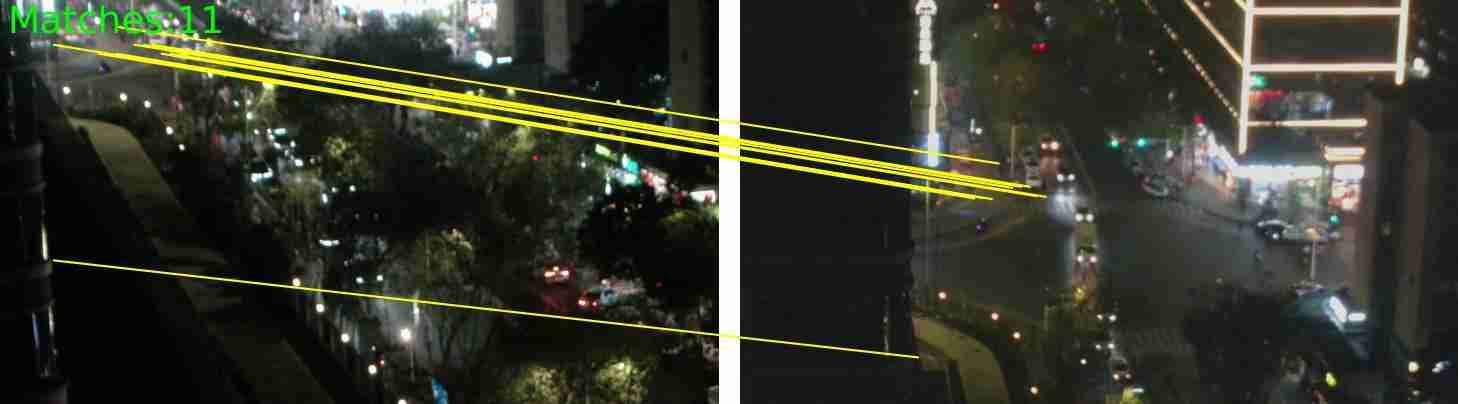} &
        \includegraphics[width=0.33\textwidth]{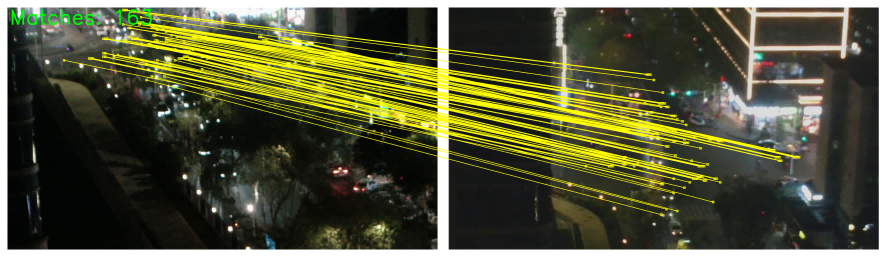}
    \end{tabular}
    
    \caption{Feature matching results on CityData-MV between SIFT, MINIMA-LoFTR and Flow Intelligence (Ours).}
    \label{fig:qualitative_comparison}
    \vspace{-2mm}
\end{figure*}

Fig.~\ref{fig:qualitative_comparison} and Fig.~\ref{fig:first_fig}(LVC) presents the matching results on the CityData-MV dataset, which is characterized by significant viewpoint variations and challenging low-light conditions. Due to the absence of ground-truth annotations, all correspondences are visualized using yellow lines, and the number of matched pairs is shown in the top-left corner of each image.


As shown in Fig.~\ref{fig:first_fig}, our method demonstrates superior performance across challenging scenarios. In the large viewpoint change (LVC) scene, where textureless regions like road surfaces dominate, MINIMA\_LoFTR only detects 4 matches, while our method successfully leverages motion cues from moving vehicles to establish 135 matched pairs with globally consistent alignment. Similar advantages are observed in extremely low-light environments (Scenes~2 and 3), where our method maintains reliable matching performance, showcasing its robustness to both geometric and photometric challenges.

 To further validate our method's matching capability across various modalities, we utilized MINIMA's Data Engine\cite{ren2025minima, scepter, depth_anything_v1, depth_anything_v2, bae2024dsine, Anime2Sketch, liu2021paint} to generate additional multi-modal datasets for testing. More detailed results can be found in Appendix D.

Additionally, we summarize the runtime characteristics of all methods (details in Appendix~D). Traditional keypoint-based approaches are computationally efficient but exhibit poor matching quality in cross-modal scenarios due to limited keypoint detectability. Learning-based methods like MINIMA offer stable runtimes (0.2–0.3s per image pair) with the aid of GPU acceleration. Our method achieves comparable or even superior efficiency, particularly on simpler datasets such as OTCBVS, where near real-time performance is achievable. Furthermore, the proposed framework supports flexible trade-offs between speed and matching density through early stopping or reduced propagation steps.



\subsection{Ablation Studies}

\subsubsection{Propagation Iterations}
\begin{figure}[htbp]
    \centering
    \includegraphics[width=0.98\columnwidth]{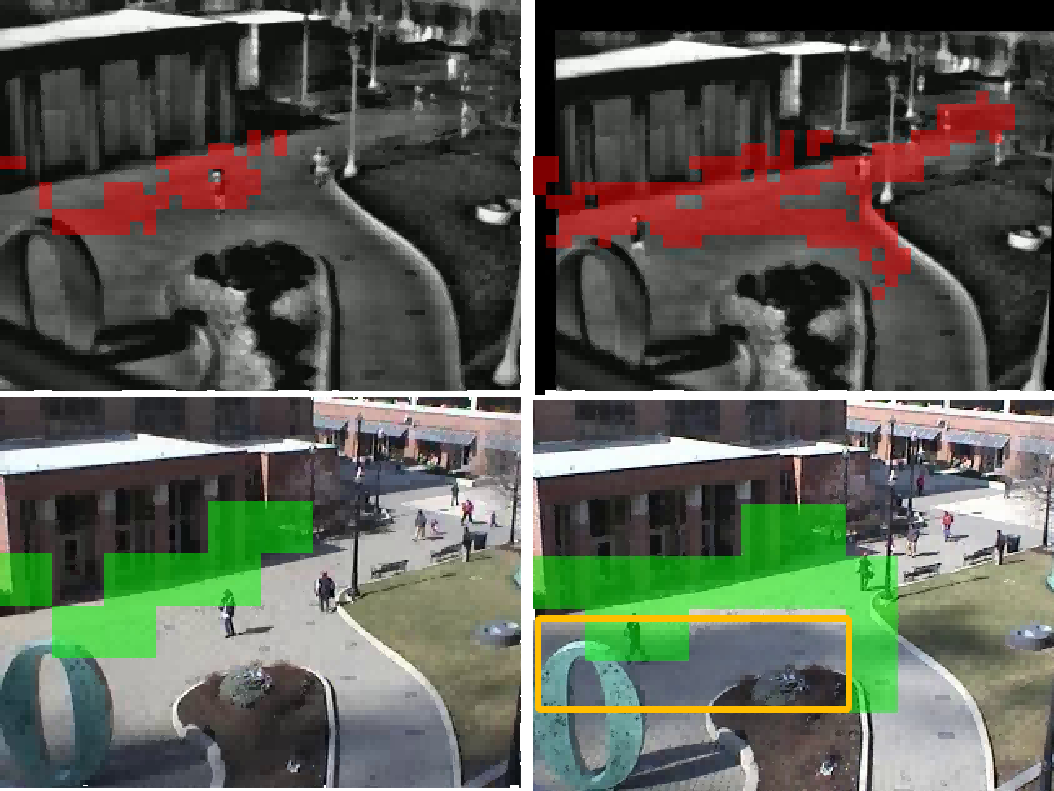} 
    \caption{{Propagation results after zero iterations (left column) and three iterations (right column). Colored regions indicate matched patches at different scales: $8 \times 8$ (top row, red) and $32 \times 32$ (bottom row, green) for the same scene.}}
    \Description{Detailed}
    \label{fig:fig5}
    \vspace{-2mm}
\end{figure}

Fig.~\ref{fig:fig5} illustrates how the matching results change with the number of propagation iterations (with frames extracted from the video as background). It can be seen that as the number of iterations increases, the matching results spread rapidly to surrounding areas. After three Propagation iterations, matching results stabilize and identifies the majority of overlapping motion regions in the video. By comparing results for different Patch sizes, we observe that during the hierarchical tree matching,  the Propagation mechanism expands smaller Patches beyond areas of larger patch matching, thus converging at global optimal solutions. 

Additionally, we emphasize that in the scenario shown in Figure \ref{fig:fig5}, the normal-light video (in the row below) is constantly affected by moving cloud shadows, which interfere with the construction of motion states (boxed in yellow). In contrast, the thermal video does not have cloud shadows, and even under these circumstances, our algorithm still manage to accurately identify overlapping moving regions in the video.

\subsubsection{Hierarchical Tree Optimization}
\vspace{-3mm}
\begin{table}[htbp]
    \centering
    \caption{Ablation study on hierarchical patch matching strategy. Direct matching uses 8$\times$8 patch directly, while hierarchical approach progressively refines from 64$\times$64 to 8$\times$8.}
    \label{tab:ablation_hierarchical}
    \begin{tabular}{l@{\hspace{2.5mm}}c@{\hspace{2.5mm}}c@{\hspace{2.5mm}}c@{\hspace{2.5mm}}c@{\hspace{2.5mm}}c@{\hspace{2.5mm}}c}
    \toprule
    \multirow{2}{*}{\textbf{Method}} & \multicolumn{2}{c}{Time (s)} & \multirow{2}{*}{Matches} & \multirow{2}{*}{Dist (px)} & \multicolumn{2}{c}{Accuracy (\%)} \\
    \cmidrule(lr){2-3} \cmidrule(l){6-7}
    & $\sum\tau_1$ & $\sum\tau_2$ & & & @1px & @5px \\
    \midrule
    \textbf{Direct}  & 2k & 0.118 & 86.6 & 3.61 & 12.1 & 75.9 \\
    \textbf{Hierarchical}  & 0.490 & 2.926 & 4879.7 & 3.59 & 5.8 & 87.0 \\
    \bottomrule
    \end{tabular}
    
    \vspace{1mm}
    \raggedright
    \footnotesize
    $\sum$ means summing across all scales; $\tau_1$: Local patch matching time; $\tau_2$: Cumulative propagation time (3 iterations). Matches: Average Matches per scene at 8$\times$8 patch;
    \vspace{-2mm}
\end{table}

To validate the effectiveness of our hierarchical matching strategy, we conducted all ablation experiments on ten randomly selected videos from the CityData-Aug dataset. As shown in Table~\ref{tab:ablation_hierarchical}, the hierarchical approach demonstrates significant advantages in both efficiency and accuracy. In terms of matching time, our hierarchical strategy dramatically reduces the patch matching time from 2000s to 0.490s. The propagation time increases from 0.118s to 2.926s as a natural consequence of processing a much larger number of matches (from 86.6 to 4879.7, approximately 56 times more matches). Regarding matching accuracy, although direct matching achieves higher precision at @1px threshold (12.1\% vs. 5.8\%) due to its exhaustive search in the complete matching space, our hierarchical approach, power by the propagation step that enables optimal match refinement beyond patch constraints, achieves superior performance at @5px threshold (87.0\% vs. 75.9\%). Notably, our approach even maintains a slightly better average distance error (3.59px vs. 3.61px), demonstrating that the hierarchical strategy with propagation can effectively balance between computational efficiency and matching accuracy.

\subsubsection{Distance Factor $\lambda$}

\begin{figure}[htbp]
    \centering
    \includegraphics[width=\columnwidth]{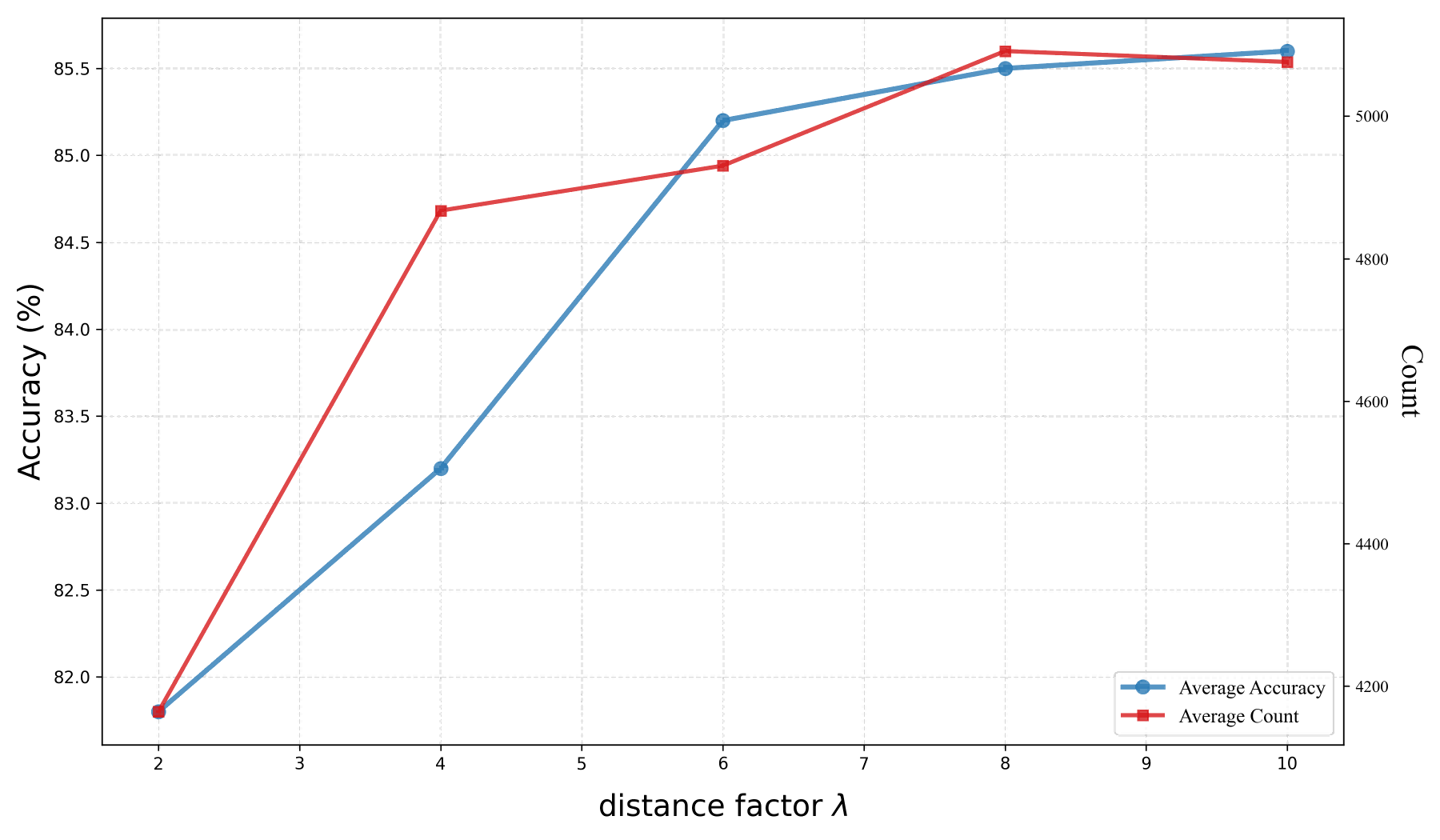} 
    \caption{Matching Accuracy and Number of Matches vs. Distance Factor $\lambda$}
    \Description{Detailed}
    \label{fig:distance_factor}
    \vspace{-5mm}
\end{figure}

The results in Fig.~\ref{fig:distance_factor} reveal that as $\lambda$ increases from 2 to 10, both the average accuracy and the number of matches exhibit an upward trend, which might appear counterintuitive but is a nature consequence of our matching strategy. The distance factor $\lambda$ defines the search radius for potential matches: increasing $\lambda$ tightens the distance threshold ($\frac{d}{\lambda}$), thereby enforcing stricter matching criteria. Rather than suppressing match quantity, this tightening unexpectedly improves both accuracy and match count because a smaller distance threshold eliminates ambiguous candidates early in the process and reduces the risk of incorrect matches occupying viable positions. This filtering effect preserves matching capacity for more reliable correspondences in later stages. Additionally, stricter thresholds yield more precise initial matches, which serve as dependable seeds for the propagation mechanism. As a result, the matching process can expand more effectively, producing a greater number of accurate matches in the final output. 


\subsubsection{Effect of Sequence Length}

\begin{table}[htbp]
    \vspace{-3mm}
    \centering
    \caption{Impact of motion state sequence length on matching performance.}
    \label{tab:sequence_length}
    \begin{tabular}{l@{\hspace{2mm}}c@{\hspace{2mm}}c@{\hspace{2mm}}c@{\hspace{2mm}}c@{\hspace{2mm}}c@{\hspace{2mm}}c}
    \toprule
    \textbf{Seq Length}  & 150 & 450 & 1000 & 1500 & 2000 \\
    \midrule
    \textbf{Matches} & 1871.2 & 2291.9 & 4249.9 & 4930.6 & 5674.1 \\
    \textbf{Dist (px)} & 9.00 & 3.87 & 3.84 & 3.70 & 4.04 \\
    \textbf{Acc@5px} (\%) & 62.0 & 81.1 & 83.2 & 85.2 & 86.1 \\
    \bottomrule
    \end{tabular}
    \vspace{-3mm}
\end{table}

To investigate the impact of video sequence length on the performance of our method, we conducted experiments using five different sequence lengths. Note that our method adopts three-frame differencing for motion detection, discarding the middle frame; thus, the actual number of required video frames is twice the sequence length.

As shown in Table~\ref{tab:sequence_length}, longer sequences consistently lead to improved matching performance. The average number of matches increases from 1871.2 to 5674.1, while accuracy at a 5-pixel threshold improves from 62.0\% to 86.1\%. In parallel, the average distance error drops sharply from 9.00 pixels at a length of 150 and stabilizes around 4 pixels for longer sequences, indicating enhanced robustness.


\section{Conclusion}

In this work, we propose a novel local feature matching algorithm that leverages temporal patterns between cubic patches (blocks of pixels over time), departing from the traditional reliance on spatial features. We demonstrates robustness to challenging conditions where traditional methods and deep learning-based approaches fail in difficult, such as geometric (scale, rotation, parallax), lighting variations, and cross modals (RGB, infrared, depth, etc.). Additionally, our approach requires no pre-training and is highly computation efficient. We believe the importance of using temporal coherence in video sequences—an inherently temporal data format—opens new doors to video feature matching and understanding. 




\bibliographystyle{ACM-Reference-Format}
\bibliography{sample-base}

\appendix 
\section{Definition of Patch-Level Error Threshold}
\label{Patch_level_th}

To better evaluate the robustness and accuracy of patch-based matching strategies, we introduce a \textit{patch-level error threshold} metric, which is more aligned with the discrete nature of patch-based matching.

Assume the patch size is $P \times P$, and let $p_1$ denote the predicted patch and $p_2$ the ground-truth patch, with center pixel coordinates $\mathbf{c}_1 = (x_1, y_1)$ and $\mathbf{c}_2 = (x_2, y_2)$, respectively. The patch-level distance $d_{\text{patch}}$ between them is computed as:
\begin{equation}
d_{\text{patch}}(p_1, p_2) = \left\lceil \frac{2 \cdot \max(|x_1 - x_2|, |y_1 - y_2|)}{P} \right\rceil
\end{equation}

\begin{figure}[h]
    \centering
    \includegraphics[width=0.9\linewidth]{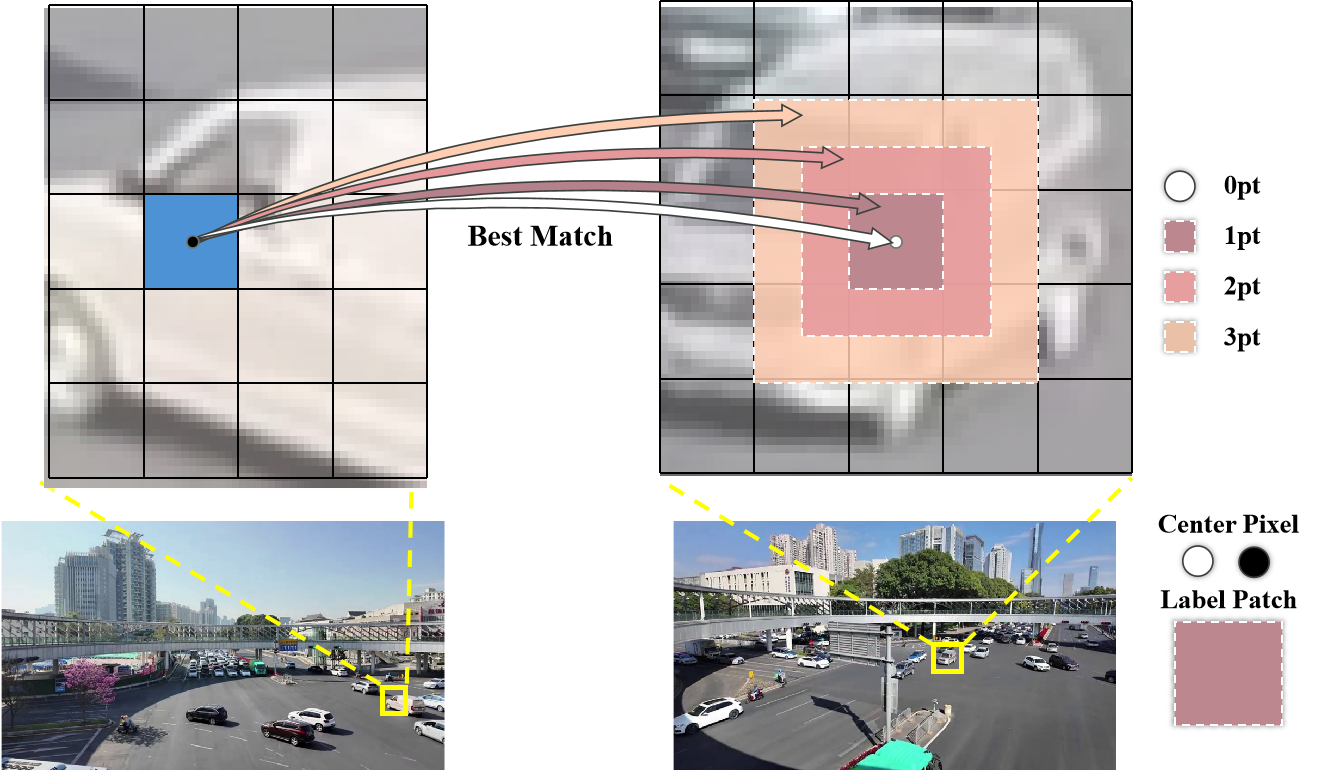}
    \caption{
        \textbf{Illustration of patch-level distance.}
        The diagram shows a patch-centered matching evaluation where the predicted patch center (yellow) is compared against the ground-truth center (red). Patch-level distance is defined based on the maximum offset between centers, normalized to patch units. For example, $d_{\text{patch}}=1$ covers the central patch, while $d_{\text{patch}}=2$ includes a $2P \times 2P$ region consisting of 4 adjacent patches. This formulation accounts for the spatial tolerance inherent to patch-based descriptors.
    }
    \label{fig:patch_distance_definition}
\end{figure}

This metric reflects how far apart the centers of the two patches are, normalized in units of the patch size. Specifically, as shown in Fig.~\ref{fig:patch_distance_definition}:

\begin{itemize}
    \item $d_{\text{patch}} = 0$ indicates that the two patch centers coincide exactly.
    \item $d_{\text{patch}} = 1$ implies that the predicted center lies within a square region of size $P \times P$ centered at the ground-truth center—i.e., within the same patch.
    \item $d_{\text{patch}} = 2$ means the predicted center lies within a $2P \times 2P$ region but outside the $P \times P$ region, corresponding to four surrounding patches.
    \item More generally, $d_{\text{patch}} = T$ implies the predicted center lies within a $(T \cdot P) \times (T \cdot P)$ square centered on the ground-truth, but outside the region defined by threshold $T-1$.
\end{itemize}

This definition accommodates the inherent quantization in patch-based representation and enables a structured tolerance for matching evaluation.

\section{Datasets}
\label{appendix:Datasets}

\begin{figure*}[htbp]
    \centering
    \includegraphics[width=0.98\textwidth]{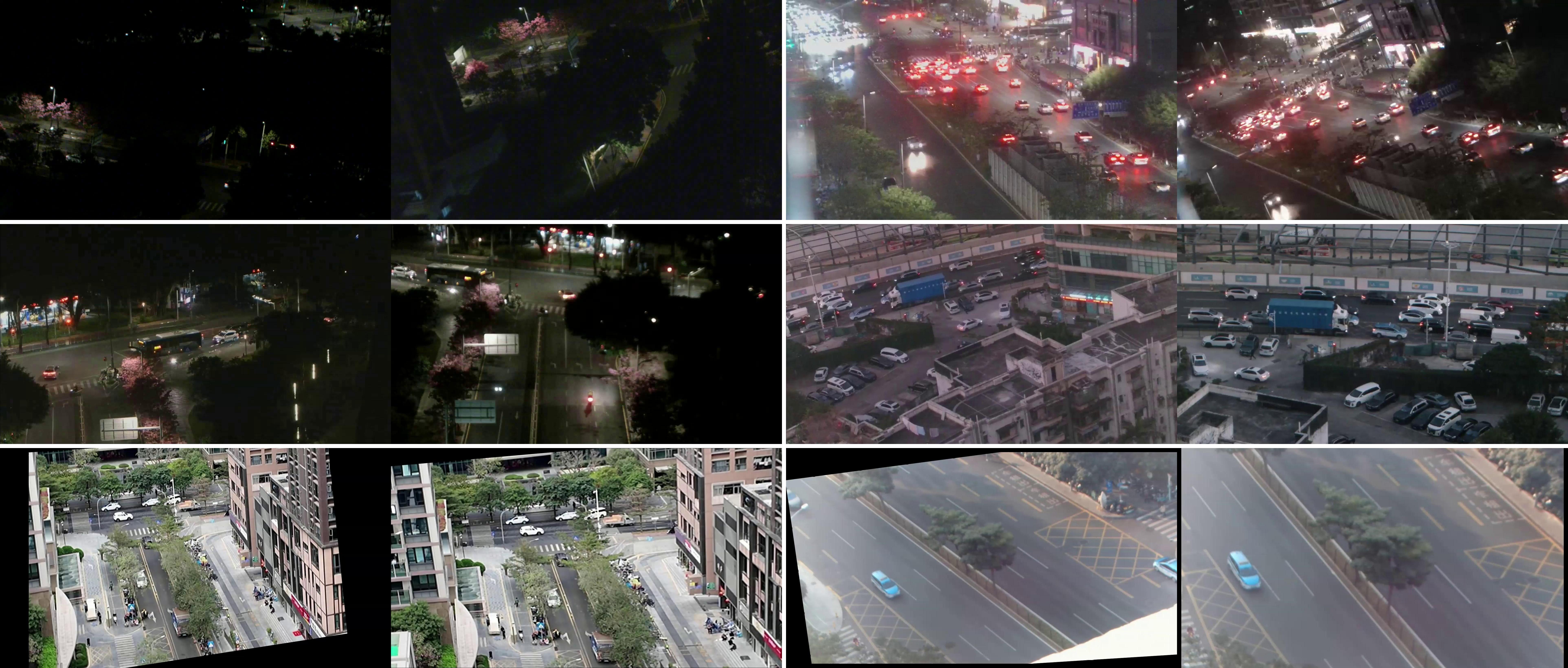}
    \caption{Cross-view Matching Samples in CityData}
    \label{fig:citydata_dataset}
\end{figure*}

\begin{figure*}[htbp]
    \centering
    \includegraphics[width=0.98\textwidth]{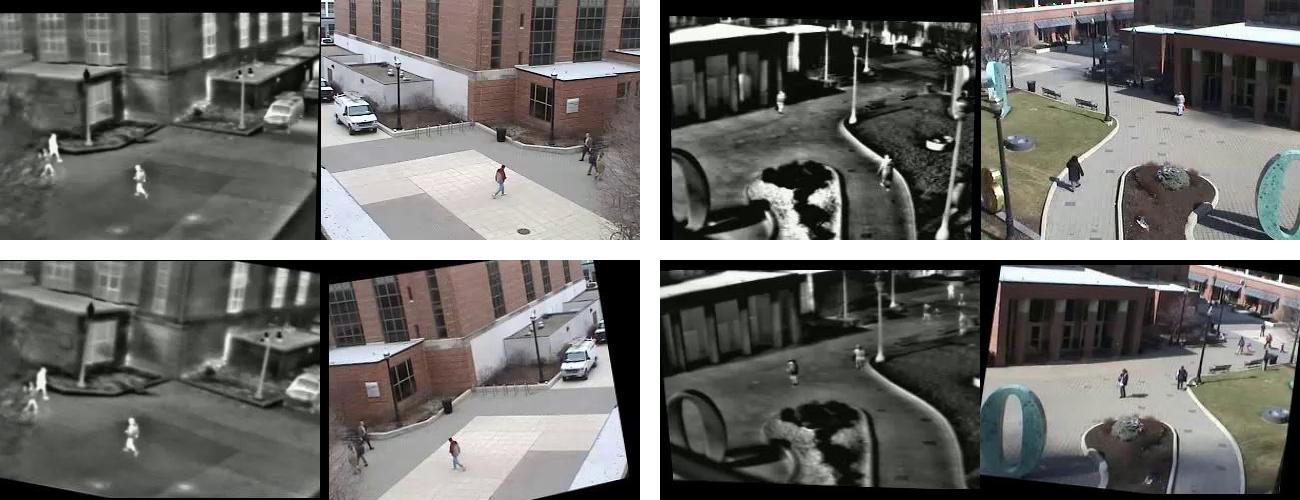}
    \caption{Augmented Video Frames in OTCBVS-AUG Dataset(Top: Horizontal Flip; Bottom: Perspective Transformation)}
    \label{fig:otcbvs_dataset}
\end{figure*}

Fig.~\ref{fig:citydata_dataset} shows some examples from our CityData dataset, and Fig.~\ref{fig:otcbvs_dataset} presents some examples from the OTVBVS-Aug dataset.

\section{Validation of Pseudo Labels Generation}
\label{appendix:Pseudo_Labels}
\begin{figure*}[htbp]
    \centering
    \includegraphics[width=0.95\textwidth]{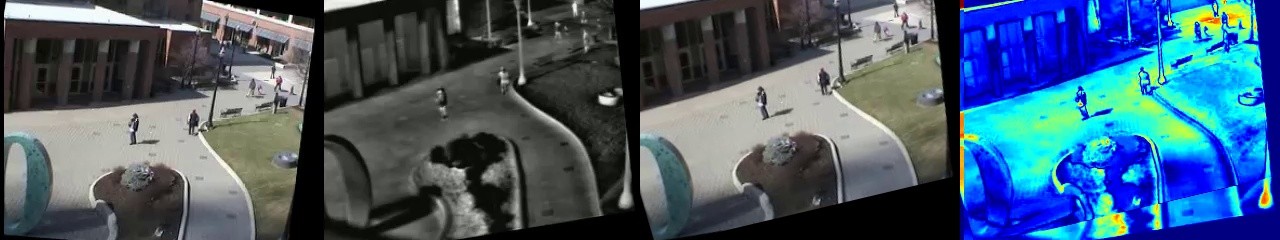}
    
    \vspace{2mm}
    \includegraphics[width=0.95\textwidth]{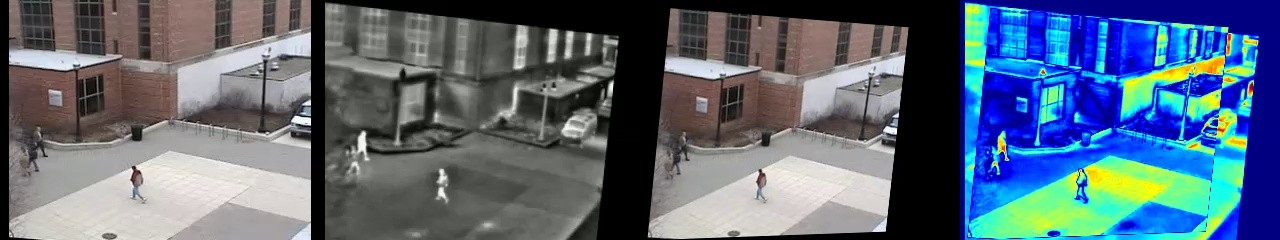}
    
    \vspace{2mm}
    \includegraphics[width=0.95\textwidth]{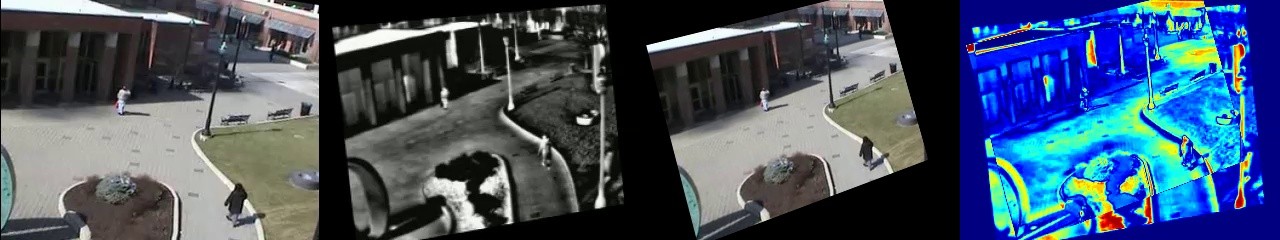}
    
    \caption{Validation of perspective transformation labels generated by MINIMA\_LoFTR on OTCBVS-Aug dataset. For each frame: original RGB image, IR image, warped RGB image using $MINIMA_{LoFTR}$-computed transformation, and difference heat map (blue: low difference, red: high difference).}
    \label{fig:heatmap_analysis}
\end{figure*}

Due to the lack of ground truth labels in the OTCBVS-Aug dataset, we propose to utilize $MINIMA_{LoFTR}$'s matching results to generate pseudo labels for quantitative evaluation. To validate the reliability of these generated labels, we conduct a visual analysis through heat map visualization, as shown in Fig. \ref{fig:heatmap_analysis}.

For each frame, we first obtain the perspective transformation matrix using $MINIMA_{LoFTR}$'s matching results. The visualization consists of four components: the original RGB frame, the IR frame, the warped RGB frame using the estimated transformation, and a heat map showing pixel-wise differences. The heat map generation follows a systematic process: (1) grayscale conversion using weighted formula (Gray = 0.299R + 0.587G + 0.114B), (2) absolute difference computation between the warped RGB and IR frames, (3) min-max normalization to [0, 255], and (4) JET colormap visualization.

The heat maps demonstrate that the perspective transformations computed from $MINIMA_{LoFTR}$ matches provide generally acceptable alignment between modalities, as evidenced by the dominant blue regions in structural areas such as building outlines and ground features. While higher differences (red) are observed in certain areas, these can be attributed to two distinct factors: inherent modality-specific characteristics and potential alignment errors. The former is expected due to the fundamental differences between RGB and IR imaging, particularly in areas with unique thermal signatures or texture details.

However, a closer inspection of the heat maps reveals noticeable "seam lines" in certain regions, indicating the existence of local misalignments in the $MINIMA_{LoFTR}$-generated transformations. \textbf{These alignment imperfections in our pseudo labels likely result in underestimated accuracy metrics for our method during quantitative evaluation.} Despite these limitations, the overall consistency of the transformations suggests that using $MINIMA_{LoFTR}$ matches as pseudo labels remains a viable approach for evaluation, albeit with the understanding that the reported accuracy might be conservative due to the inherent errors in the labels themselves.

\section{More Results}

\subsection{Time Consumption}

The table \ref{tab:time_consumption} compares the time consumption of each method across various datasets. Traditional keypoint-based approaches generally exhibit the shortest runtime. However, this is primarily because only a very limited number of keypoints can be detected and matched in these challenging datasets—particularly in RGB-IR datasets, where keypoint-based methods almost completely fail.

\textbf{MINIMA} demonstrates highly consistent time consumption across different datasets, with \texttt{MINIMA\textsubscript{LG}} requiring approximately 0.24 seconds and \texttt{MINIMA\textsubscript{LoFTR}} around 0.35 seconds to process a single image pair, highlighting its efficiency. It is important to note, however, that MINIMA is a model-based approach evaluated on a GPU, which offers a significant computational advantage over CPU-based methods.

Even for the low-resolution OTCBVS dataset, MINIMA requires this level of processing time. In contrast, our method achieves near real-time performance on the OTCBVS dataset, which also benefits from the relatively simple motion patterns in the videos. Results from the LLVIP and CityData datasets indicate that, despite the adoption of segment-wise matching, video duration significantly affects our method's overall runtime. This is because we filter out patches with minimal motion before matching. As the video length increases, more motion is captured, leading to fewer patches being discarded. As a result, both the complexity of the matching process and the number of matched patches increase.

Furthermore, our method allows for a flexible trade-off between computation time and the number of matched patches. In the experiments, the method was configured to perform one patch matching iteration followed by three propagation steps to maximize the number of matches. However, if faster processing is required, the method can be adjusted to terminate earlier once a sufficient number of matches are found. This adaptability makes the method suitable for scenarios where real-time performance is critical, even at the expense of reduced matching density.

Lastly, it is worth noting that our current implementation is written in Python and remains relatively unoptimized, as the primary focus of this work is to validate the feasibility of using temporal signals for matching. Improving performance through optimization is a key direction for future work.

\begin{table}[htbp]
    \centering
    \small
    \setlength{\tabcolsep}{3pt}
    \caption{Computation Time Comparison (S) of Feature Matching Paradigms: Video-Based Multi-Patch Propagation ($\tau_1$/$\tau_2$ Decomposition) vs. Image-Based Methods on Extracted Video Frames}
    \label{tab:time_consumption}
    \begin{tabular}{@{}l*{4}{c}@{}}
    \toprule
    Method & OTCBVS & CityData & \multicolumn{2}{c}{LLVIP} \\
    \cmidrule(l){4-5}
    & & & (2$\sim$4min) & (5$\sim$11min) \\
    \midrule
    SIFT & 0.011$\pm$0.001 & 0.843$\pm$2.116 & 0.141$\pm$0.091 & - \\
    ORB & 0.005$\pm$0.006 & 0.138$\pm$0.862 & 0.033$\pm$0.025 & - \\
    AKAZE & 0.041$\pm$0.001 & 0.884$\pm$0.069 & 0.484$\pm$0.316 & - \\
    \midrule
    MINIMA (LG) & 0.325$\pm$0.021 & 0.341$\pm$0.061 & 0.323$\pm$0.060 & - \\
    MINIMA (LoFTR) & 0.204$\pm$0.003 & 0.233$\pm$0.009 & 0.238$\pm$0.006 & - \\
    \midrule
    Ours (64) & & & & \\
    \quad $\tau_1$ & 0.007$\pm$0.030 & 0.497$\pm$0.348 & 0.385$\pm$0.459 & 1.003$\pm$1.737 \\
    \quad $\tau_2$ & 0.001$\pm$0.001 & 0.007$\pm$0.002 & 0.010$\pm$0.004 & 0.045$\pm$0.021 \\
    \addlinespace
    Ours (32) & & & & \\
    \quad $\tau_1$ & 0.002$\pm$0.005 & 0.010$\pm$0.004 & 0.026$\pm$0.011 & 0.056$\pm$0.050 \\
    \quad $\tau_2$ & 0.005$\pm$0.015 & 0.034$\pm$0.020 & 0.068$\pm$0.029 & 0.228$\pm$0.120 \\
    \addlinespace
    Ours (16) & & & & \\
    \quad $\tau_1$ & 0.005$\pm$0.016 & 0.086$\pm$0.025 & 0.118$\pm$0.047 & 0.238$\pm$0.194 \\
    \quad $\tau_2$ & 0.016$\pm$0.055 & 0.198$\pm$0.160 & 0.400$\pm$0.198 & 1.039$\pm$0.370 \\
    \addlinespace
    Ours (8) & & & & \\
    \quad $\tau_1$ & 0.033$\pm$0.131 & 0.973$\pm$0.190 & 0.826$\pm$0.203 & 1.300$\pm$0.538 \\
    \quad $\tau_2$ & 0.119$\pm$0.449 & 1.911$\pm$1.593 & 3.942$\pm$2.515 & 10.457$\pm$5.819 \\
    \bottomrule
    \end{tabular}

    \vspace{1mm}
    \raggedright
    \footnotesize
    \textbf{Note:} 
    The number after our method refers to the Patch size. And $\tau_1$: Local patch matching time; $\tau_2$: Cumulative propagation time (3 iterations).\\
    The sequence construction time is excluded from the reported matching time, as it can be performed in parallel during video capture. \\
    The baseline methods were exclusively evaluated on short video segments (2--4 minutes) in LLVIP, as their frame-wise matching latency is duration-invariant.
\end{table}

\subsection{Results on CityData-Aug Dataset}

\begin{figure*}[htbp]
    \centering
    \begin{tabular}{c@{\hspace{1mm}}c@{\hspace{1mm}}c@{\hspace{1mm}}c@{\hspace{1mm}}c}
        & SIFT & MINIMA\_LoFTR & MINIMA\_LG & Ours \\
        \raisebox{0pt}{\rotatebox{90}{S1}} &
        \includegraphics[width=0.23\textwidth]{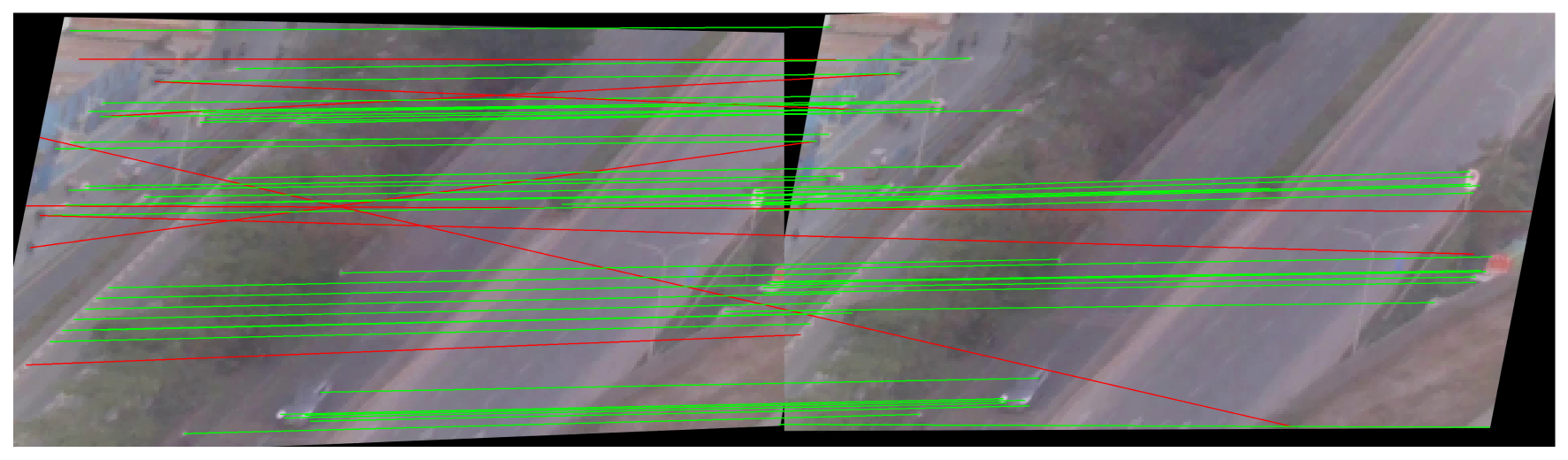} &
        \includegraphics[width=0.23\textwidth]{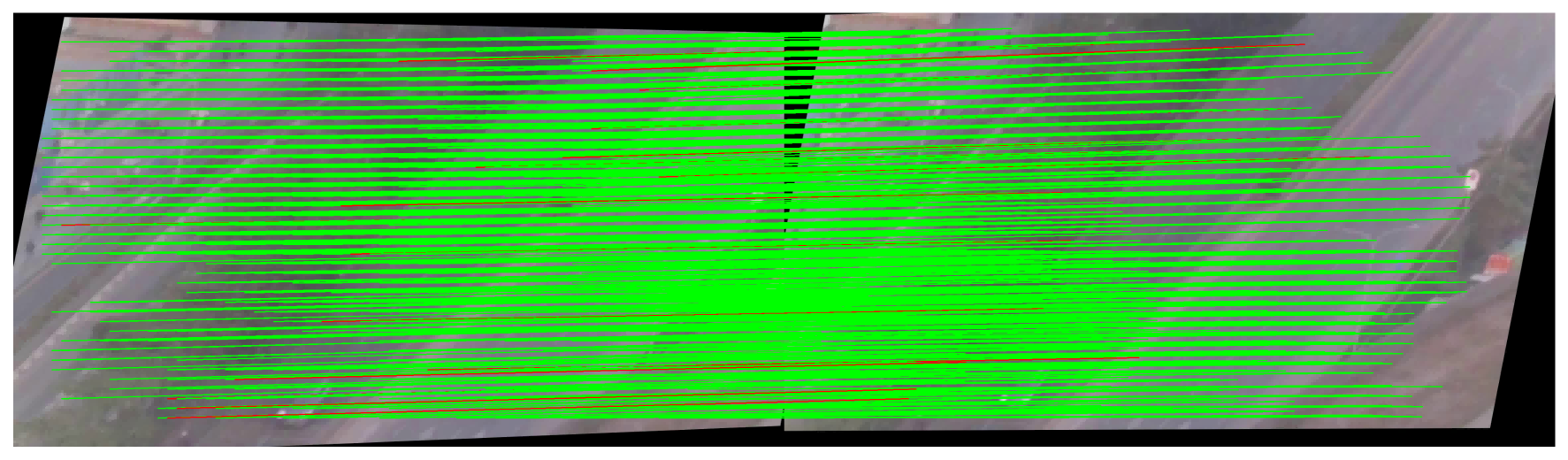} &
        \includegraphics[width=0.23\textwidth]{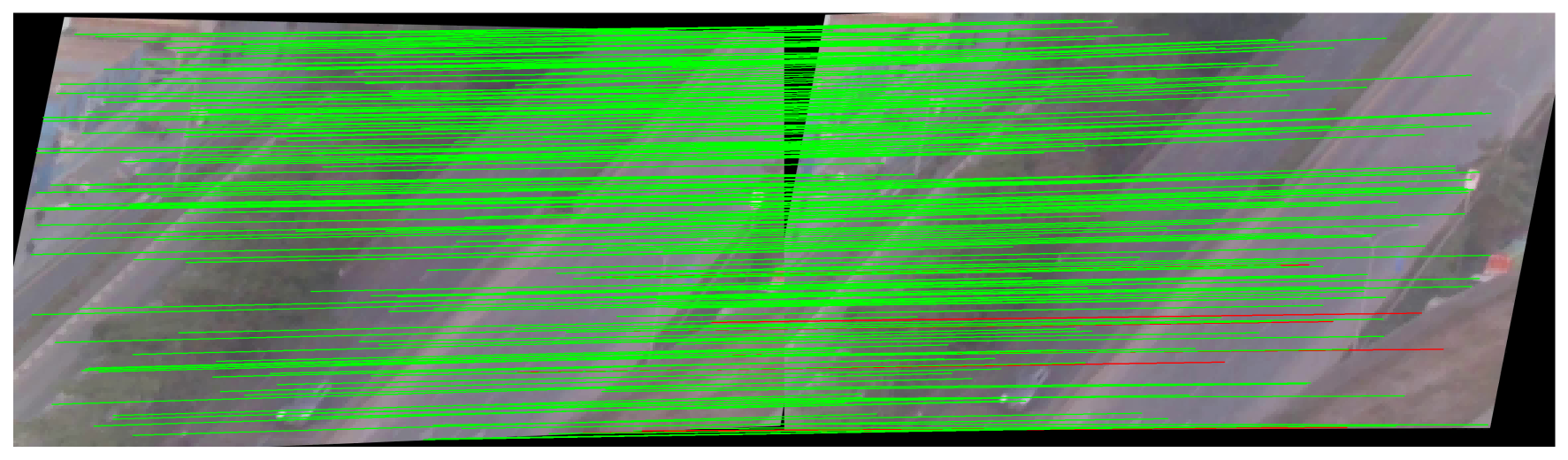} &
        \includegraphics[width=0.23\textwidth]{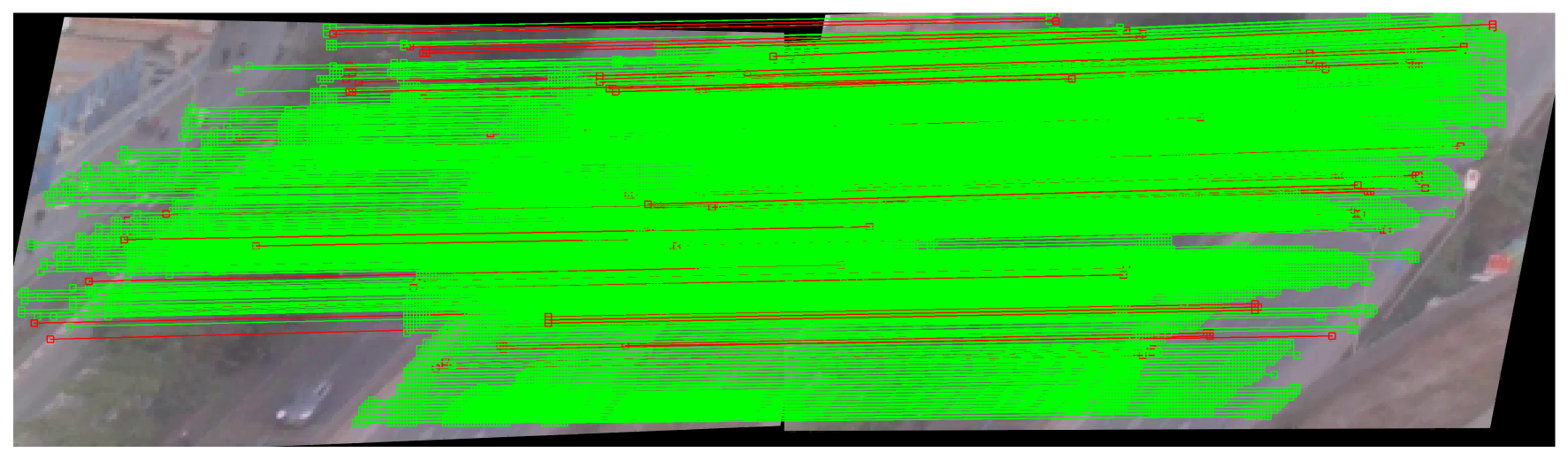} \\[2mm]

        \raisebox{0pt}{\rotatebox{90}{S2}} &
        \includegraphics[width=0.23\textwidth]{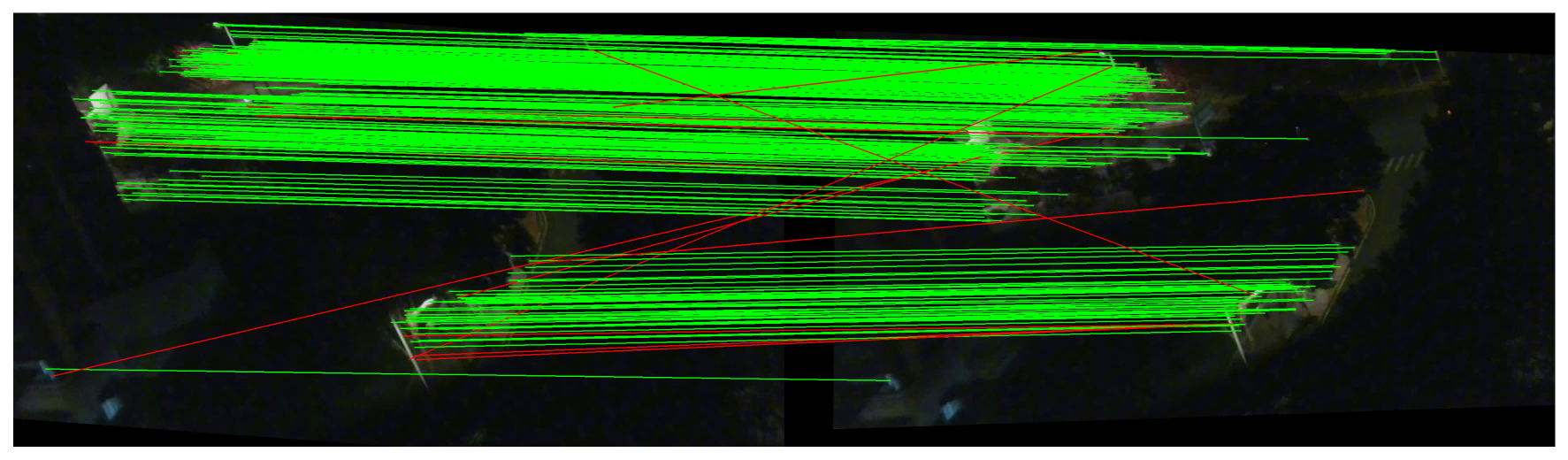} &
        \includegraphics[width=0.23\textwidth]{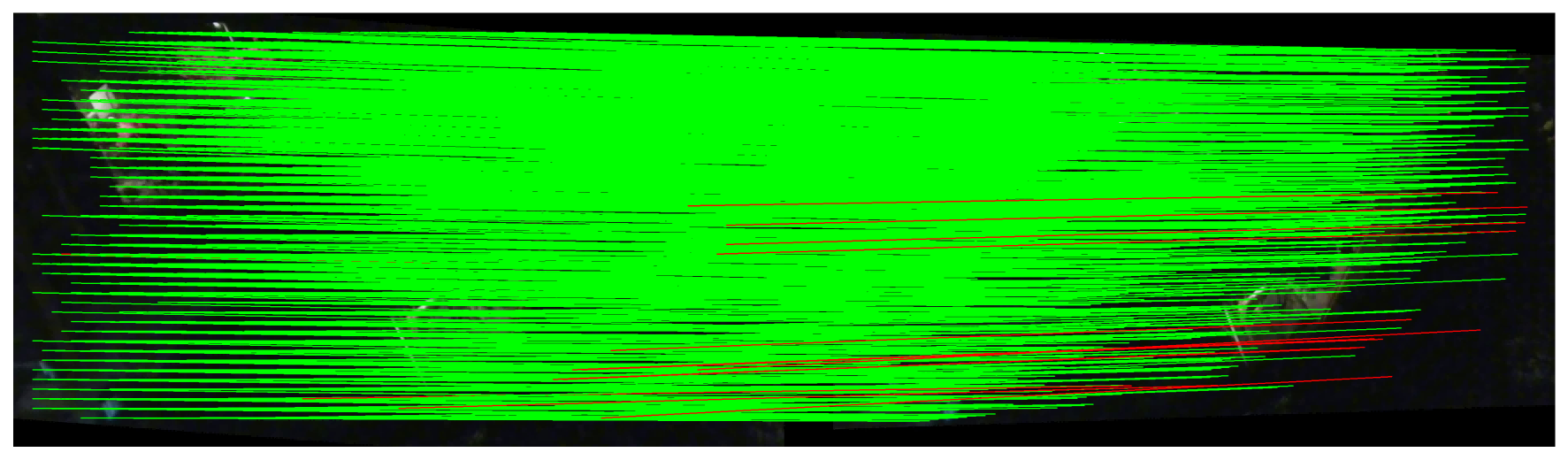} &
        \includegraphics[width=0.23\textwidth]{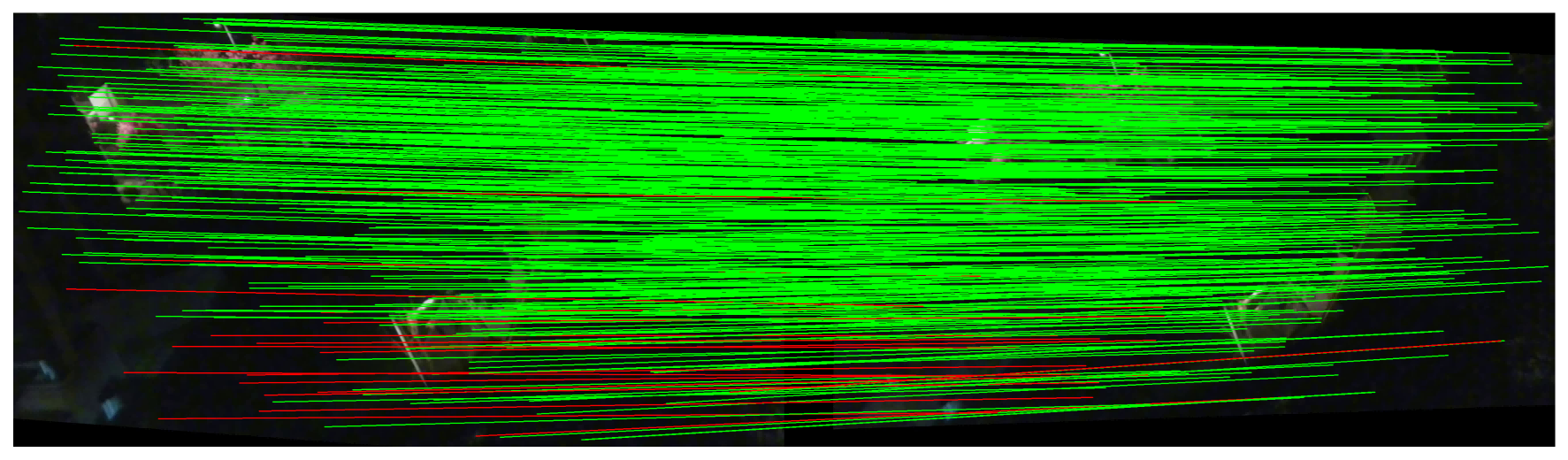} &
        \includegraphics[width=0.23\textwidth]{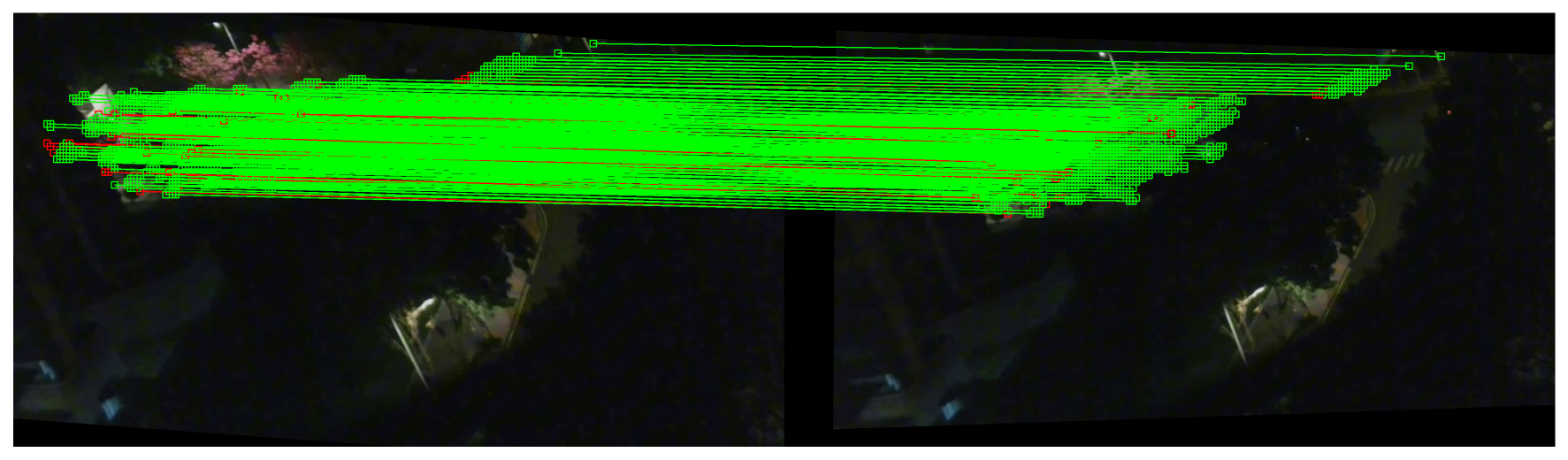} 
    \end{tabular}
    
    \caption{Qualitative comparison of feature matching results on CityData-Aug with 5-pixel threshold.}
    \label{fig:citydata}
\end{figure*}

Figure~\ref{fig:citydata} illustrates the matching results of different methods on the CityData-Aug dataset, where correspondences with a matching error below 20 pixels are shown in green, and those exceeding this threshold are displayed in red.

In Scene S1, all evaluated methods manage to establish a certain number of correct matches. However, due to the dominance of textureless regions (e.g., road surfaces), both SIFT and MINIMA produce relatively sparse correspondences. In contrast, our method, which leverages motion cues to inform the matching process, achieves superior performance in both the number and spatial accuracy of matches.

In Scene S2, although the environment is characterized by low lighting conditions, the presence of high-intensity areas (such as regions near streetlights) enables traditional feature-based methods to detect some keypoints. Nonetheless, these methods tend to focus primarily on brightly lit regions, leading to spatially biased results. In this scenario, the MINIMA method delivers the most accurate correspondences among the baselines, benefiting from its learning-based architecture. However, our method still maintains a competitive number of high-quality matches distributed more evenly across the scene, underscoring its robustness in handling challenging illumination conditions.

\subsection{Results on OTCBVS-Aug Dataset}

\begin{figure*}[htbp]
    \centering
    \begin{tabular}{c@{\hspace{1mm}}c@{\hspace{1mm}}c@{\hspace{1mm}}c@{\hspace{1mm}}c}
        & SIFT & MINIMA\_LoFTR & MINIMA\_LG & Ours \\
        \raisebox{0pt}{\rotatebox{90}{S1}} &
        \includegraphics[width=0.23\textwidth]{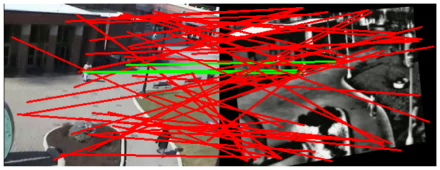} &
        \includegraphics[width=0.23\textwidth]{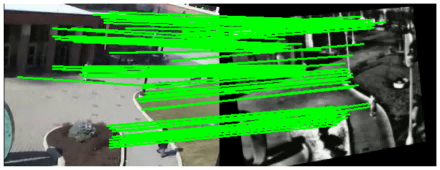} &
        \includegraphics[width=0.23\textwidth]{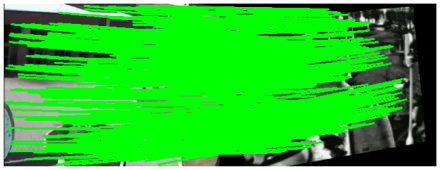} &
        \includegraphics[width=0.23\textwidth]{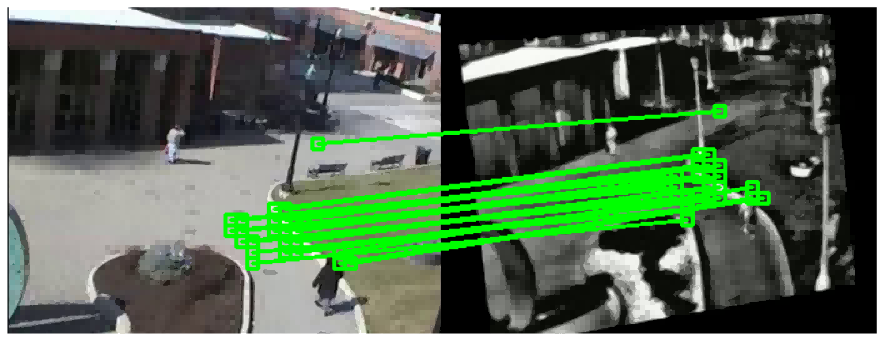} \\[2mm]

        \raisebox{0pt}{\rotatebox{90}{S2}} &
        \includegraphics[width=0.23\textwidth]{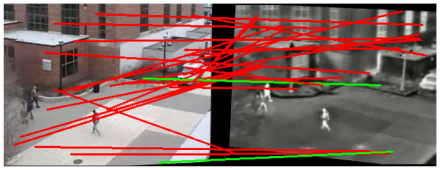} &
        \includegraphics[width=0.23\textwidth]{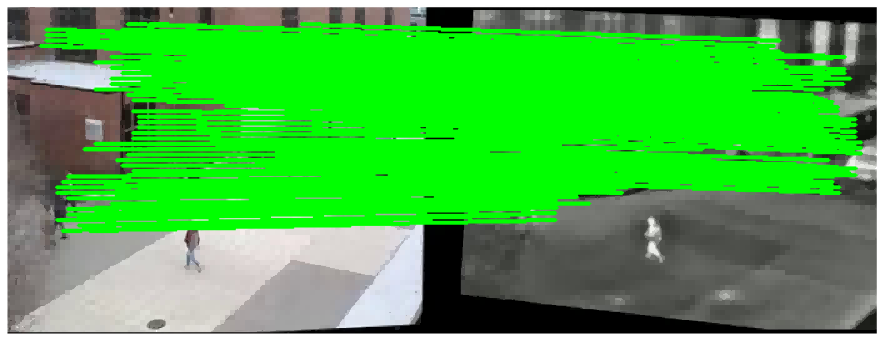} &
        \includegraphics[width=0.23\textwidth]{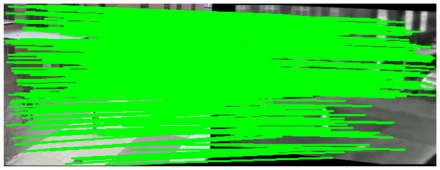} &
        \includegraphics[width=0.23\textwidth]{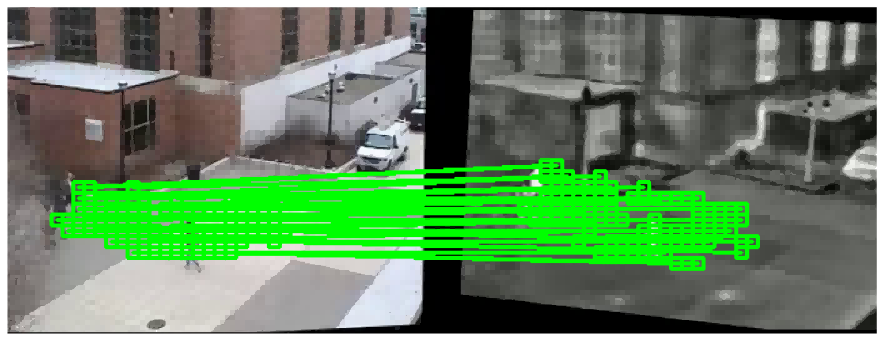} 
    \end{tabular}
    
    \caption{Qualitative comparison of feature matching results on OTCBVS-Aug with 20-pixel threshold.}
    \label{fig:otcbvs}
\end{figure*}

\subsection{Results on LLVIP Dataset}

\begin{figure*}[htbp]
    \centering
    \begin{tabular}{c@{\hspace{1mm}}c@{\hspace{1mm}}c@{\hspace{1mm}}c@{\hspace{1mm}}c}
        & SIFT & MINIMA\_LoFTR & MINIMA\_LG & Ours \\
        \raisebox{0pt}{\rotatebox{90}{S1}} &
        \includegraphics[width=0.23\textwidth]{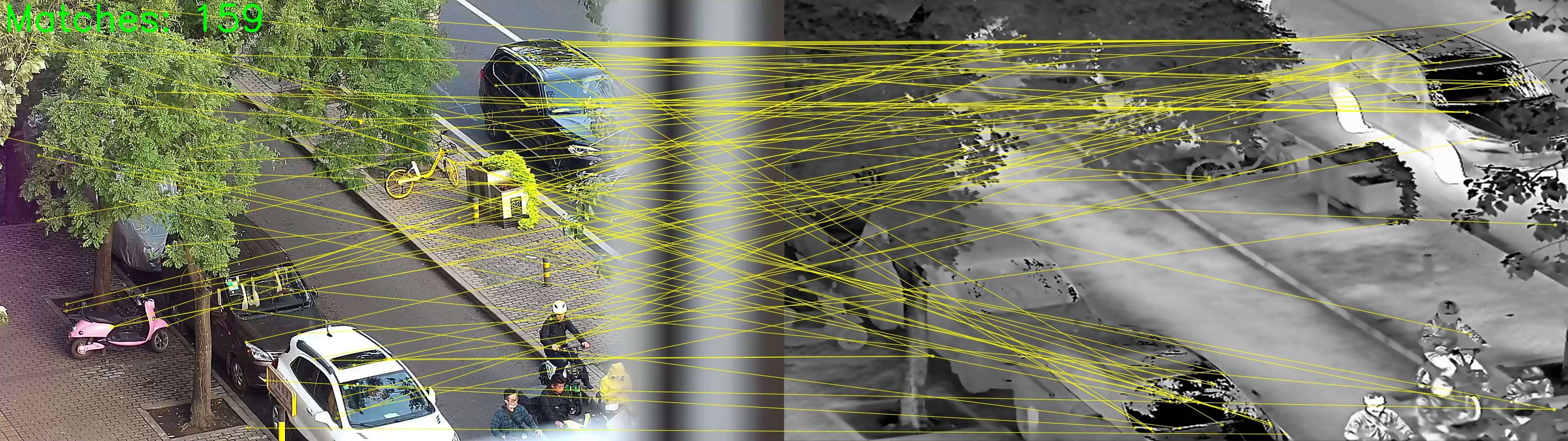} &
        \includegraphics[width=0.23\textwidth]{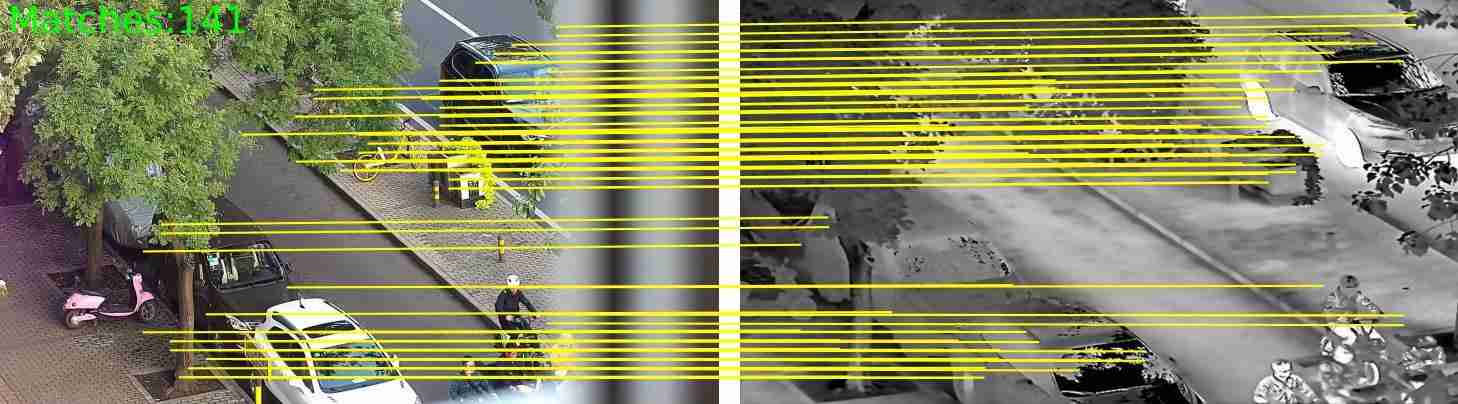} &
        \includegraphics[width=0.23\textwidth]{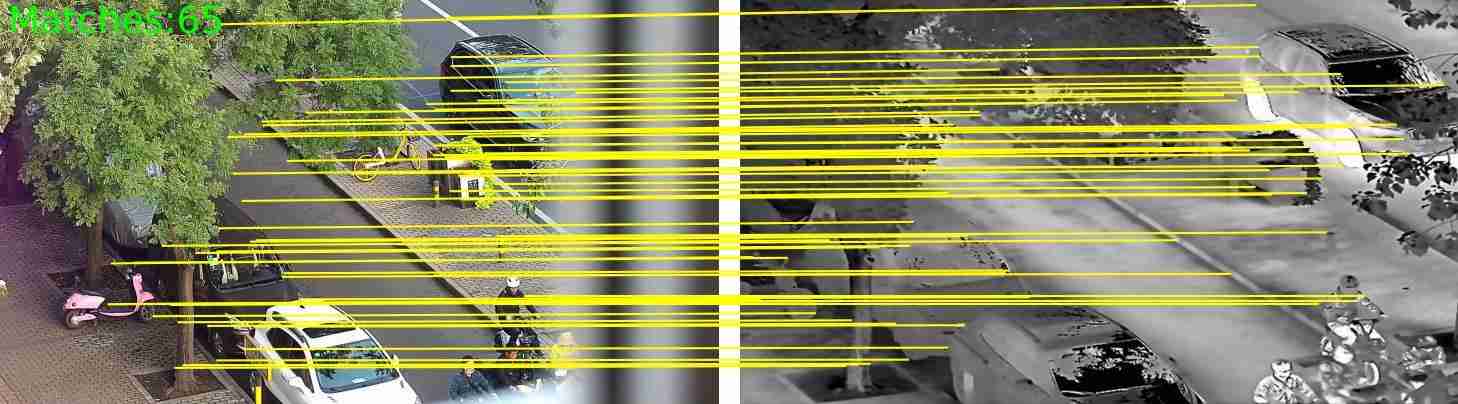} &
        \includegraphics[width=0.23\textwidth]{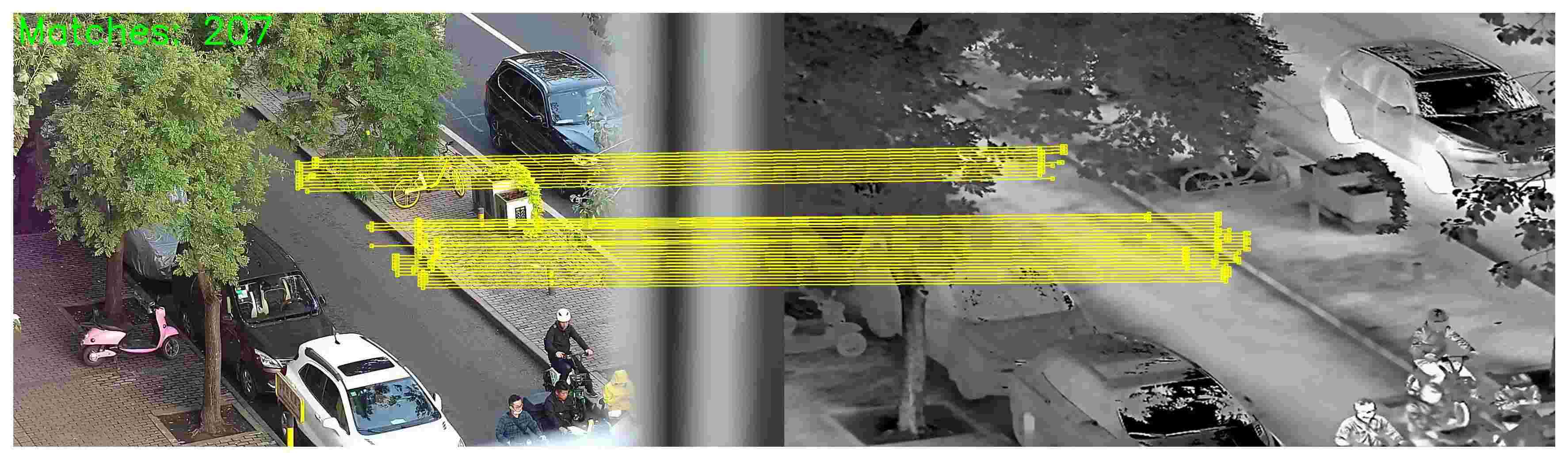} \\[2mm]

        \raisebox{0pt}{\rotatebox{90}{S2}} &
        \includegraphics[width=0.23\textwidth]{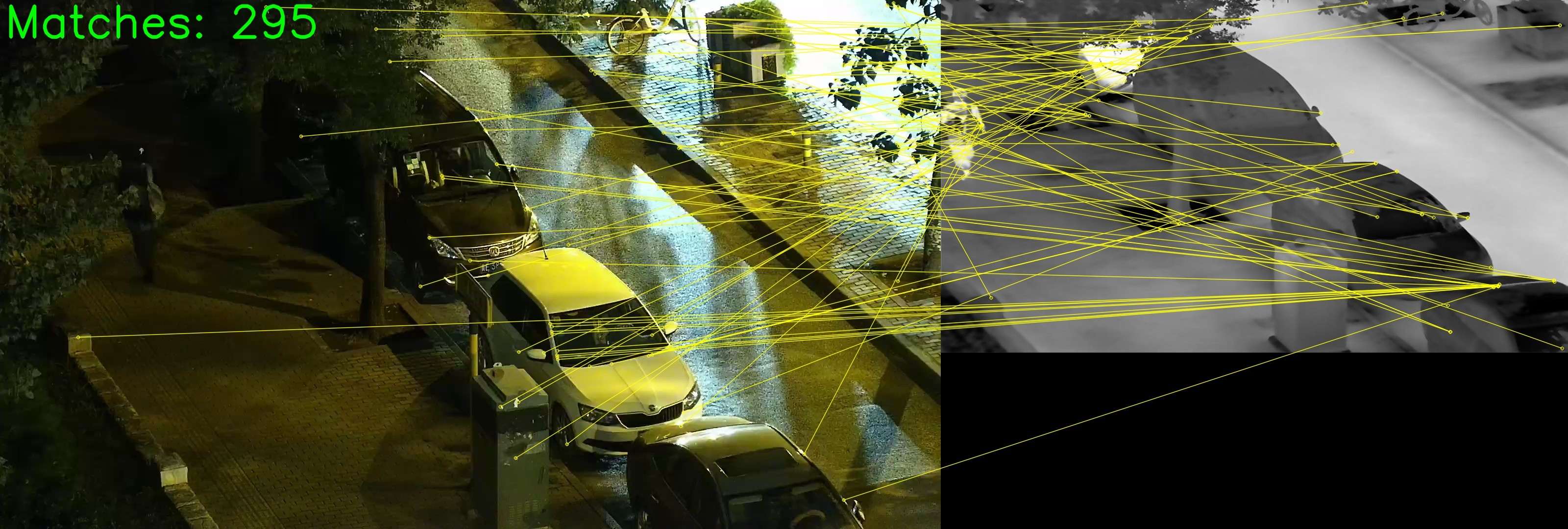} &
        \includegraphics[width=0.23\textwidth]{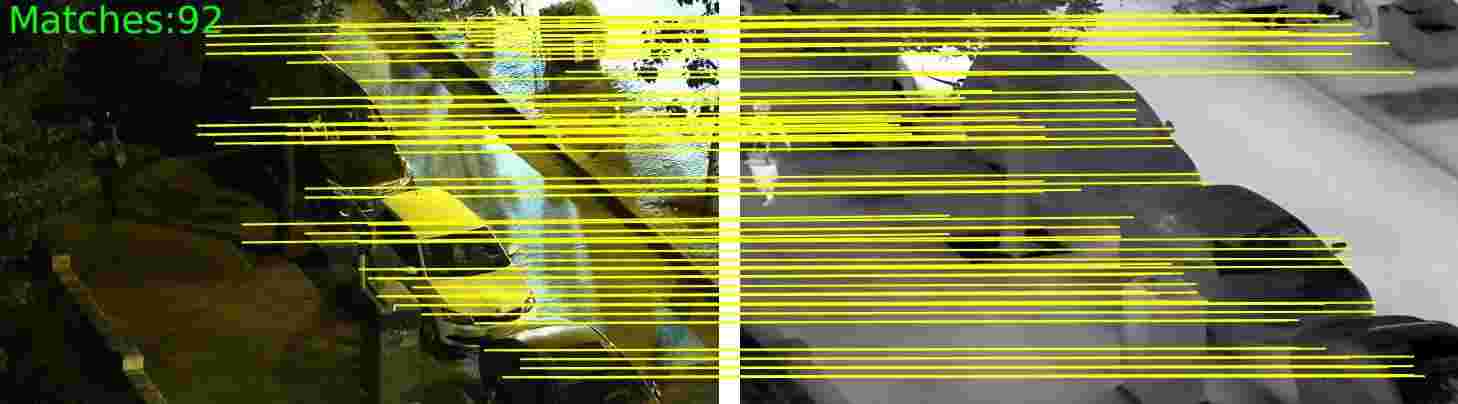} &
        \includegraphics[width=0.23\textwidth]{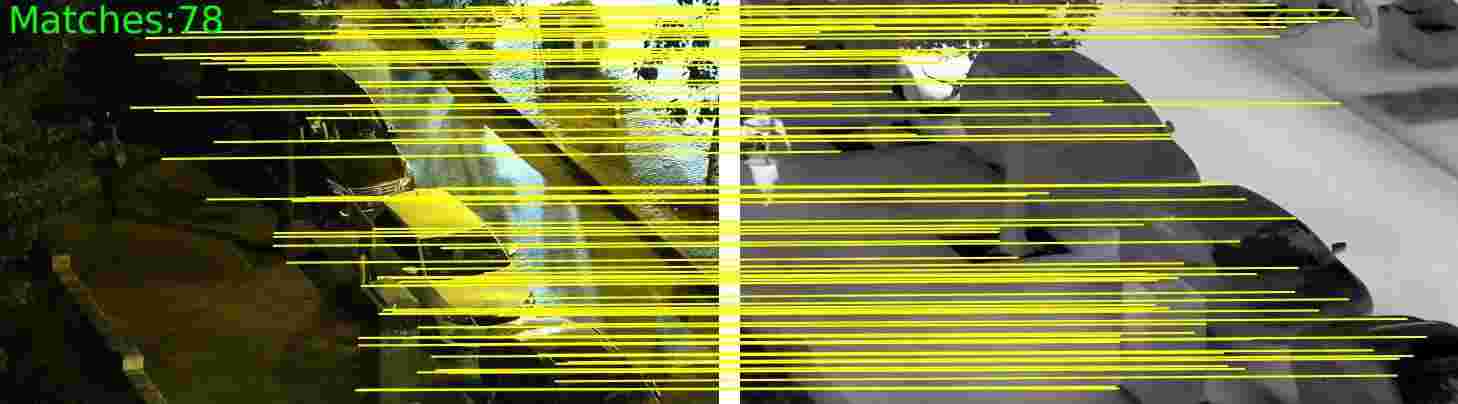} &
        \includegraphics[width=0.23\textwidth]{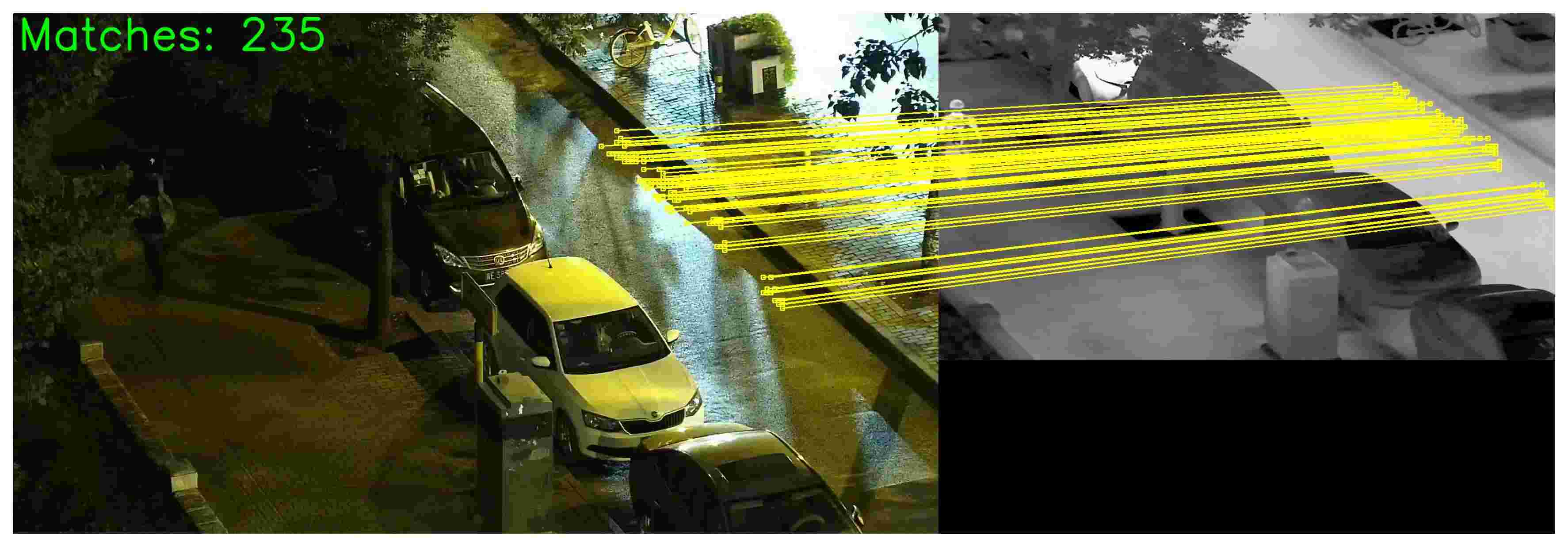} \\[2mm]

        \raisebox{0pt}{\rotatebox{90}{S3}} &
        \includegraphics[width=0.23\textwidth]{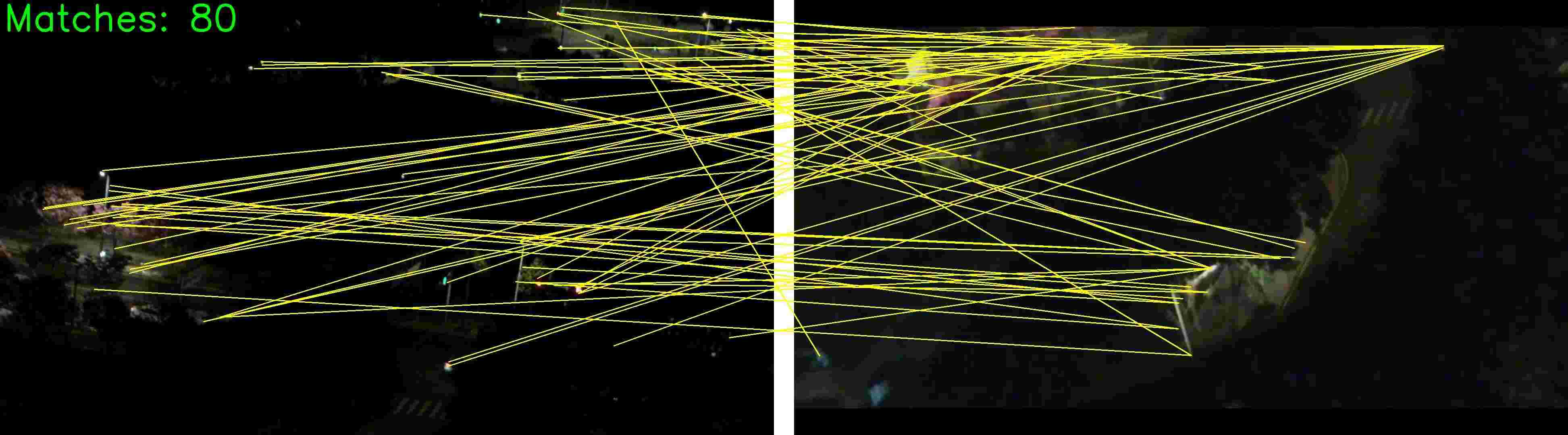} &
        \includegraphics[width=0.23\textwidth]{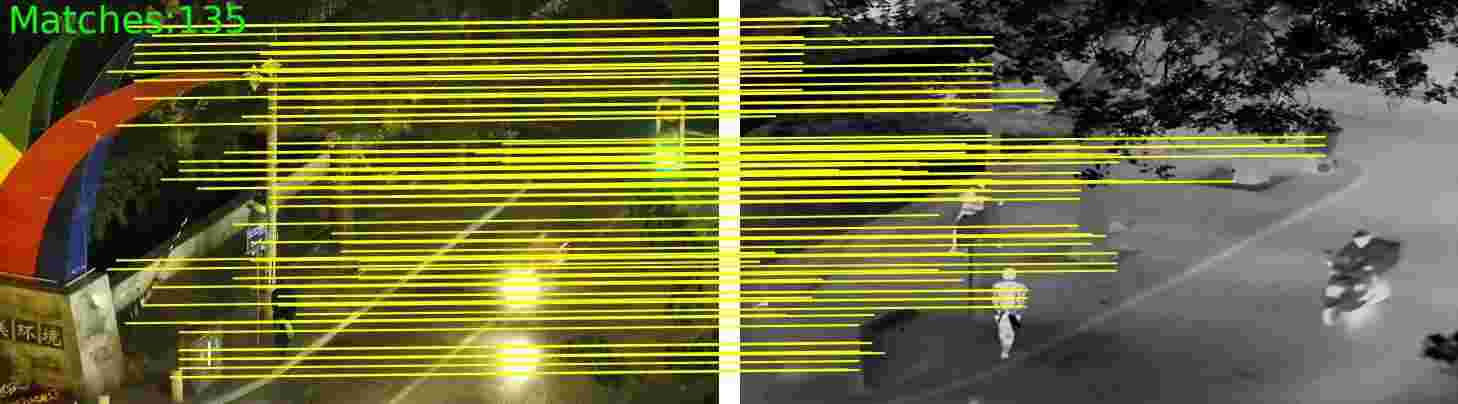} &
        \includegraphics[width=0.23\textwidth]{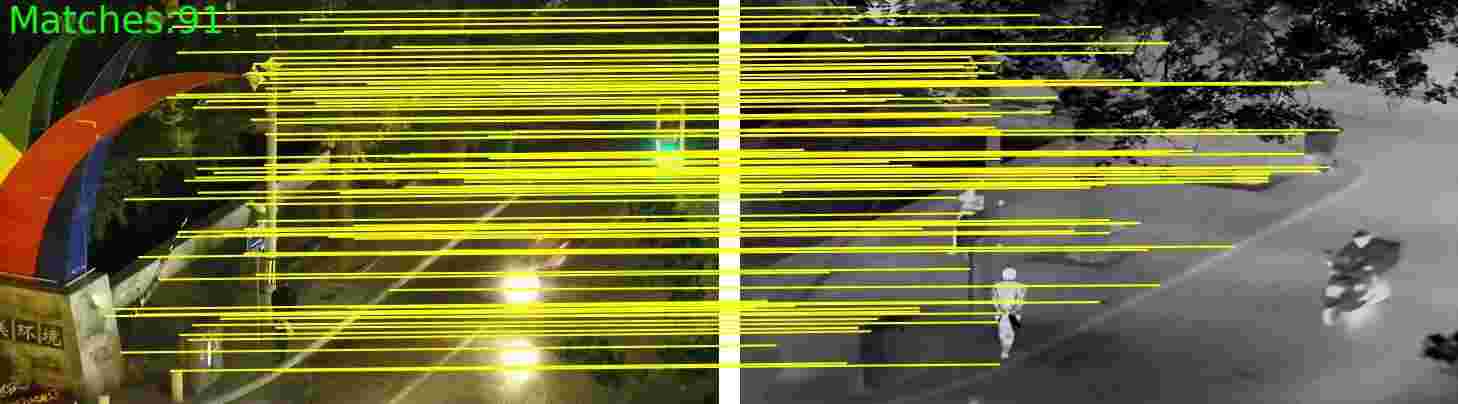} &
        \includegraphics[width=0.23\textwidth]{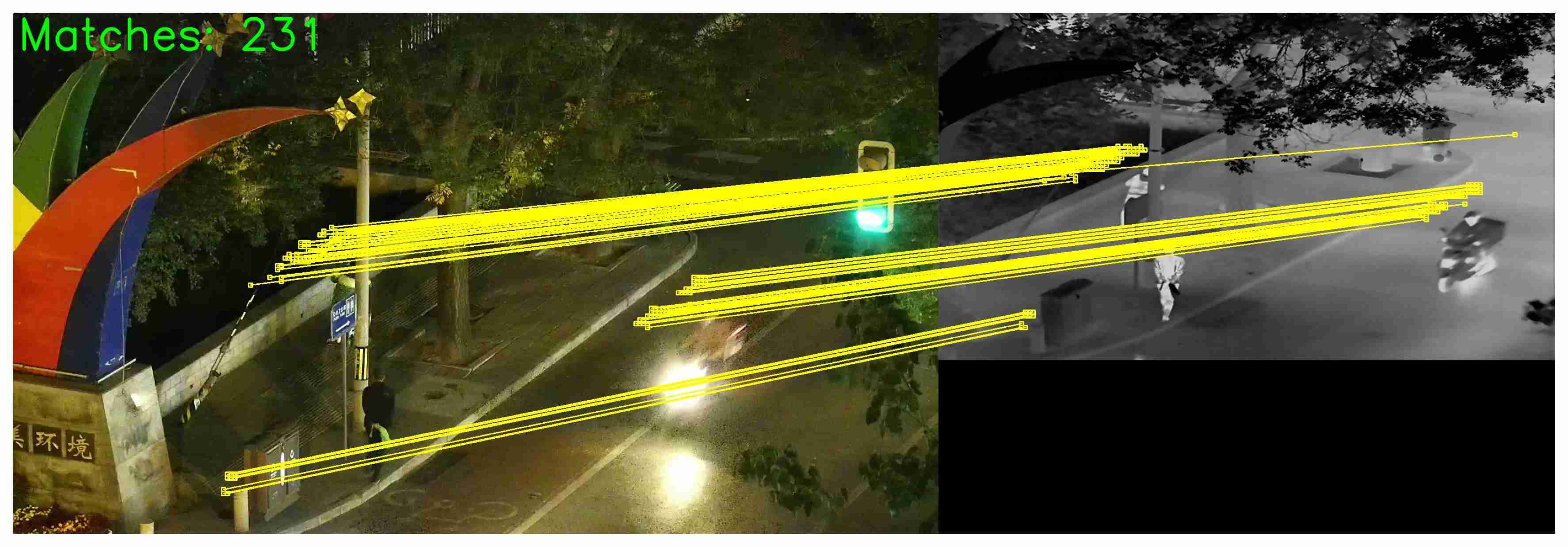} \\[2mm]

        \raisebox{0pt}{\rotatebox{90}{S4}} &
        \includegraphics[width=0.23\textwidth]{Img/appendix/LLVIP/3_SIFT.jpg} &
        \includegraphics[width=0.23\textwidth]{Img/appendix/LLVIP/3_minima_loftr.jpg} &
        \includegraphics[width=0.23\textwidth]{Img/appendix/LLVIP/3_minima_sp_lg.jpg} &
        \includegraphics[width=0.23\textwidth]{Img/appendix/LLVIP/3_ours.jpg}
    \end{tabular}
    
    \caption{Qualitative comparison of feature matching results on LLVIP.}
    \label{fig:LLVIP}
\end{figure*}

\subsection{Results on Additional Modalities }

\begin{figure*}[htbp]
    \centering
    \begin{tabular}{c@{\hspace{1mm}}c@{\hspace{1mm}}c@{\hspace{1mm}}c@{\hspace{1mm}}c}
        & SIFT & MINIMA\_LoFTR & MINIMA\_LG & Ours \\
        \raisebox{0pt}{\rotatebox{90}{Depth}} &
        \includegraphics[width=0.23\textwidth]{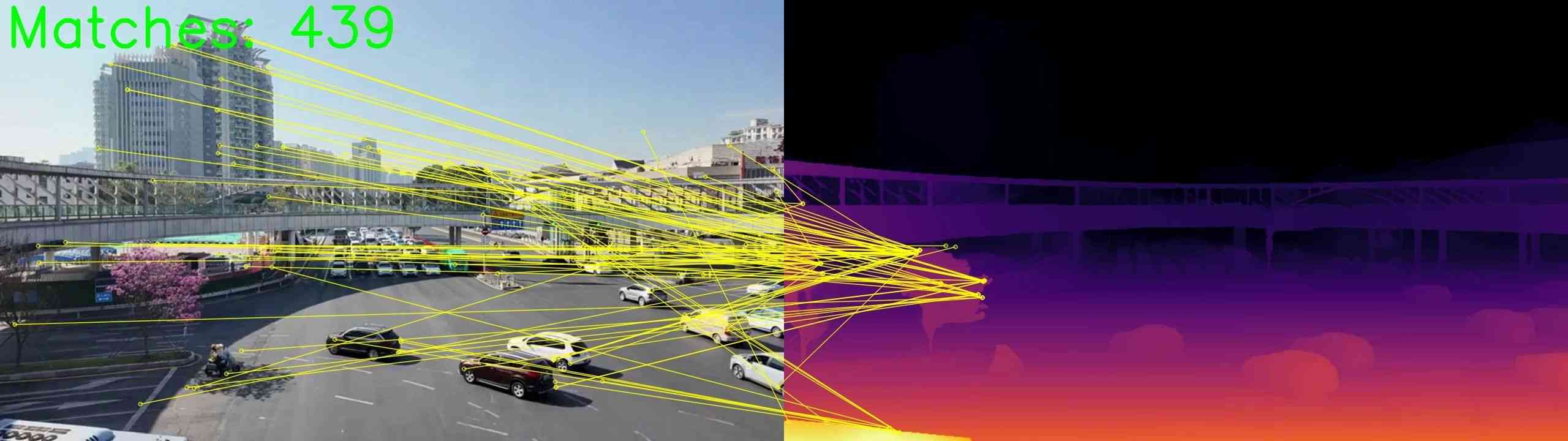} &
        \includegraphics[width=0.23\textwidth]{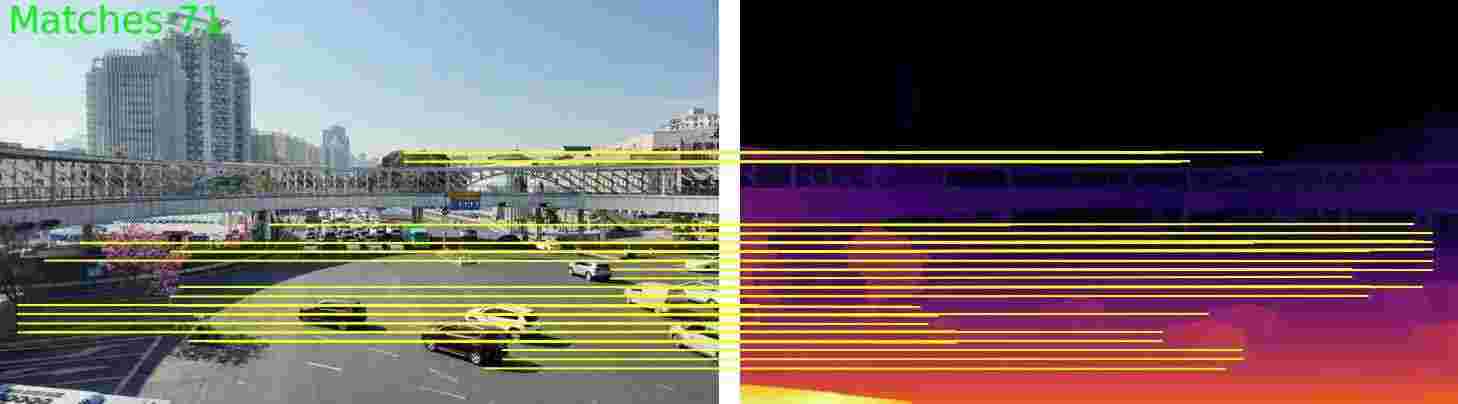} &
        \includegraphics[width=0.23\textwidth]{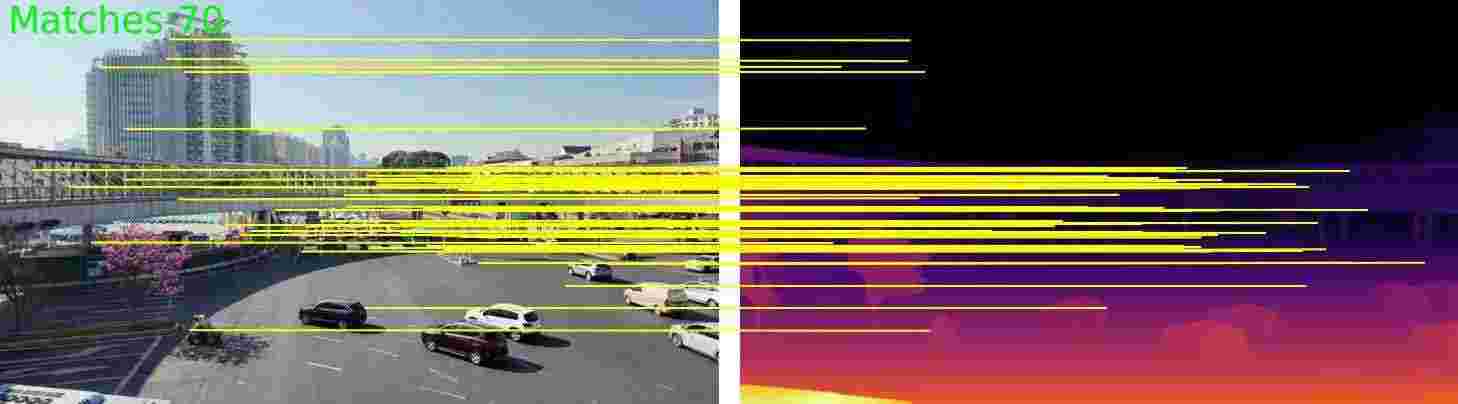} &
        \includegraphics[width=0.23\textwidth]{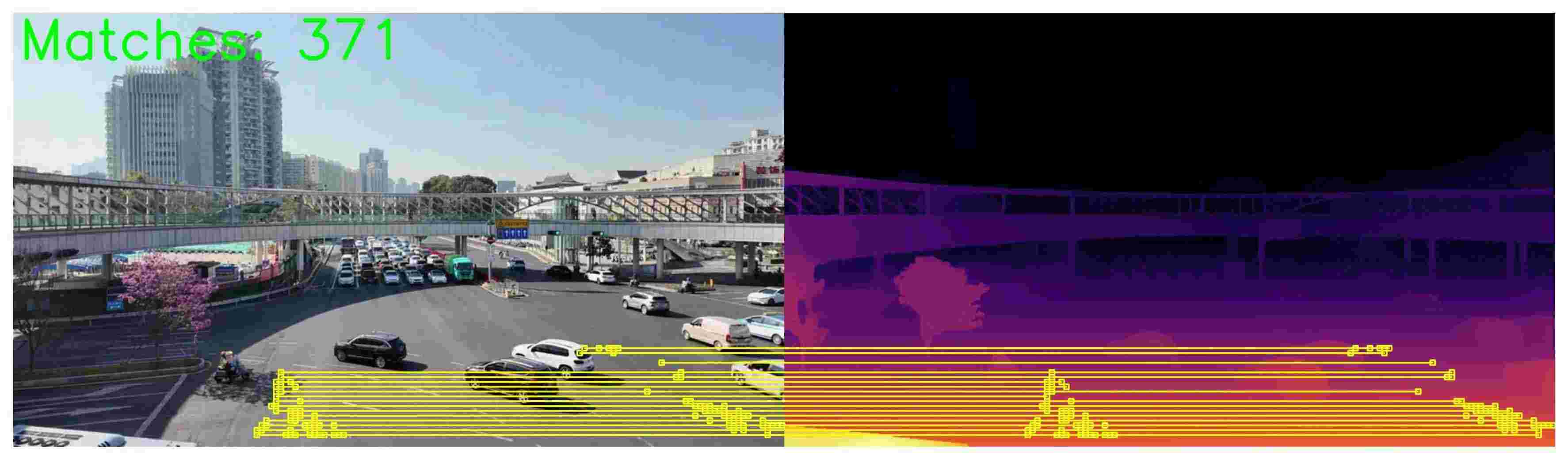} \\[2mm]

        \raisebox{0pt}{\rotatebox{90}{Event}} &
        \includegraphics[width=0.23\textwidth]{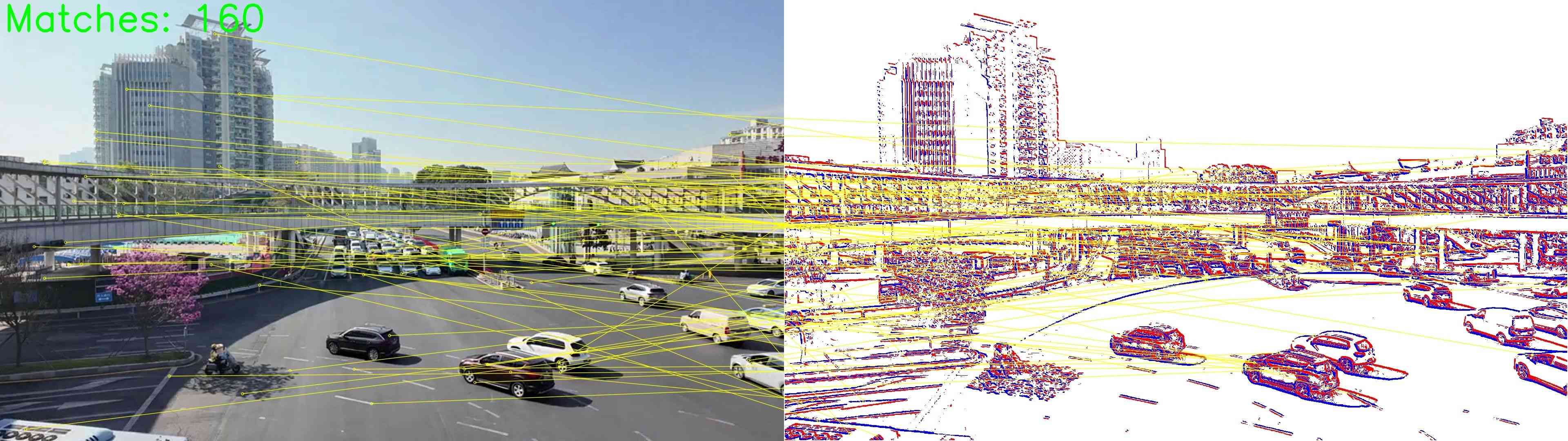} &
        \includegraphics[width=0.23\textwidth]{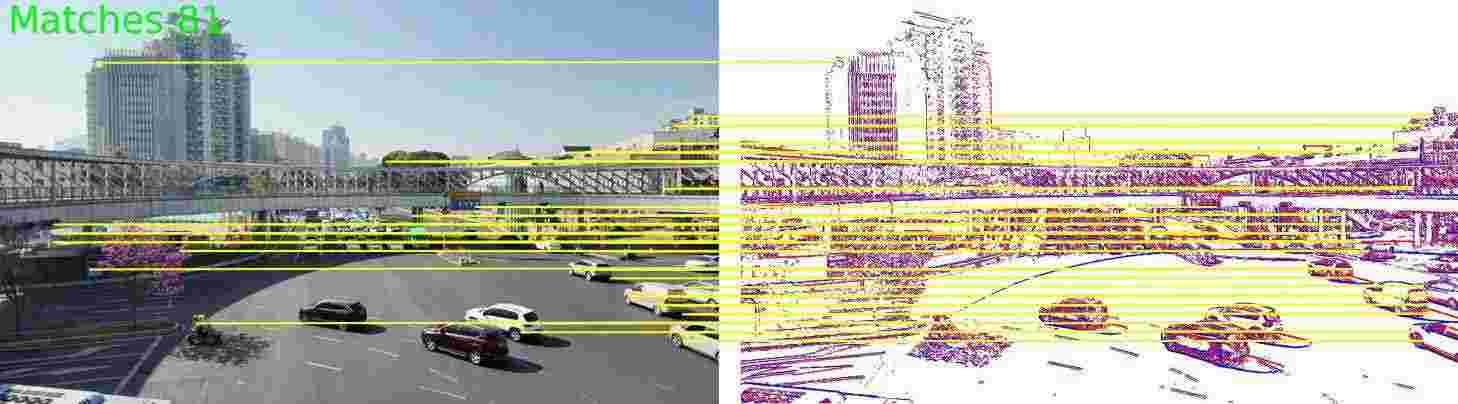} &
        \includegraphics[width=0.23\textwidth]{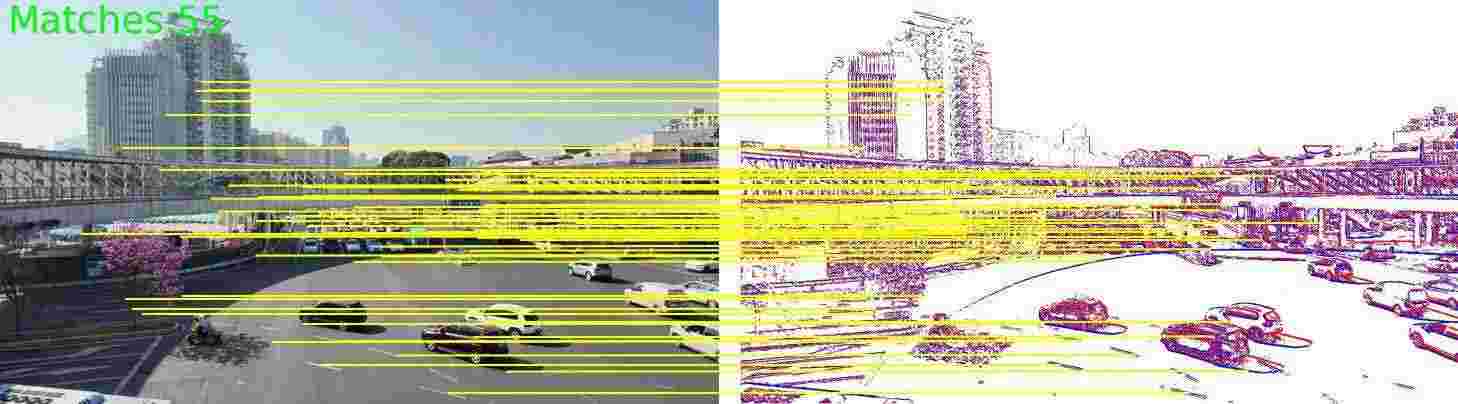} &
        \includegraphics[width=0.23\textwidth]{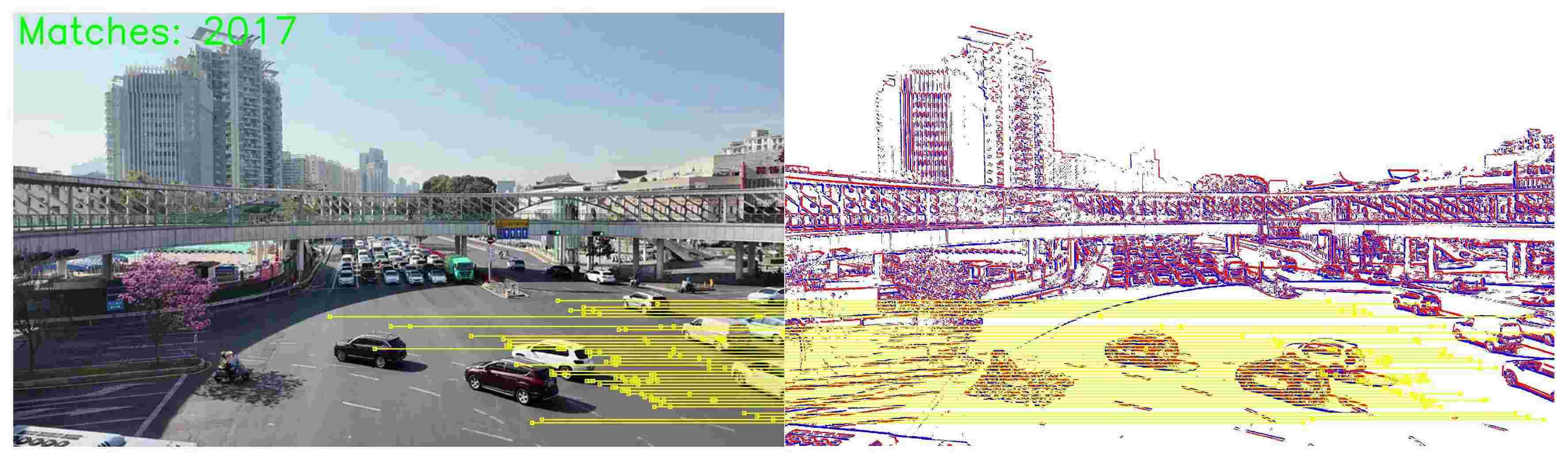} \\[2mm]

        \raisebox{0pt}{\rotatebox{90}{Normal}} &
        \includegraphics[width=0.23\textwidth]{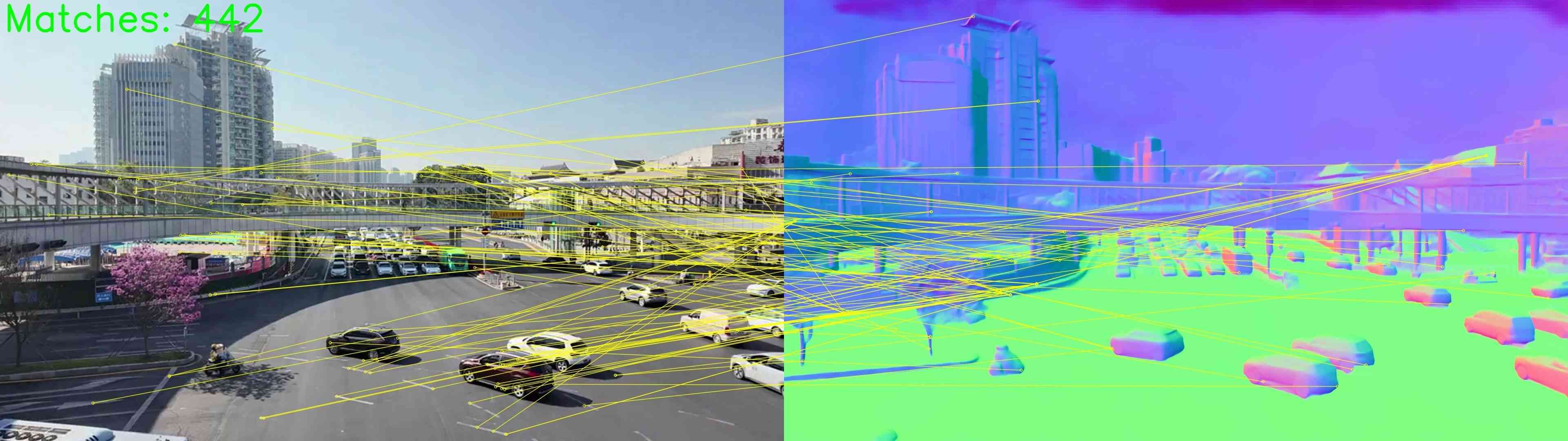} &
        \includegraphics[width=0.23\textwidth]{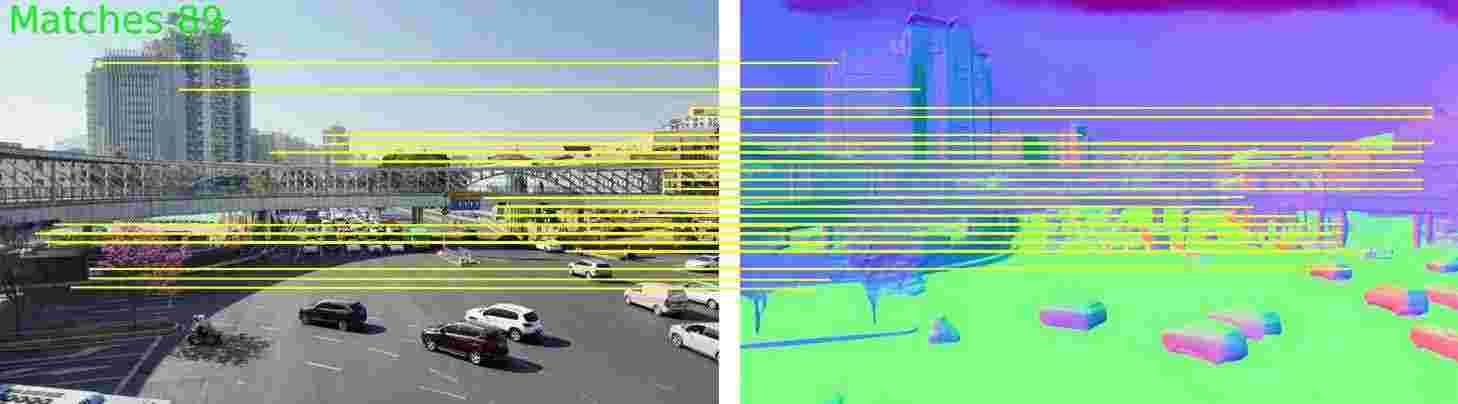} &
        \includegraphics[width=0.23\textwidth]{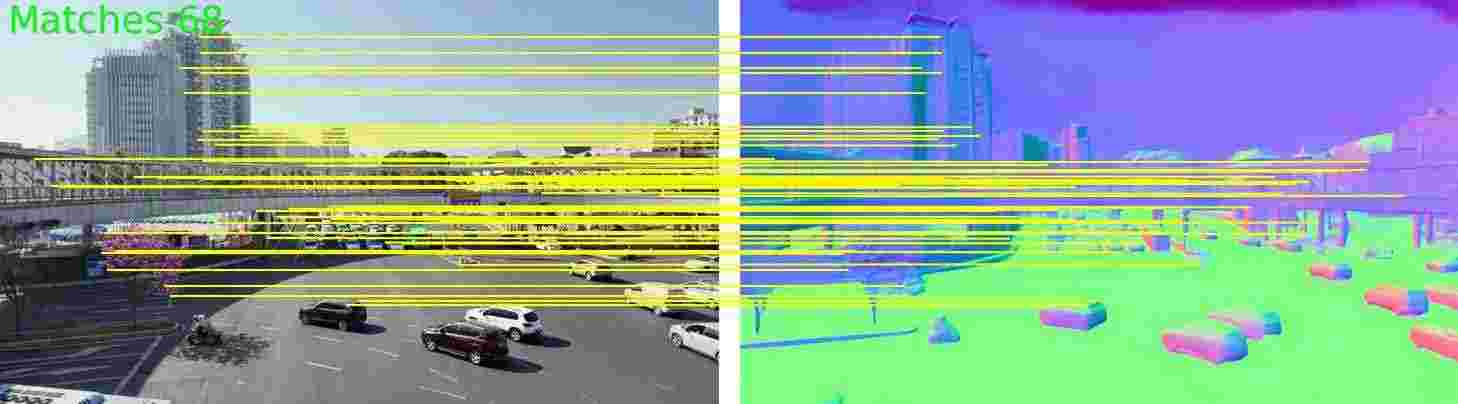} &
        \includegraphics[width=0.23\textwidth]{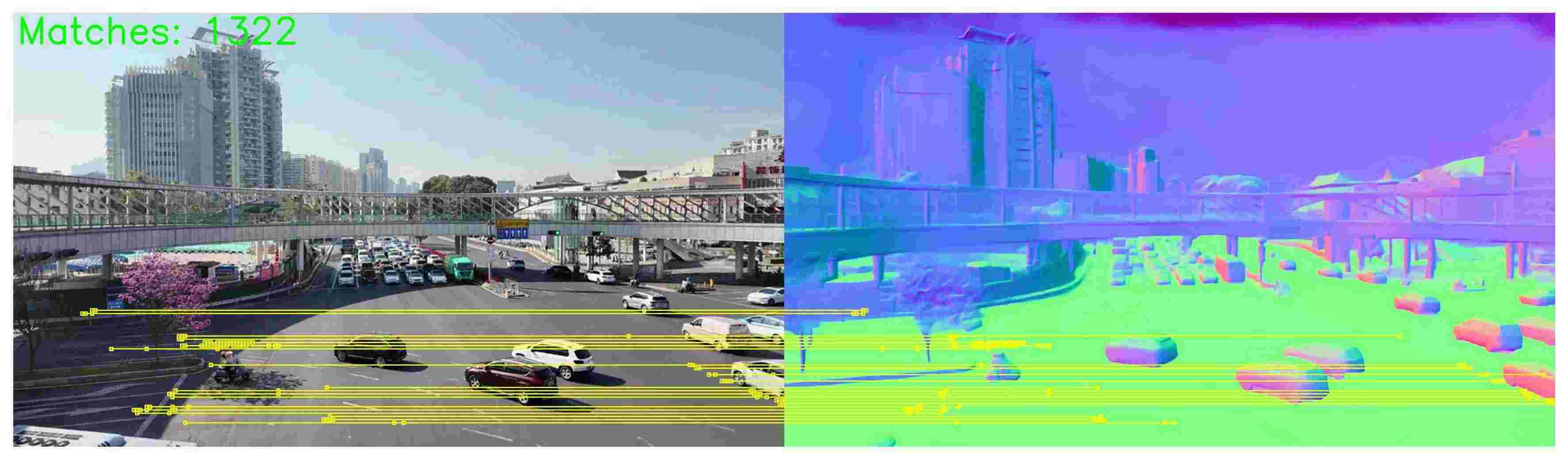} \\[2mm]

        \raisebox{0pt}{\rotatebox{90}{Paint}} &
        \includegraphics[width=0.23\textwidth]{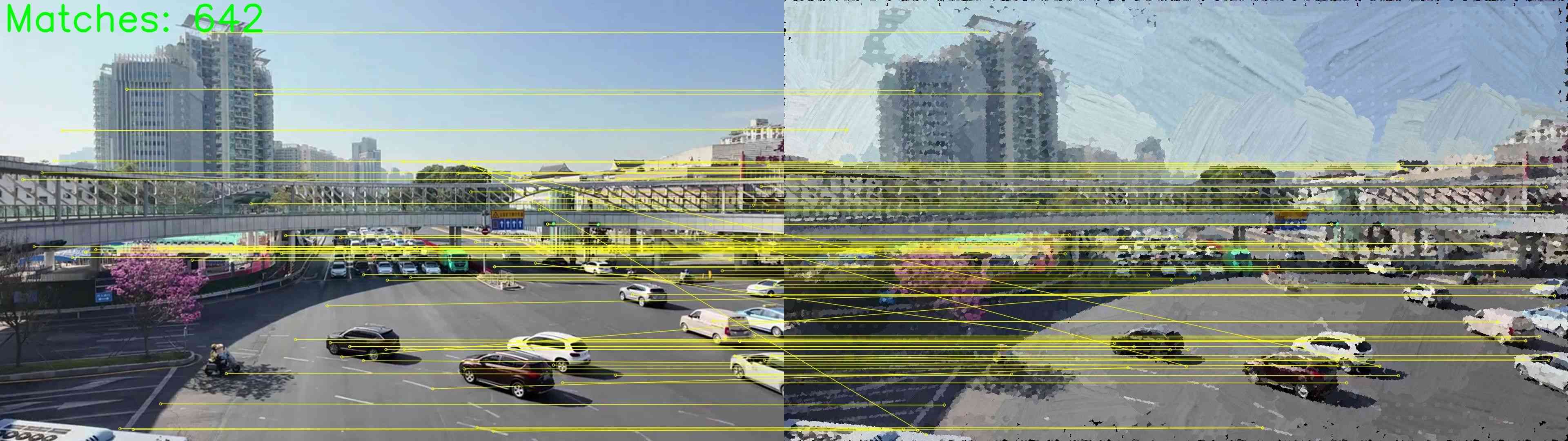} &
        \includegraphics[width=0.23\textwidth]{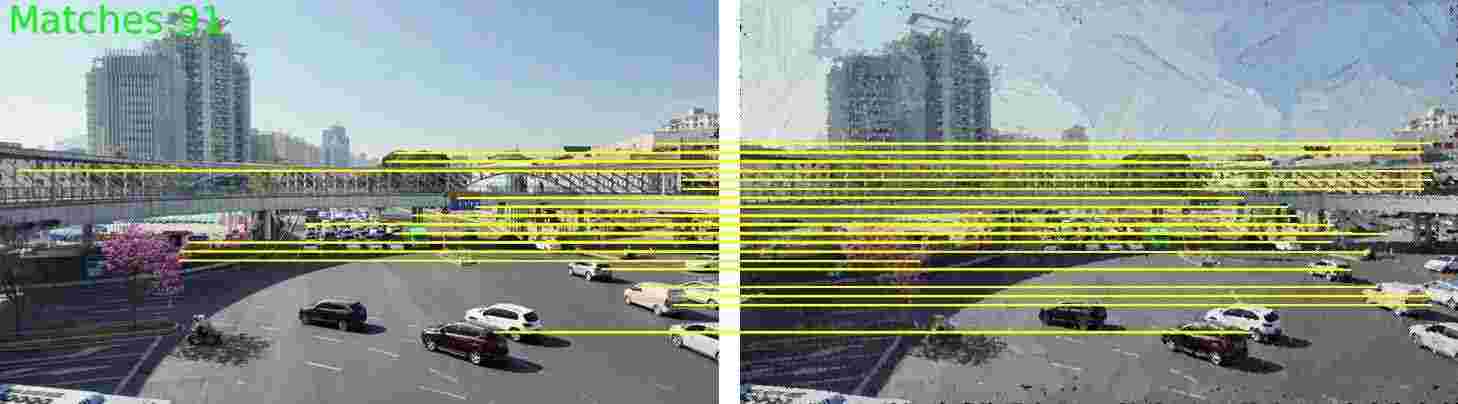} &
        \includegraphics[width=0.23\textwidth]{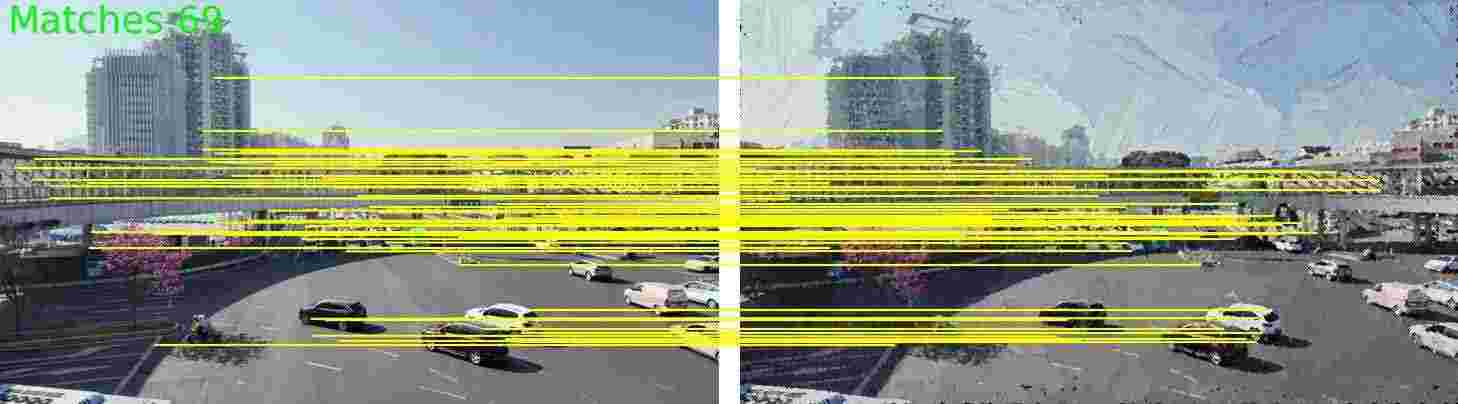} &
        \includegraphics[width=0.23\textwidth]{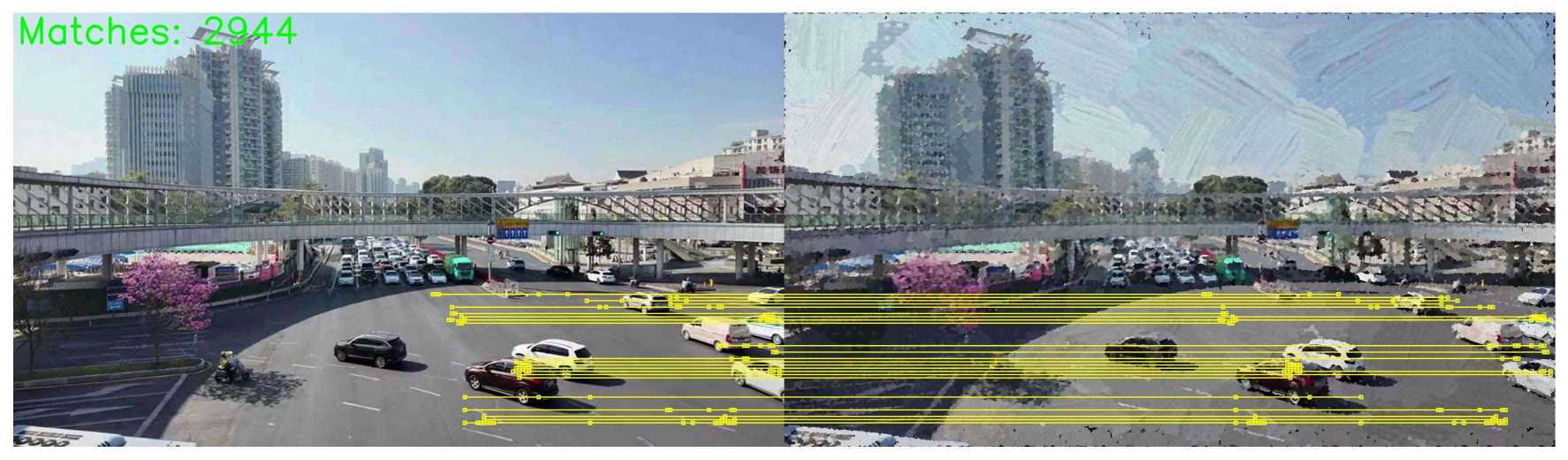} \\[2mm]

        \raisebox{0pt}{\rotatebox{90}{Sketch}} &
        \includegraphics[width=0.23\textwidth]{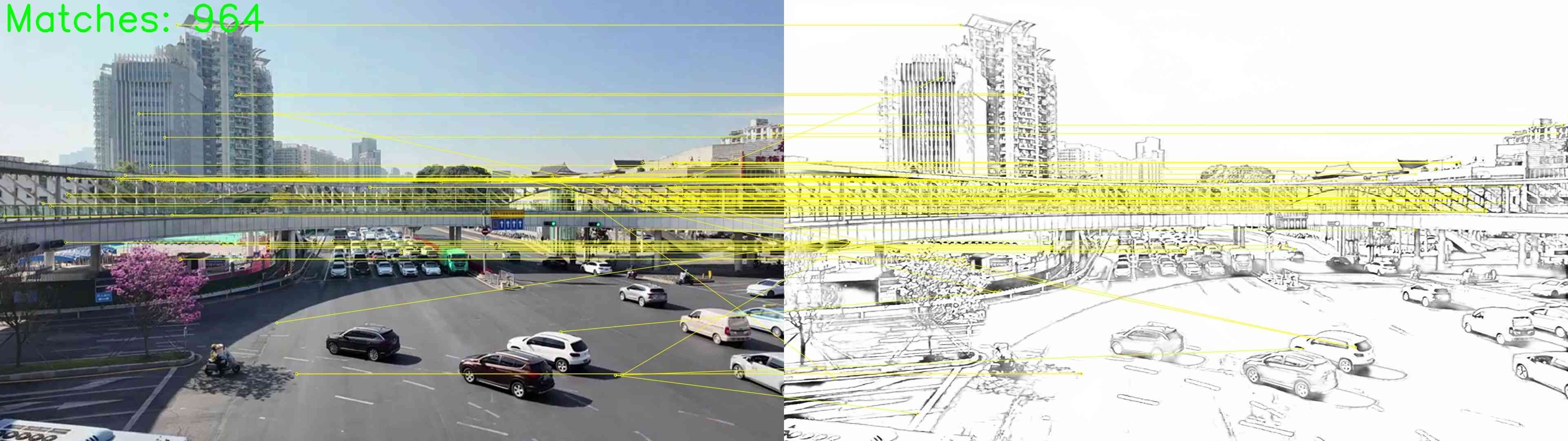} &
        \includegraphics[width=0.23\textwidth]{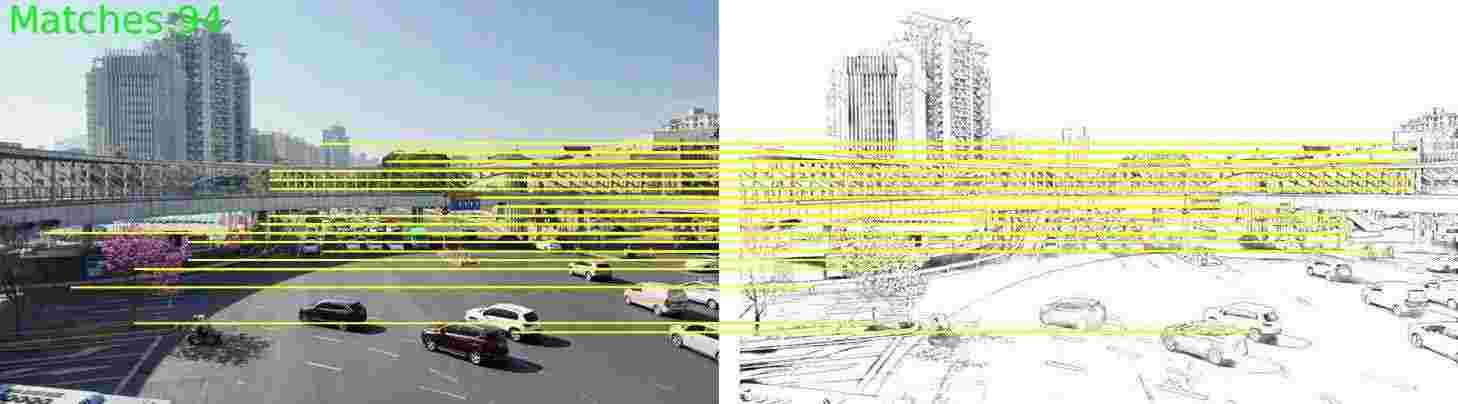} &
        \includegraphics[width=0.23\textwidth]{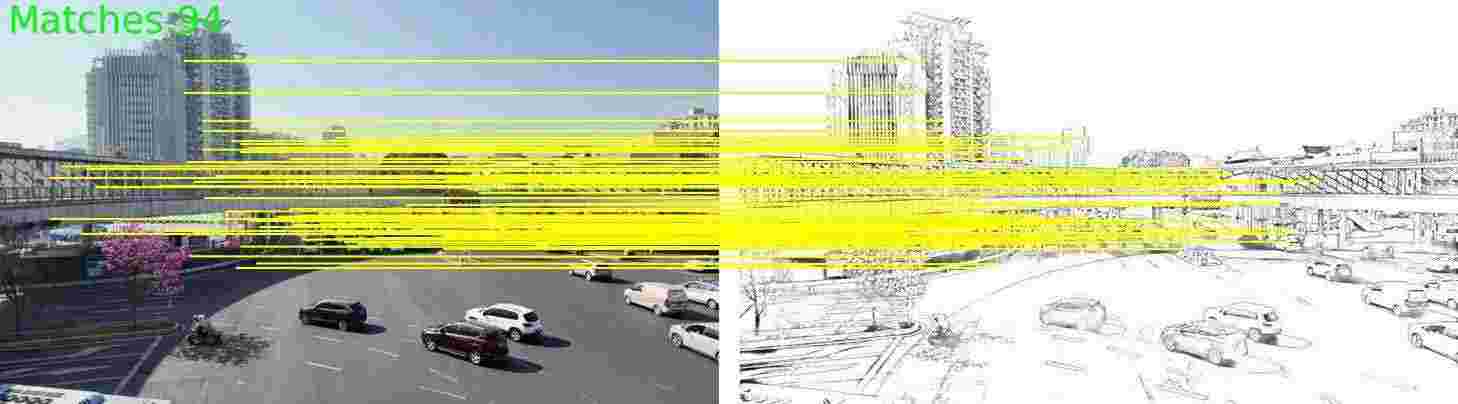} &
        \includegraphics[width=0.23\textwidth]{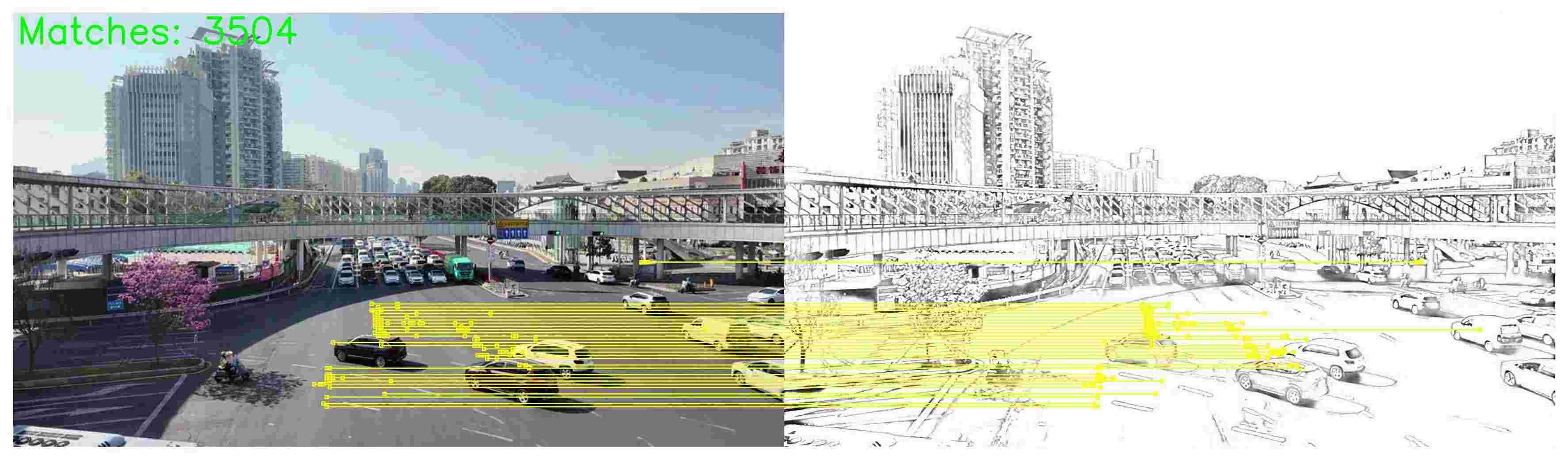}
    \end{tabular}
    
    \caption{Qualitative comparison of feature matching results across different modalities and methods(1).}
    \label{fig:more_modalities1}
\end{figure*}

\begin{figure*}[htbp]
    \centering
    \begin{tabular}{c@{\hspace{1mm}}c@{\hspace{1mm}}c@{\hspace{1mm}}c@{\hspace{1mm}}c}
        & SIFT & MINIMA\_LoFTR & MINIMA\_LG & Ours \\
        \raisebox{0pt}{\rotatebox{90}{Depth}} &
        \includegraphics[width=0.23\textwidth]{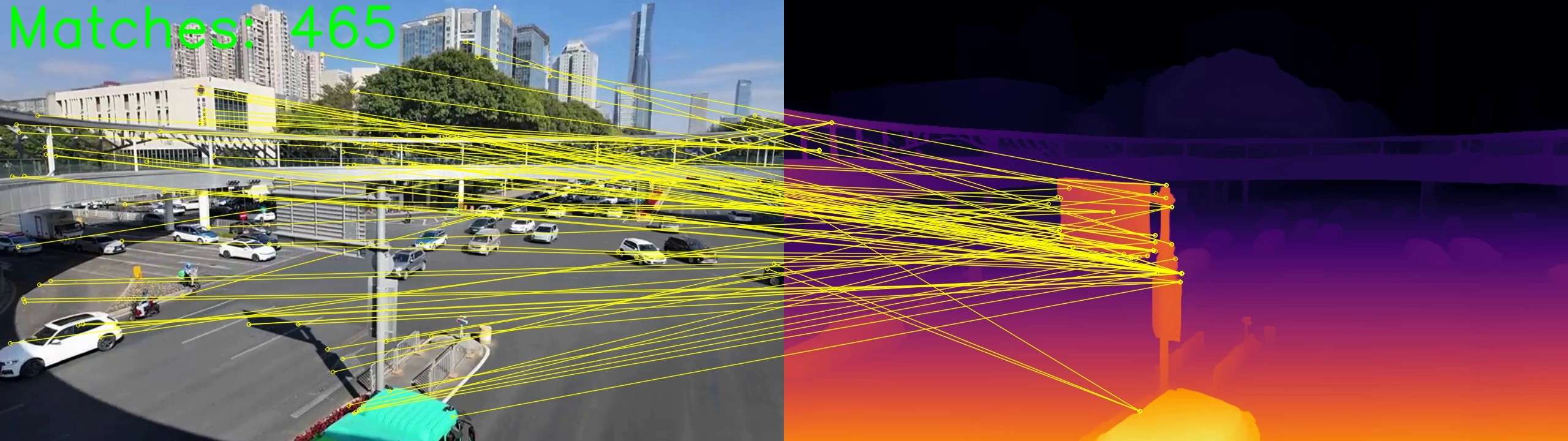} &
        \includegraphics[width=0.23\textwidth]{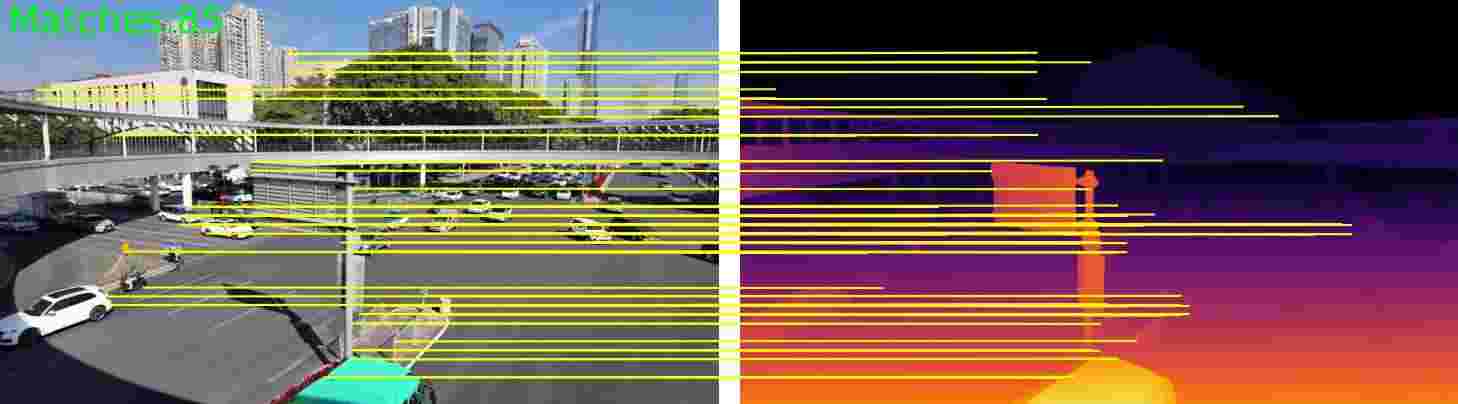} &
        \includegraphics[width=0.23\textwidth]{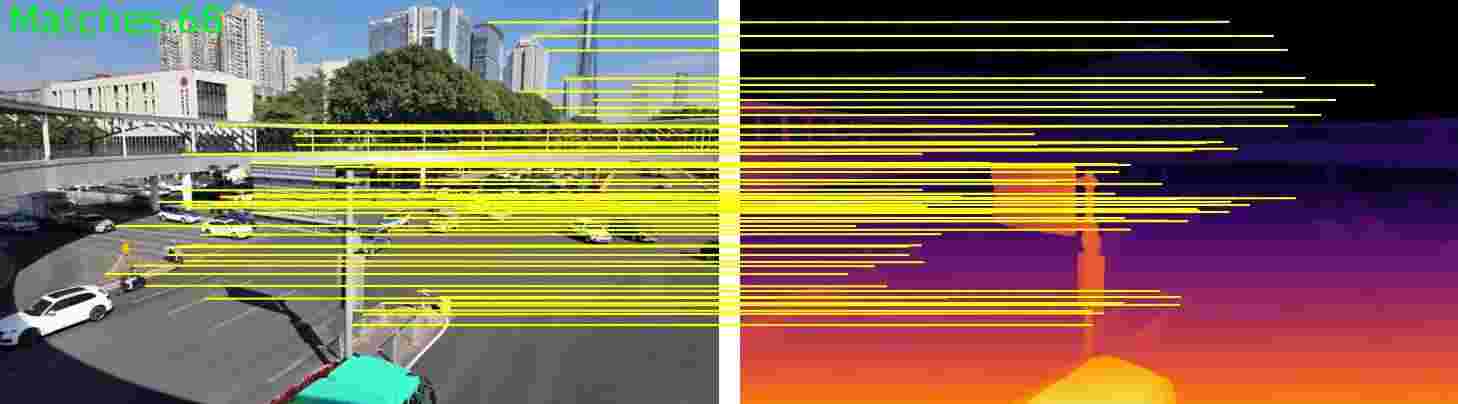} &
        \includegraphics[width=0.23\textwidth]{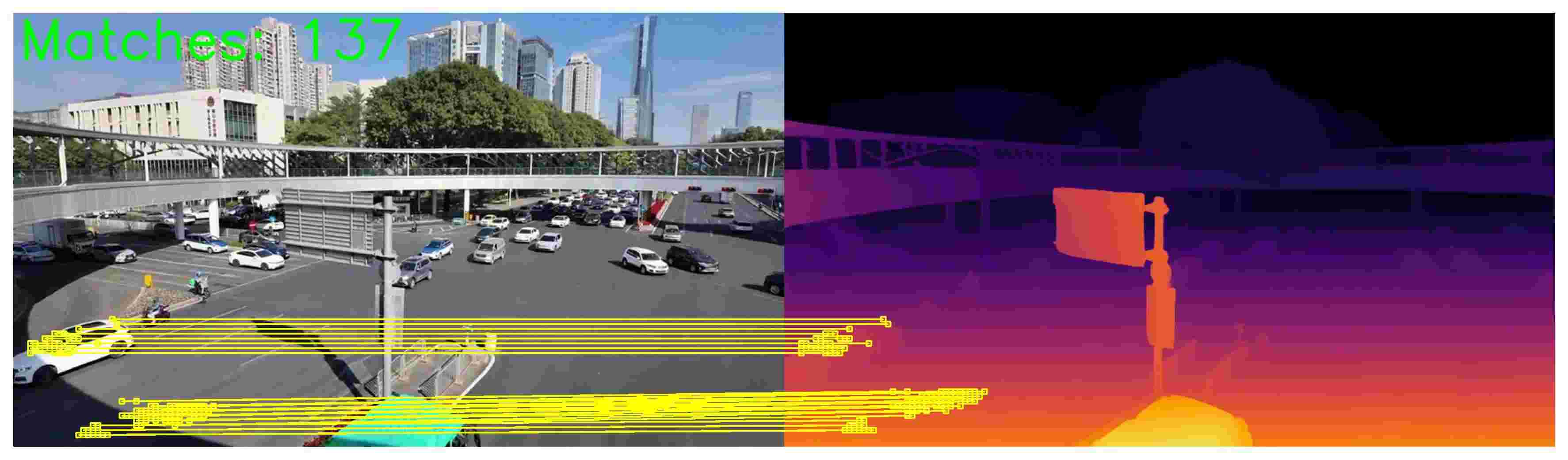} \\[2mm]

        \raisebox{0pt}{\rotatebox{90}{Event}} &
        \includegraphics[width=0.23\textwidth]{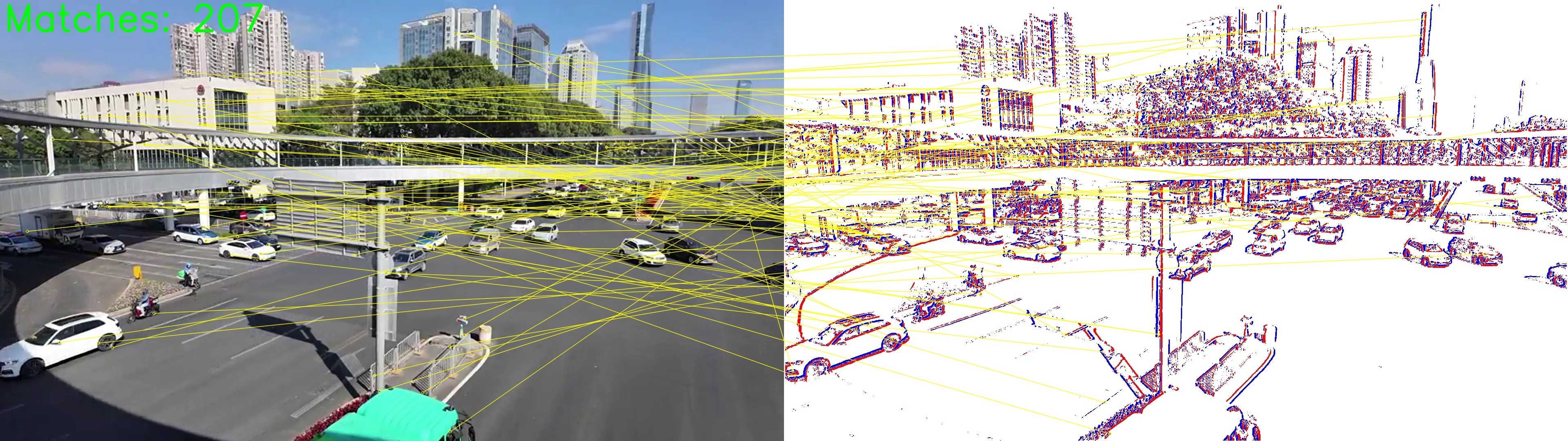} &
        \includegraphics[width=0.23\textwidth]{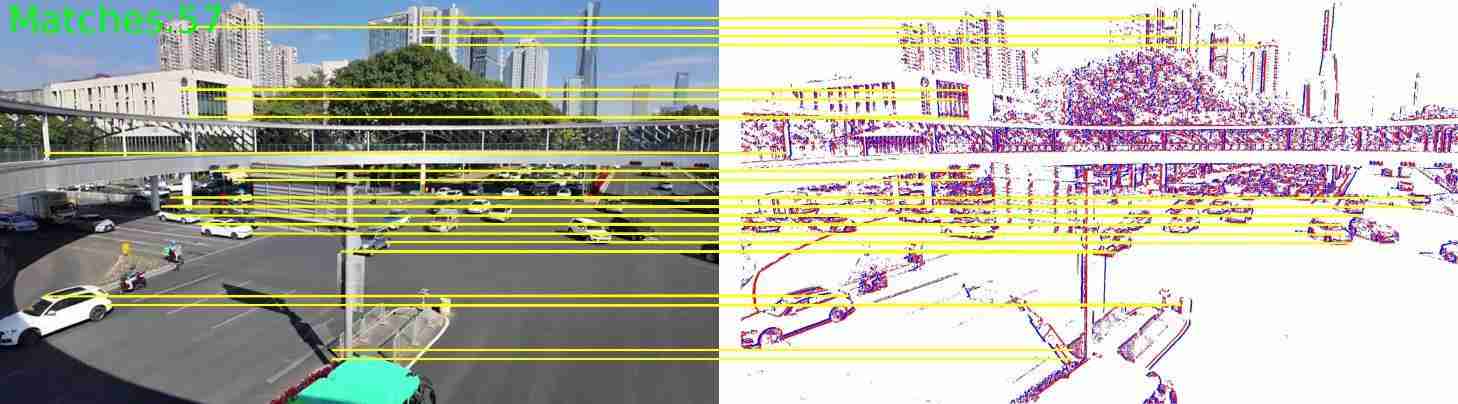} &
        \includegraphics[width=0.23\textwidth]{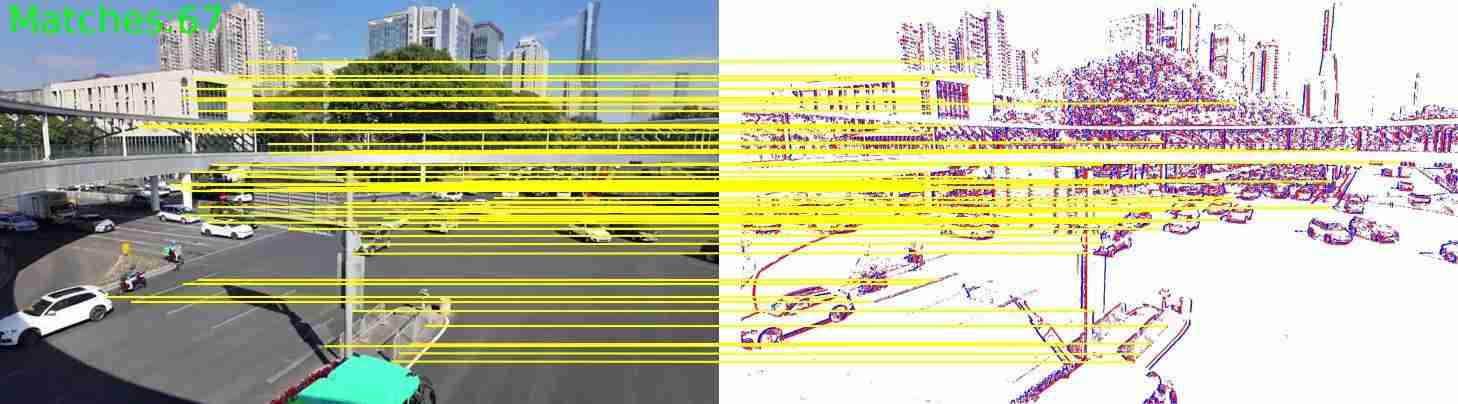} &
        \includegraphics[width=0.23\textwidth]{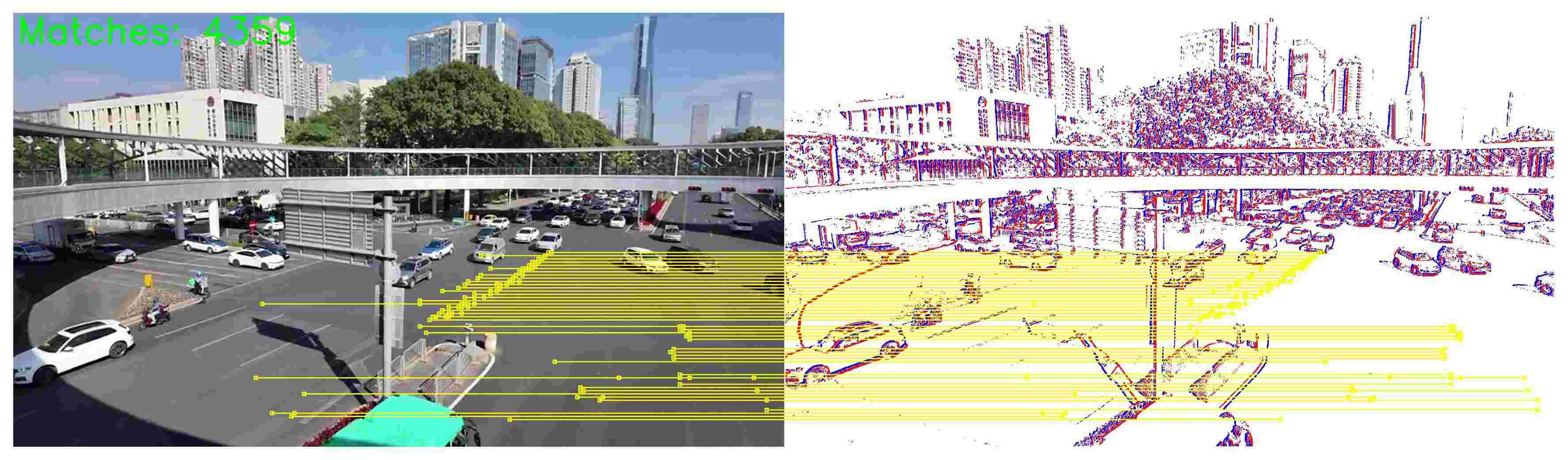} \\[2mm]

        \raisebox{0pt}{\rotatebox{90}{Normal}} &
        \includegraphics[width=0.23\textwidth]{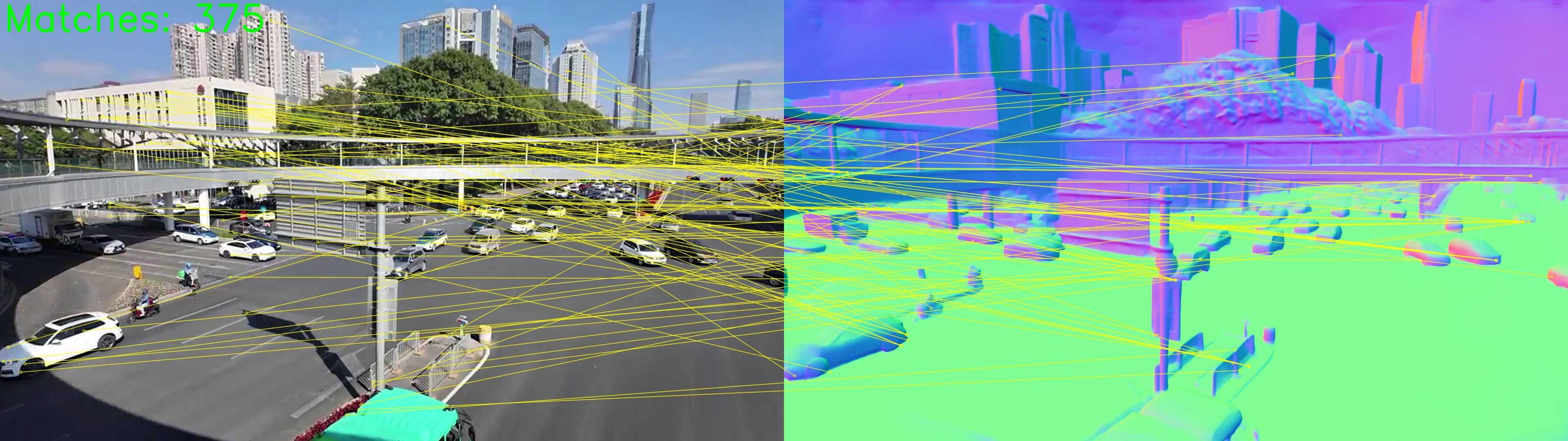} &
        \includegraphics[width=0.23\textwidth]{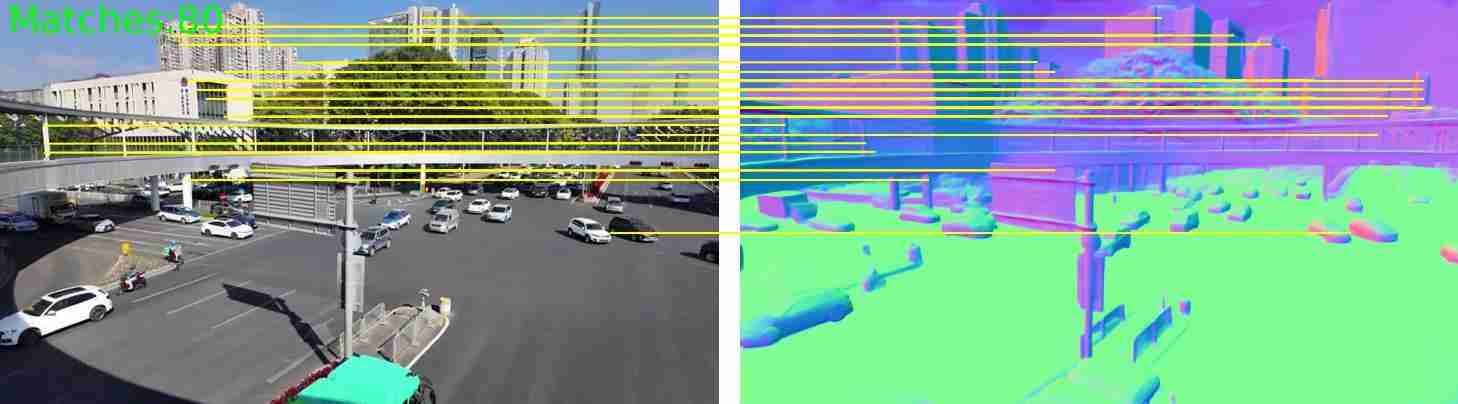} &
        \includegraphics[width=0.23\textwidth]{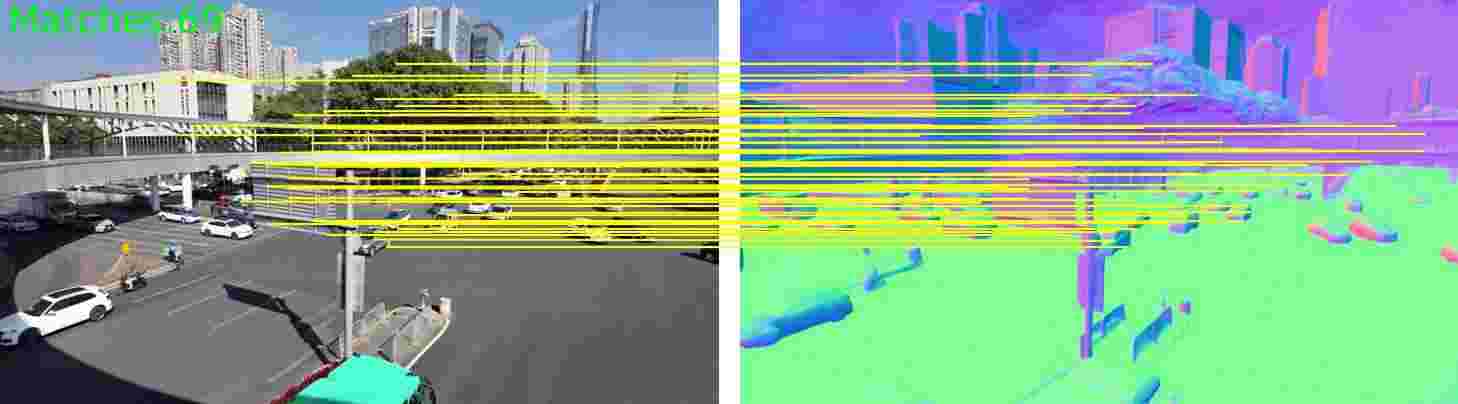} &
        \includegraphics[width=0.23\textwidth]{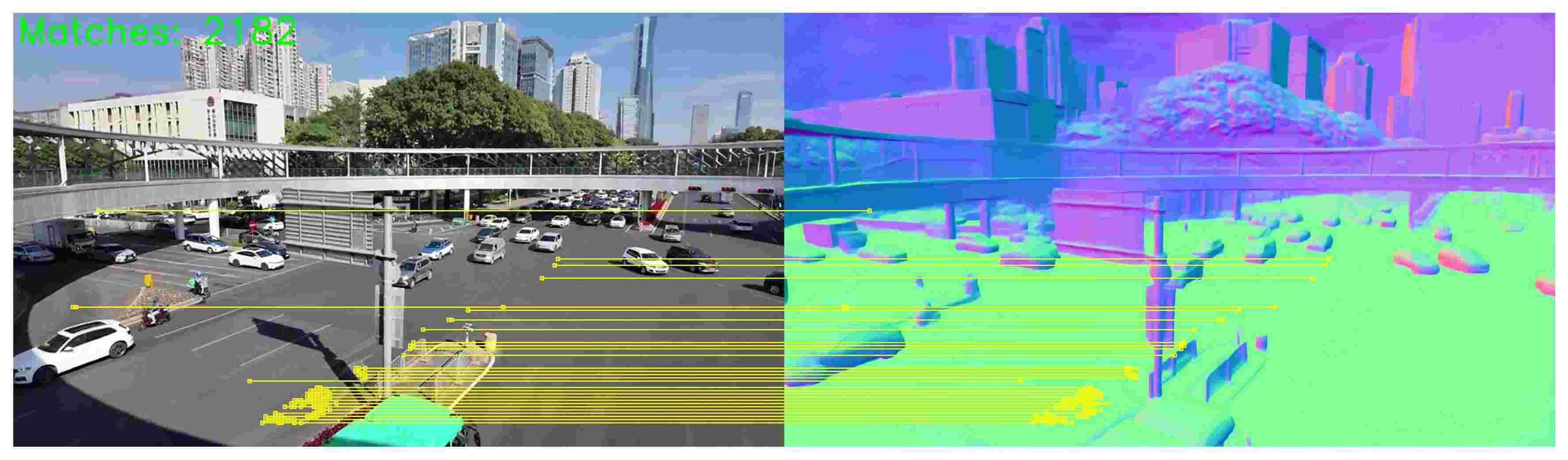} \\[2mm]

        \raisebox{0pt}{\rotatebox{90}{Paint}} &
        \includegraphics[width=0.23\textwidth]{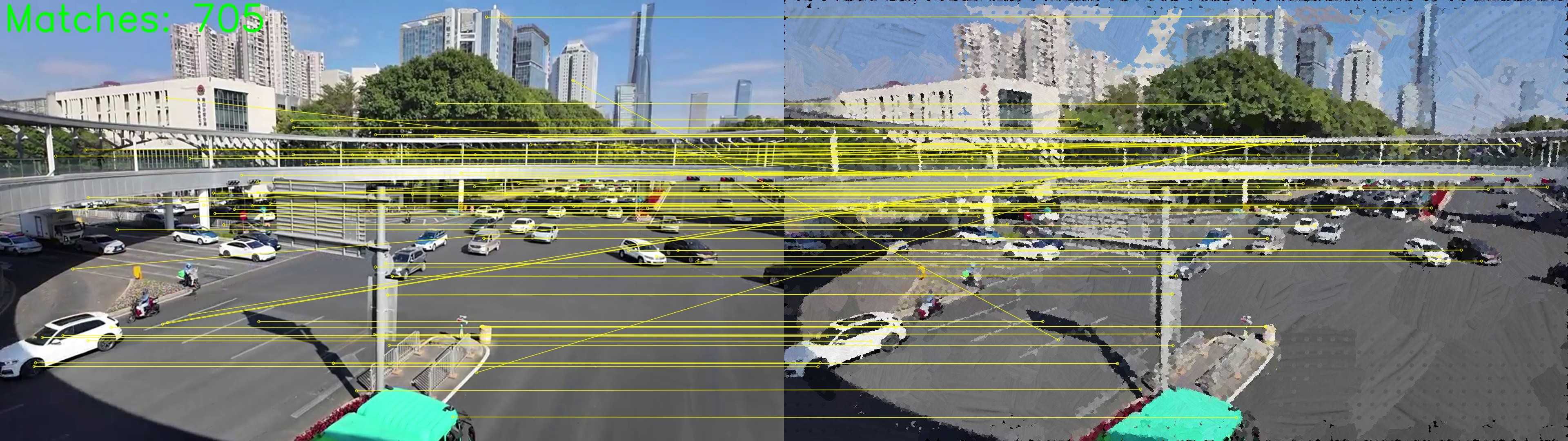} &
        \includegraphics[width=0.23\textwidth]{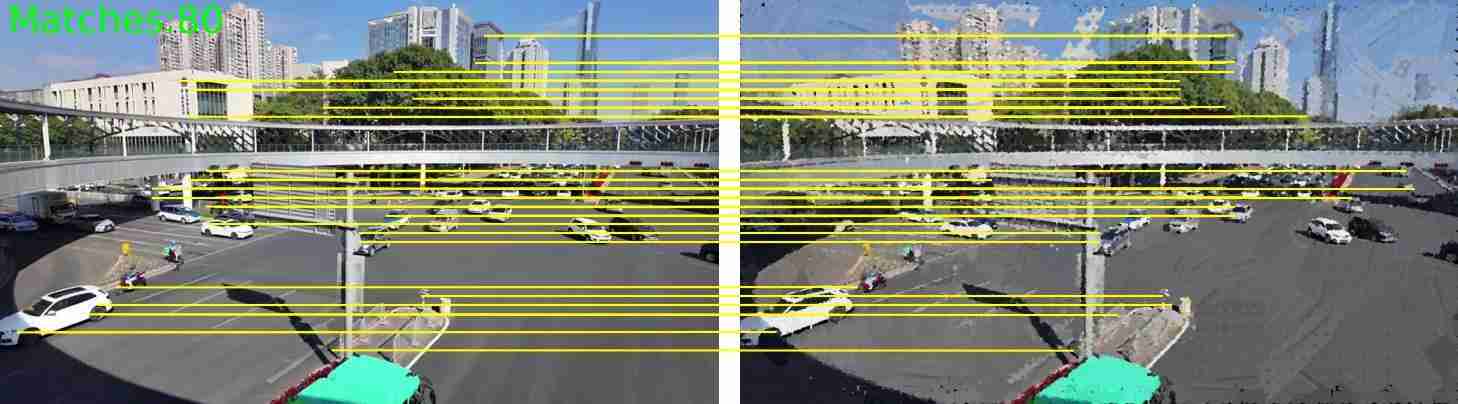} &
        \includegraphics[width=0.23\textwidth]{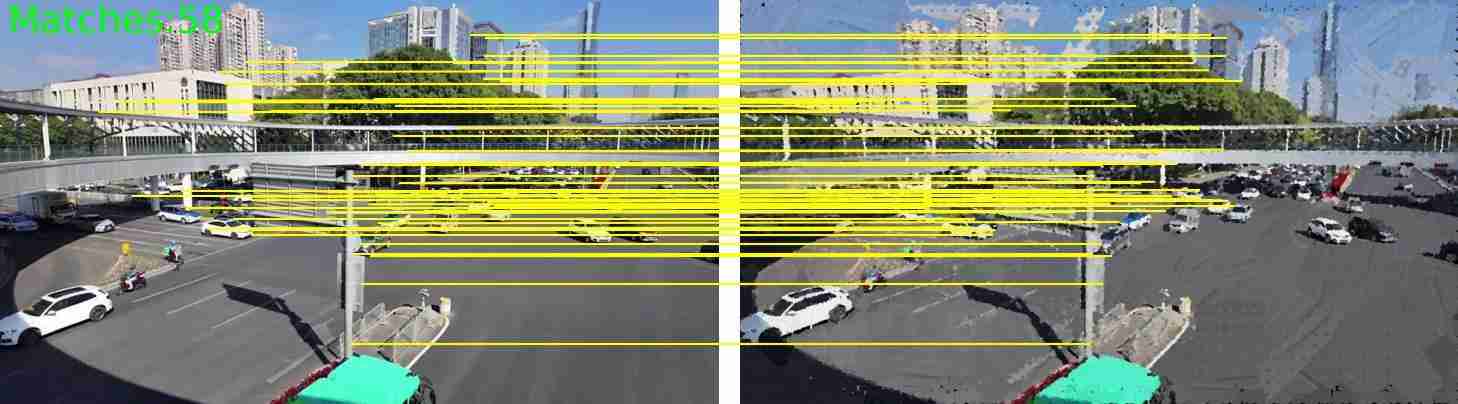} &
        \includegraphics[width=0.23\textwidth]{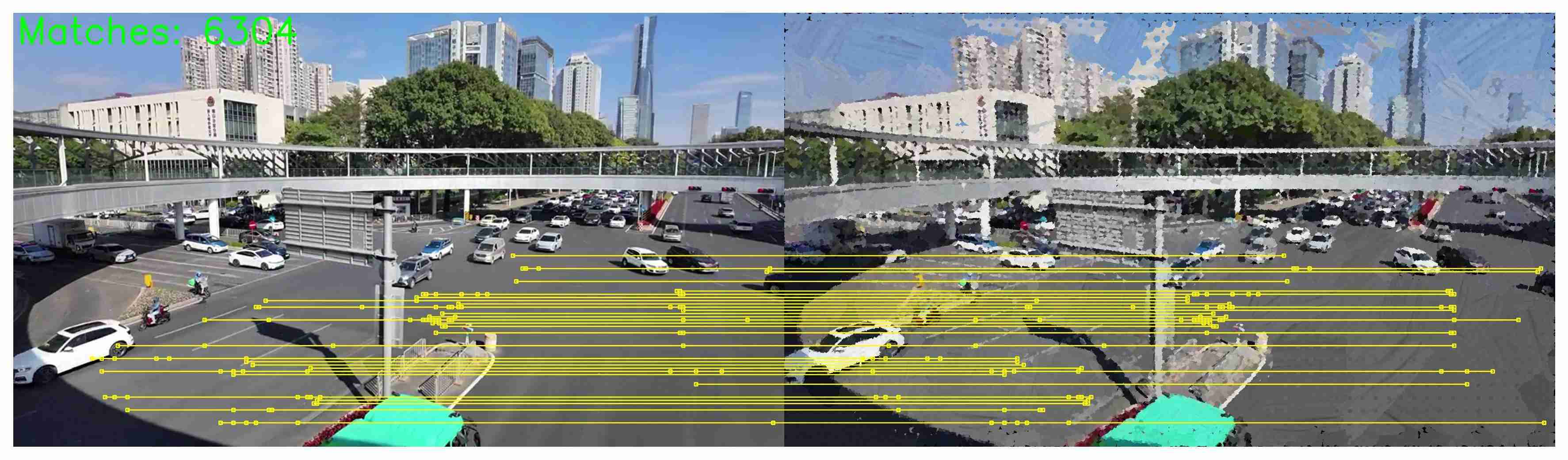} \\[2mm]

        \raisebox{0pt}{\rotatebox{90}{Sketch}} &
        \includegraphics[width=0.23\textwidth]{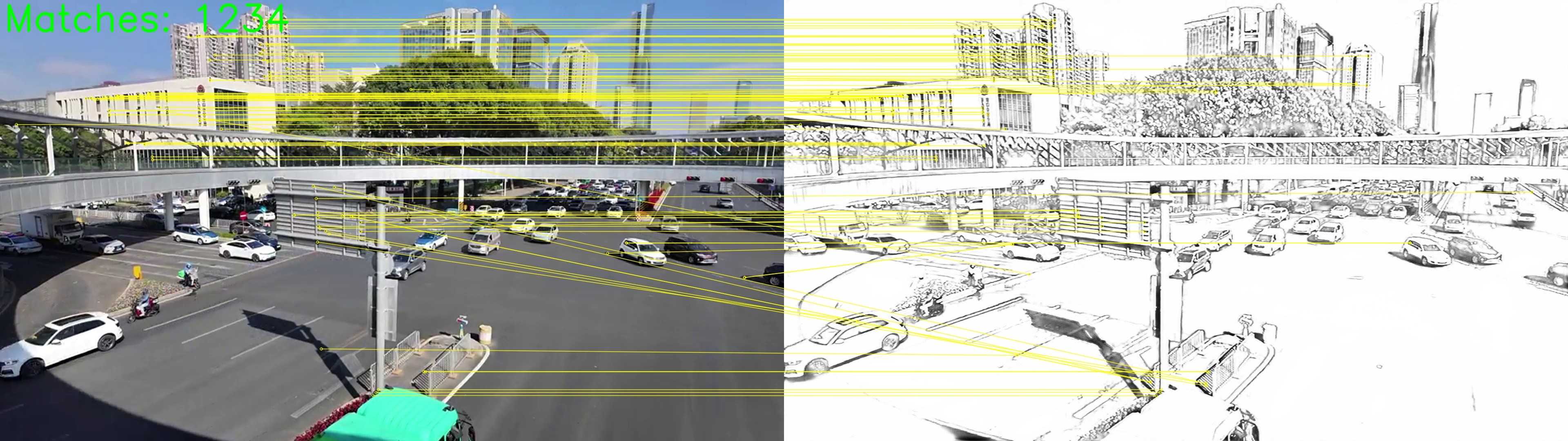} &
        \includegraphics[width=0.23\textwidth]{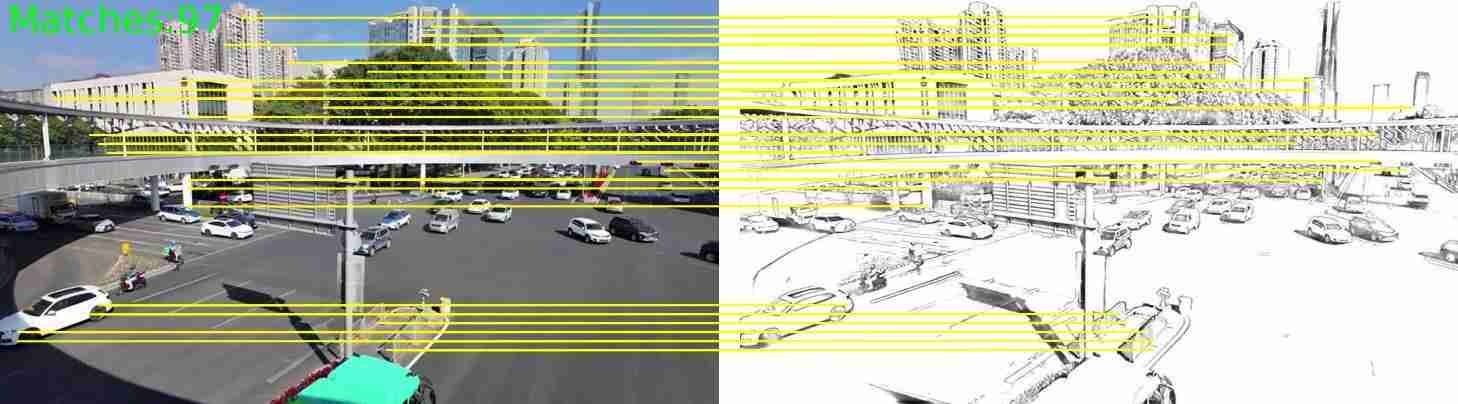} &
        \includegraphics[width=0.23\textwidth]{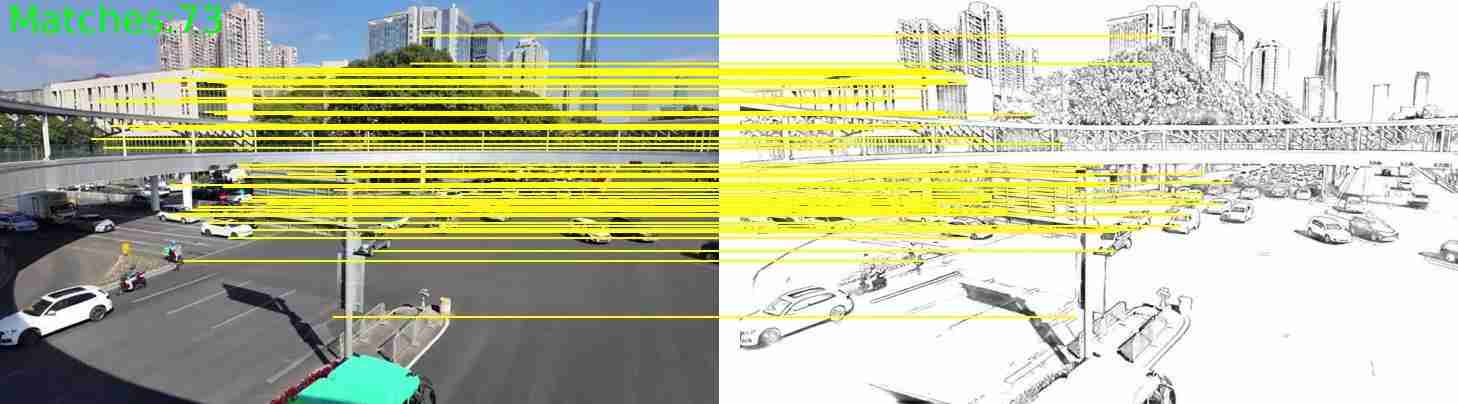} &
        \includegraphics[width=0.23\textwidth]{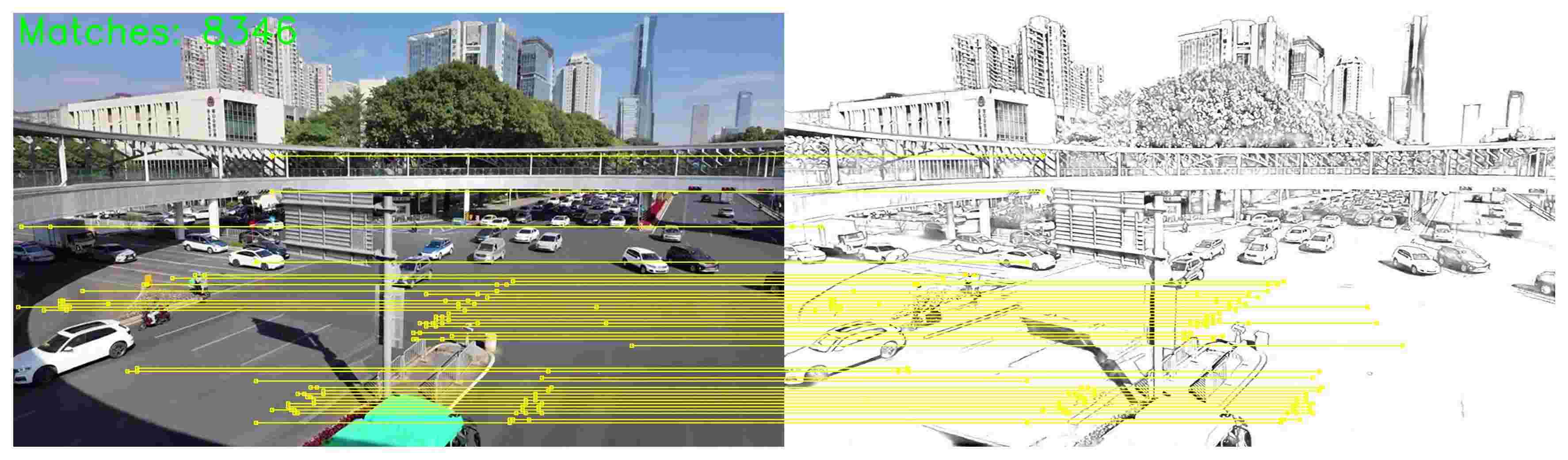}
    \end{tabular}
    
    \caption{Qualitative comparison of feature matching results across different modalities and methods(2).}
    \label{fig:more_modalities2}
\end{figure*}

Fig.~\ref{fig:more_modalities1} and Fig.~\ref{fig:more_modalities2} shows additional matching results across various modalities generated by the MINIMA Data Engine. It is important to note that although the Data Engine outputs single images, our input consists of continuous video frames. The generated images for other modalities may exhibit considerable inconsistencies between consecutive frames. As a result, when the generated image sequences are converted back into video form, severe flickering and jitter artifacts occur.

Such instability introduces significant interference to our method. Nevertheless, our approach still achieves reasonably good matching results, demonstrating its robustness in handling modality inconsistencies and temporal artifacts.

\end{document}